\documentclass[lettersize,journal]{IEEEtran}
\usepackage{amsmath,amsfonts}

\usepackage[linesnumbered,ruled,vlined]{algorithm2e}

\usepackage{array}
\usepackage{accents}
\usepackage{subfigure}

\usepackage{textcomp}
\usepackage{stfloats}
\usepackage{url}
\usepackage{verbatim}
\usepackage{graphicx}
\usepackage{cite}
\usepackage{float}
 \usepackage{tabularx} 

\def\BibTeX{{\rm B\kern-.05em{\sc i\kern-.025em b}\kern-.08em
    T\kern-.1667em\lower.7ex\hbox{E}\kern-.125emX}}
\usepackage{balance}

\usepackage{amssymb}
\usepackage{booktabs}
\usepackage{threeparttable}
\usepackage{multirow}
\usepackage{bigstrut}
\usepackage{bigdelim}
\usepackage{adjustbox}
\usepackage{makecell}
\usepackage{nicematrix}
\usepackage{xcolor}

\usepackage{tikz} 
\usetikzlibrary{arrows,shapes,chains}

\usepackage{colortbl}
\usepackage{rotating}

\DeclareMathOperator{\wrap}{wrap}
\DeclareMathOperator{\diag}{diag}

\begin{document}

\title{Occupancy-SLAM: An Efficient and Robust Algorithm for Simultaneously Optimizing Robot Poses and Occupancy Map}

\author{\authorblockN{Yingyu Wang, Liang Zhao, and Shoudong Huang} 
\thanks{Yingyu Wang and Shoudong Huang are with the Robotics Institute, University of Technology Sydney, Australia (e-mail: Yingyu.Wang-1@student.uts.edu.au; Shoudong.Huang@uts.edu.au).} 

\thanks{Liang Zhao was with the Robotics Institute, University of Technology Sydney, Australia, and is now with the School of Informatics, University of Edinburgh, Edinburgh, UK (e-mail: Liang.Zhao@ed.ac.uk).}}

\markboth{}
{Shell \MakeLowercase{\textit{et al.}}: A Sample Article Using IEEEtran.cls for IEEE Journals}


\maketitle


\begin{abstract}
Joint optimization of poses and features has been extensively studied and demonstrated to yield more accurate results in feature-based SLAM problems. However, research on jointly optimizing poses and non-feature-based maps remains limited. Occupancy maps are widely used non-feature-based environment representations because they effectively classify spaces into obstacles, free areas, and unknown regions, providing robots with spatial information for various tasks. In this paper, we propose Occupancy-SLAM, a novel optimization-based SLAM method that enables the joint optimization of robot trajectory and the occupancy map through a parameterized map representation. The key novelty lies in optimizing both robot poses and occupancy values at different cell vertices simultaneously, a significant departure from existing methods where the robot poses need to be optimized first before the map can be estimated. 

This paper focuses on 2D laser-based SLAM to investigate how to jointly optimize robot poses and the occupancy map. In our formulation, the state variables in optimization include all the robot poses and the occupancy values at discrete cell vertices in the occupancy map. Moreover, a multi-resolution optimization framework that utilizes occupancy maps with varying resolutions in different stages is introduced. A variation of Gauss-Newton method is proposed to solve the optimization problem at different stages to obtain the optimized occupancy map and robot trajectory. The proposed algorithm is efficient and converges easily with initialization from either odometry inputs or scan matching, even when only limited key-frame scans are used. Furthermore, we propose an occupancy submap joining method, enabling more effective handling of large-scale problems by incorporating the submap joining process into the Occupancy-SLAM framework. Evaluations using simulations and practical 2D laser datasets demonstrate that the proposed approach can robustly obtain more accurate robot trajectories and occupancy maps than state-of-the-art techniques with comparable computational time. Preliminary results in the 3D case further confirm the potential of the proposed method in practical 3D applications, achieving more accurate results than existing methods. The code is made available to benefit the robotics community\footnote{\url{https://github.com/WANGYINGYU/Occupancy-SLAM}}. 
\end{abstract}


\section{Introduction}


\IEEEPARstart{S}{imultaneous} localization and mapping (SLAM) is an important problem in robotics that has been studied for decades \cite{cadena2016past}. Jointly optimizing the robot poses and map can enhance SLAM performance, as this formulation utilizes the available information more directly without approximations. While joint optimization has been widely explored in feature-based SLAM (e.g., \cite{kaess2008isam,kaess2011isam2}), research on the joint optimization of robot poses and non-feature-based maps remains limited.

Occupancy grid maps are widely used in robotic tasks for their ability to clearly represent obstacles, free space, and unknown areas, facilitating collision-free navigation and path planning. Assuming the robot poses used to collect the sensor information are known exactly, the evidence grid mapping technique \cite{moravec1985high,
moravec1989sensor,elfes1989occupancy,martin1996robot,hornung2013octomap} provides an elegant and efficient approach for building occupancy grid maps from the collected information. However, when a robot is navigating in an unknown environment and performing SLAM, its own poses need to be estimated, and the estimates inherently contain uncertainties. Achieving both accurate robot localization and precise occupancy mapping simultaneously is not trivial.


In some occupancy grid map based SLAM approaches such as Cartographer \cite{hess2016real}, the problem is tackled in two steps. First, the robot poses are estimated by solving a pose-graph SLAM problem, where the relative pose measurements are derived using odometry, scan matching, loop closure detection, or other similar techniques. Second, the optimized poses are assumed to be the correct poses and are used to build up the map using evidence grid mapping techniques. However, in these two-step approaches, the uncertainties in the robot poses obtained during the first step are not considered when building the map. Therefore, it is crucial to achieve highly accurate pose estimates to construct a reliable occupancy grid map. As a result, it can be expected that the occupancy map obtained using a two-step approach may not represent the best result that one can achieve using all the available information.

In feature-based SLAM approaches, jointly optimizing the poses and the feature map is common, as the relationship between observations and the map is straightforward to model. However, for occupancy map based SLAM, jointly optimizing the robot poses and the occupancy map is not trivial because: 
\begin{itemize}
	\item [1)] \textbf{The relation between the observations and the map is complex.} The observations are laser beams (the endpoint of a beam represents ``hit" and the other positions along the beam represent ``free"), and the map is a function representing the occupancy values at different positions. This is significantly different from feature-based SLAM where both the observations and the map are about feature parameters such as positions.
	\item [2)] \textbf{The data association is not easy to do.} When the robot poses are noisy, the correct correspondence between laser beams and occupancy grid cells is hard to find. In contrast, for feature-based SLAM, there are well-established front-end methods for data association.
	\item [3)] \textbf{The resolution of the map has a significant impact on the optimization problem.} A high-resolution map helps to establish a more accurate connection between the observations and the map, but it leads to a sharp increase in the number of variables. However, for feature-based SLAM, there is no such issue.
\end{itemize}


\subsection{Contributions}
In this paper, we propose Occupancy-SLAM algorithm, which jointly optimizes the robot poses and the occupancy map using 2D laser scans (and odometry) information. Moreover, we propose a multi-resolution optimization framework for improving convergence and robustness to initial guesses. To better handle the case of large-scale environments and long-term trajectories, we further propose an occupancy submap joining method. Experiments on both simulated and practical datasets verify the superior performance of our method compared with state-of-the-art approaches (e.g., Cartographer \cite{hess2016real}). In addition, we extend our method to the 3D case, and preliminary results confirm its effectiveness in improving accuracy. The main contributions are summarized as follows: 


\begin{enumerate}
	\item We formulate the occupancy grid map based SLAM problem as a joint optimization problem where the poses and the occupancy map are optimized together. 
	\item We propose a variation of Gauss-Newton method to solve the new formulation, enabling the estimation of more accurate robot poses and occupancy maps compared to existing state-of-the-art techniques.
	\item To enhance efficiency, convergence, and robustness, we propose a multi-resolution optimization strategy using occupancy maps of different resolutions across stages.
    
    
    \item We propose a submap joining algorithm to address the cases of large-scale environments and long-term trajectories through our joint poses and occupancy map optimization idea.
	\item Our method achieves robust convergence even with key frames of limited overlap, outperforming state-of-the-art approaches like Cartographer in efficiency while maintaining superior accuracy.
    \item We extend our method to 3D, with preliminary results demonstrating superior accuracy compared to other state-of-the-art approaches.
\end{enumerate}

This paper is an extension to our conference paper \cite{Zhao-RSS-22}, with major improvements in contributions 3, 4, 5, and 6, significantly enhancing the robustness and efficiency of the algorithm while extending the method to 3D.


\begin{figure}
\centering
\includegraphics[width=0.49\textwidth]{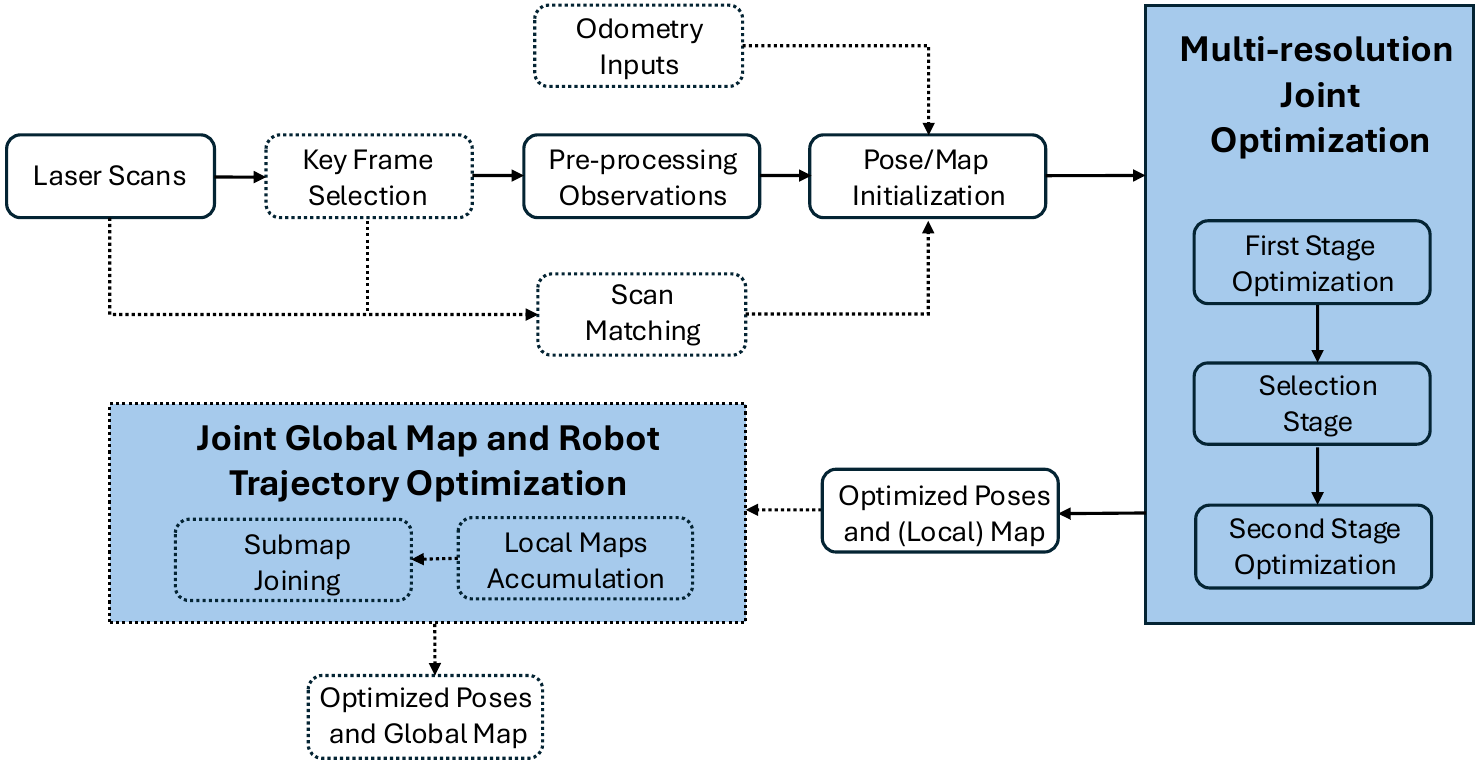}
\caption{\label{fig_overview} Main components of our proposed method. The blue-colored components represent our core approaches, while the dashed portions are optional. Our multi-resolution joint optimization is covered in Section \ref{sec_formulation}, Section \ref{Sec_Algorithm_1}, and Section \ref{Sec_multi}. The joint global map and robot trajectory optimization approach is presented in Section \ref{Sec_submap}. }
\end{figure}


\subsection{Outline}
Fig. \ref{fig_overview} illustrates the flowchart of applying our proposed methods in practice. The blue components represent our core approaches, while the dashed portions are optional. The rest of the paper is organized as follows: Section \ref{Sec_related_work} provides a review of related work on non-feature-based SLAM, submap joining, and joint optimization of poses and maps. In Section \ref{sec_formulation}, we introduce our novel formulation for jointly optimizing the robot poses and occupancy map. A variation of the Gauss-Newton method to solve our nonlinear least squares (NLLS) formulation is presented in Section \ref{Sec_Algorithm_1}. In Section \ref{Sec_multi}, we introduce our multi-resolution strategy to improve the efficiency and robustness of the algorithm. Section \ref{Sec_submap} presents our submap joining algorithm for handling large-scale environments and long-term trajectories. Experimental results are provided in Section \ref{Sec_experiment}. In Section \ref{sec_3d}, we extend our method to the 3D case and present preliminary results. Finally, the conclusions are given in Section \ref{Sec_conclusion}.

\section{Related Work}\label{Sec_related_work}

In this section, we discuss some related work on non-feature based map representations for SLAM, submap joining techniques, and joint optimization of poses and maps. 

\subsection{Non-feature based map representations for SLAM}\label{Sec_related_a}
One widely used non-feature based SLAM approach is occupancy grid map-based SLAM, which probabilistically classifies spaces into obstacles, free areas, and unknown regions while accounting for uncertainty during observation updates \cite{moravec1985high, moravec1989sensor, elfes1989occupancy, martin1996robot, hornung2013octomap}. Classic examples, such as FastSLAM \cite{montemerlo2002fastslam} and GMapping \cite{grisetti2005improving}, use particle filters for mapping and localization but struggle with high computational demand and long-term accuracy in large-scale environments.

Recent optimization-based approaches, such as Hector SLAM \cite{kohlbrecher2011flexible}, Karto-SLAM \cite{konolige2010efficient}, and Cartographer \cite{hess2016real}, address cumulative errors effectively. Hector SLAM uses scan-to-map matching but lacks loop closure, restricting it to small-scale scenarios. Karto-SLAM incorporates loop closure detection with sparse pose adjustment for global optimization, while Cartographer integrates scan-to-map matching and pose graph optimization with a branch-and-bound strategy for efficient loop closure detection. However, by treating pose optimization and map construction as independent processes, these methods fail to account for the interdependencies of their uncertainties.

Multi-resolution occupancy mapping techniques can be integrated into occupancy grid map based SLAM frameworks to enable a more compact and efficient mapping process. For instance, approaches like OctoMap \cite{hornung2013octomap} use memory-efficient octrees to balance map compactness and accessibility. Adaptive-resolution methods, such as RMAP \cite{khan2014rmap} and ColMap \cite{fisher2021colmap}, dynamically adjust grid resolution to enhance mapping efficiency. Recently, \cite{Reijgwart-RSS-23} applies wavelet compression for hierarchical occupancy map storage, allowing efficient updates and queries. However, integrating multi-resolution maps as state variables into a unified framework for joint poses and map optimization remains an open challenge.

Another widely used non-feature-based map is the signed-distance function (SDF), which discretizes the environment into grid cells storing the distance to the nearest surface. This representation encodes the space, with the object surfaces defined by the zero crossings of the distance functions \cite{curless1996volumetric}. Some SLAM systems adopt SDF to improve localization accuracy and mapping quality. For example, supereight \cite{vespa2018efficient} integrates tracking, mapping, and planning using an octree-based truncated SDF (TSDF). It aligns camera frames to the TSDF map with iterative closest point (ICP) \cite{besl1992method}. A follow-up work \cite{vespa2019adaptive} improves this by introducing adaptive-resolution octree structures, achieving denser environment representation and reduced noise, leading to more accurate localization.

Other non-feature based map representations have also been used in SLAM, including mesh-based \cite{rosinol2021kimera}, normal distributions transform based \cite{einhorn2015generic}, neural radiance fields based \cite{rosinol2023nerf} and Gaussian splatting based \cite{matsuki2024gaussian}. Although these approaches differ in the type of non-feature representations they use, they all aim to provide more effective environmental modeling, improve robot localization accuracy, or achieve both.

However, all the optimization-based SLAM approaches that utilize non-feature based maps need to optimize the poses first and then build the non-feature based map using the optimized poses. This separation prevents these approaches from jointly considering the uncertainties in both localization and mapping during the optimization process. In contrast, this paper considers unifying the optimization of both the robot poses and occupancy values at each cell vertex of the occupancy map into a single optimization problem, which can be expected to yield better accuracy.

\subsection{Submap Joining}\label{Sec_related_b}

Submap joining, as proposed by \cite{bosse2003atlas}, is a widely used scheme for SLAM in large-scale environments due to its efficiency and reduced risk of being trapped in local minima compared to full optimization-based SLAM. Feature-based submap joining approaches \cite{huang2008sparse,zhao2013linear,wang2019submap} have been well investigated, with many demonstrating properties that enable efficient problem-solving while maintaining a high level of accuracy. To extend non-feature-based SLAM to large-scale environments and long-term operations, recent efforts have explored non-feature-based submap joining methods.

For example, \cite{wagner2014graph} divides the environment into overlapping submaps composed of small TSDF grids from KinectFusion \cite{newcombe2011kinectfusion}. Submap joining is then formulated as a pose graph optimization problem, where submap poses are nodes, and relative transformations from ICP are edges. Similarly, VOG-map \cite{ho2018virtual} represents submaps as 3D occupancy grids, converts them to point clouds for ICP-based relative transformations, and solves submap joining via pose graph optimization. Voxgraph \cite{reijgwart2019voxgraph} improves accuracy by employing SDF-to-SDF registration for overlapping submaps created with C-blox \cite{millane2018c}. Unlike time-sequence-based submap partitioning, \cite{wang2021elastic} uses spatial partitioning, merging submaps during loop closures by solving a pose graph containing only submap frames, with reconstruction decisions based on environmental changes.

All the aforementioned non-feature-based submap joining approaches estimate relative measurements between overlapping submaps to formulate and solve the pose graph problem for submap frames. In contrast, this paper jointly optimizes submap frames and the global occupancy map.

\subsection{Joint Optimization of Poses and Maps}
Joint optimization of poses and maps can result in better accuracy, as it utilizes the information more directly. In feature-based SLAM and bundle adjustment approaches, the most common form is to jointly optimize poses and positions of features, such as \cite{dellaert2006square,triggs2000bundle,konolige2008frameslam,sibley2009adaptive,zhao2015parallaxba}. Some approaches extend this idea to planar feature parameters. For instance, \cite{kaess2015simultaneous,hsiao2017keyframe} minimize the difference between plane measured in a scan and predicted planes, while \cite{trevor2012planar,geneva2018lips,zhou2021pi,zhou2021lidar} minimize the Euclidean distance between points in a scan and the predicted planes. Based on the idea of minimizing Euclidean distance between points in scans, BALM \cite{liu2021balm} demonstrates that planar parameters can be solved analytically in closed form, reducing the dimensionality of the optimization. BALM2 \cite{liu2023efficient} further improves efficiency by using point clusters, avoiding individual point enumeration. HBA \cite{liu2023large} introduces a hierarchical structure to address the scalability challenges of BALM and BALM2 in large environments. In summary, jointly optimizing poses and feature-based maps is well-studied, as features naturally link positions, observations, and poses, making them straightforward to integrate into optimization problems. In contrast, establishing constraints between observations, poses, and non-feature-based maps (e.g., occupancy grid maps) for joint optimization remains a significant challenge.



Research on jointly optimizing the poses and non-feature based maps is limited. Kimera-PGMO proposed in \cite{rosinol2021kimera} represents a notable attempt, integrating pose optimization with mesh deformation. It constructs a deformation graph of a simplified mesh and a pose graph, formulating the problem as a factor graph solvable by GTSAM \cite{dellaert2012factor}. 
While Kimera-PGMO \cite{rosinol2021kimera} has similar motivations as our paper, aiming to achieve better quality maps and more accurate poses through joint optimization, its mesh-based representation differs fundamentally from the occupancy grid maps used in our approach. Meshes are naturally represented through point positions and their relationships, which facilitates factor graph formulations.


\begin{table}[t]
 		\centering
 		\caption{Summary of Some Important Notations.}\label{tab_notation}
 		\setlength{\tabcolsep}{0.5 mm}{
 		\begin{tabular}{|c|l|p{3cm}p{3cm}p{3cm}}
   \hline
   \multicolumn{1}{|c|}{Notation} & \multicolumn{1}{|c|}{Explanation} \\ \hline
   $\mathbb{M}$  & \begin{tabular}[c]{@{}l@{}} A set includes occupancy values at all discrete cell vertices in \\occupancy map, as defined in Section \ref{sec_discrete_occupancy}. $\mathbb{M}^{l}$, $\mathbb{M}^{h}$, and $\mathbb{M}^{s}$ \\represent the sets include occupancy values at all cell vertices \\in low-resolution map, high-resolution map and selected \\high-resolution map, respectively. In addition, $\mathbb{M}_L$ and $\mathbb{M}_G$ \\represent the sets including occupancy values of all cell vertices \\in local maps and the global map, as defined in Section \ref{Sec_submap}.\end{tabular}\\ \hline

   $M(\cdot)$ &\begin{tabular}[c]{@{}l@{}}A function to lookup occupancy value at an arbitrary position in \\the occupancy map by bilinear interpolation using $\mathbb{M}$.\end{tabular} \\ \hline

    $\mathbf{x}^M$ & \begin{tabular}[c]{@{}l@{}} A vector including occupancy values at all cell vertices in discrete \\occupancy map $\mathbb{M}$, as defined in (\ref{eq_map_state}). $\mathbf{x}^{lM}$ and $\mathbf{x}^{sM}$ are vectors \\including occupancy values at cell vertices in $\mathbb{M}^{l}$ and $\mathbb{M}^{s}$. In \\addition, $\mathbf{x}^M_G$ represents the vector which includes occupancy \\values at cell vertices from $M_G$, as described in Section \ref{Sec_submap}.\end{tabular} \\ \hline

    $N(\cdot)$ & \begin{tabular}[c]{@{}l@{}} A function to lookup hit number at arbitrary position in the map \\by bilinear interpolation using hit map $\mathbb{N}$, where $\mathbb{N}$ is defined as a \\set includes hit number at all discrete cell vertices in the map, as \\described in Section \ref{sec_hit}. $\mathbb{N}^{l}$ and $\mathbb{N}^{s}$ represent hit maps used in \\different optimization stages.\end{tabular}\\ \hline

    $\mathbf{x}^P$ & \begin{tabular}[c]{@{}l@{}} A vector including all robot poses for optimization, as defined \\in (\ref{eq_pose_state}). In addition, $\mathbf{x}^P_L$ denotes a vector including all local map \\coordinate frames for submap joining problem in Section \ref{Sec_submap}.\end{tabular} \\ \hline
   \rule{0pt}{1.5em}
    $\mathbf{x}$ & \begin{tabular}[c]{@{}l@{}} State vector of optimization, $\mathbf{x} = {\left[{\mathbf{x}^P}^\top, {\mathbf{x}^M}^\top\right]}^\top$. $\mathbf{x}^{l}$ and $\mathbf{x}^{s}$ \\represent state vectors of different optimization stages. \end{tabular} \\ \hline

    $\mathbb{S}$  &\begin{tabular}[c]{@{}l@{}} $\mathbb{S} = \{\mathbb{S}_i\}_{0 \leq i \leq n}$ where $\mathbb{S}_i$ is defined in (\ref{S_i}), a set including \\observations, as defined in Section \ref{Sec_Info_1}. $\mathbb{S}^{l}$, $\mathbb{S}^{h}$, and $\mathbb{S}^{s}$ are \\observations used for occupancy maps $\mathbb{M}^{l}$, $\mathbb{M}^{h}$, and $\mathbb{M}^{s}$, \\respectively. \end{tabular}\\ \hline

     $s$  &\begin{tabular}[c]{@{}l@{}} Resolution of the occupancy map, which indicates the distance\\ between two nearby cell vertices. $s^{l}$ and $s^{h}$ represent the \\resolutions of low-resolution map and high-resolution map, \\respectively. $s^L$ and $s^G$ denote the resolutions of local maps and \\the global map, respectively, as described in Section \ref{Sec_submap}. \end{tabular}\\ \hline
    
     $\mathbb{O}$ & \begin{tabular}[c]{@{}l@{}}Set including all odometry inputs, as defined in Section \ref{sec_odometry}. \end{tabular}\\ \hline

    $r$  & Ratio between resolutions of two stages, $r = {s^{l}}/{s^h}$.  \\ \hline

    $\mathbf{m}$ & \begin{tabular}[c]{@{}l@{}} Discrete coordinate of a cell vertex, detailed explanation in the \\second paragraph in Section \ref{sec_discrete_occupancy}.\end{tabular} \\ \hline

    $\mathbf{p}$ & \begin{tabular}[c]{@{}l@{}}Continuous coordinate of a point, see the second paragraph in \\Section \ref{sec_relationship}.\end{tabular} \\
        
 		\hline
 		\end{tabular}
 		}

 \end{table}

\section{Problem Formulation}\label{sec_formulation}
Our approach considers the joint optimization of the robot poses and the occupancy map using information from 2D laser observations (and odometry). In this section, we will explain how the observations from the laser can be linked to the robot poses and the occupancy map to formulate the NLLS problem. 


\subsection{Notation}
Throughout this paper, unless otherwise noted, we use specific typographical conventions: typefaces denote sets, bold uppercase letters represent matrices, bold lowercase letters indicate vectors, and regular (unbolded) lowercase letters signify scalars. Key notations used in this paper are summarized in Table \ref{tab_notation}, while others are introduced within the text as needed.

\subsection{Occupancy Map Representation and State in Optimization} \label{sec_discrete_occupancy}
Suppose the environment is discretized into $c_w\times c_h$ grid cells. We use $\mathbf{m}_{wh}=[w,h]^\top~(0 \leq w \leq c_w, 0 \leq h \leq c_h)$ to represent the coordinate of a discrete cell vertex in the map. The occupancy value at the cell vertex $\mathbf{m}_{wh}$, denoted as $M(\mathbf{m}_{wh})$, is defined using evidence, which is the natural logarithm of odds (the ratio between the probability of being occupied and the probability of being free) \cite{martin1996robot,hornung2013octomap,ProbabilisticRobotics}. 
The occupancy values of all $(c_w+1) \times (c_h+1)$ cell vertices consist of the discrete occupancy map $\mathbb{M}=\{M(\mathbf{m}_{wh})\}_{0 \leq w \leq c_w, 0 \leq h \leq c_h}$.

\begin{figure}[t]
\centering
\includegraphics[width=0.48\textwidth]{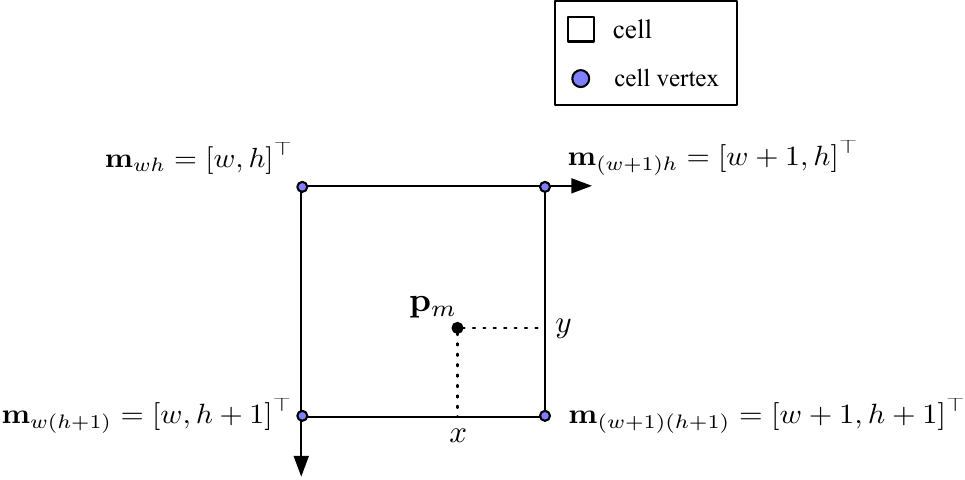}
\caption{\label{fig_interpolation} Parameterizing the entire map by bilinear interpolation of discrete map $\mathbb{M}$.}
\end{figure}

To represent the entire environment using a finite number of parameters, we describe the occupancy value at an arbitrary position $\mathbf{p}_m=[x,y]^{\top}$ on the map using bilinear interpolation of the occupancy values at its four surrounding cell vertices: $\mathbf{m}_{wh}, \mathbf{m}_{({w+1})h}, \mathbf{m}_{w({h+1})}, \mathbf{m}_{({w+1})({h+1})}$, as shown in Fig. \ref{fig_interpolation}, i.e.,

\begin{equation}
	M(\mathbf{p}_{m})= \begin{bmatrix}
a_1b_1,a_0b_1,a_1b_0,a_0b_0
\end{bmatrix}\left[
\begin{aligned}\label{eq_interp}
&M(\mathbf{m}_{wh})\\&M(\mathbf{m}_{(w+1)h})\\&M(\mathbf{m}_{w(h+1)})\\&M(\mathbf{m}_{(w+1)(h+1)})
\end{aligned}\right] 
\end{equation}
in which 
\begin{equation}
\begin{aligned}
	a_0 &= x - w\\
	a_1 &= w+1 - x\\
	b_0 &= y - h\\
	b_1 &= h+1 - y .\\
\end{aligned} 
\end{equation}

Our method jointly optimizes robot poses and the occupancy map, combining them into the state vector of the proposed optimization problem. Using bilinear interpolation with the discrete occupancy map $\mathbb{M}$, estimating the entire map is equivalent to estimating $\mathbb{M}$. Thus, the map component of the state vector can be expressed as

\begin{equation}
    \mathbf{x}^M =\left[M(\mathbf{m}_{00}),\cdots,M(\mathbf{m}_{c_wc_h}) \right]^\top. \label{eq_map_state}
\end{equation}

We define the $n+1$ robot poses as \rule{0pt}{1em}$\{\mathbf{x}^P_i \triangleq [\mathbf{t}_i^\top,\theta_i]^\top\}_{0 \leq i \leq n}$, where $\mathbf{t}_i$ is the $x$-$y$ position of the robot and $\theta_i$ is the orientation with the corresponding rotation matrix $\mathbf{R}_i=\begin{bmatrix}
\cos(\theta_i), \sin(\theta_i)\\ -\sin(\theta_i), \cos(\theta_i)
\end{bmatrix}$. As in most of the SLAM problem formulations, we assume the first robot pose defines the coordinate system, $\mathbf{x}^P_0 \triangleq [0,0,0]^\top$, so only the other $n$ robot poses are variables that need to be estimated, thus the pose component of the state vector is represented as
\begin{equation}
    \mathbf{x}^P = \left[ (\mathbf{x}^P_1)^\top, \cdots, (\mathbf{x}^P_n)^\top \right]^\top.
\end{equation}

Accordingly, the state vector of the proposed optimization problem is
\begin{equation}
    \mathbf{x} = \left[{(\mathbf{x}^P)}^\top,{(\mathbf{x}^M)}^\top \right]^\top. \label{eq_pose_state}
\end{equation}

In our method, the occupancy map $\mathbb{M}$ is initialized by the Bayesian occupancy mapping method \cite{ProbabilisticRobotics} with initially estimated poses (derived from odometry or scan matching) and updated throughout the optimization process.

\subsection {Scan Points Sampling Strategy}\label{Sec_Info_1} 


 \begin{figure}[tbp]
\centering 
\subfigure[Equidistant Sampling Strategy] {\label{fig_sampling_strategy}
\includegraphics[width=0.23\textwidth]{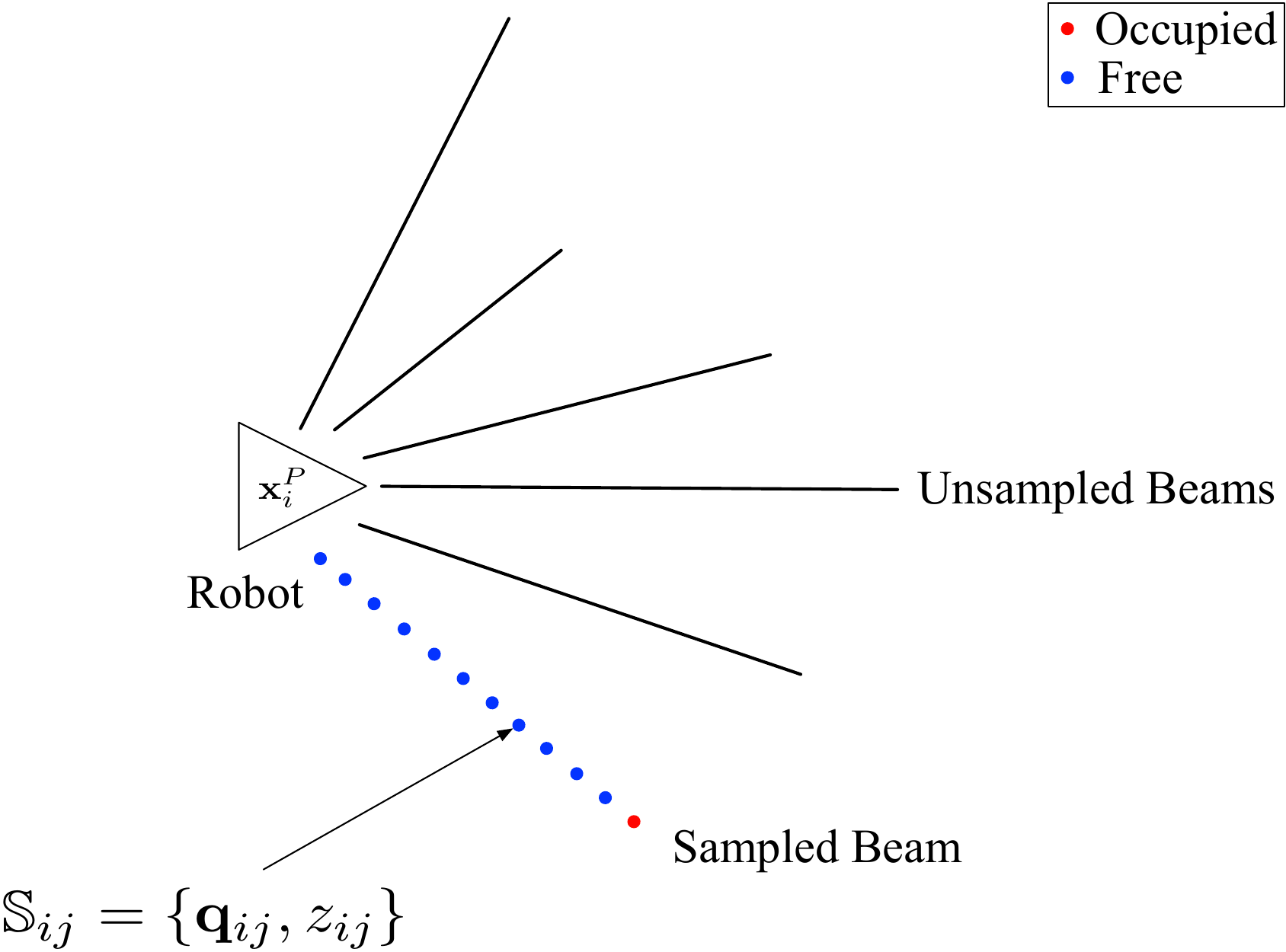}}
\subfigure[Observation Points in One Scan] {\label{fig_scan}
\includegraphics[width=0.23\textwidth]{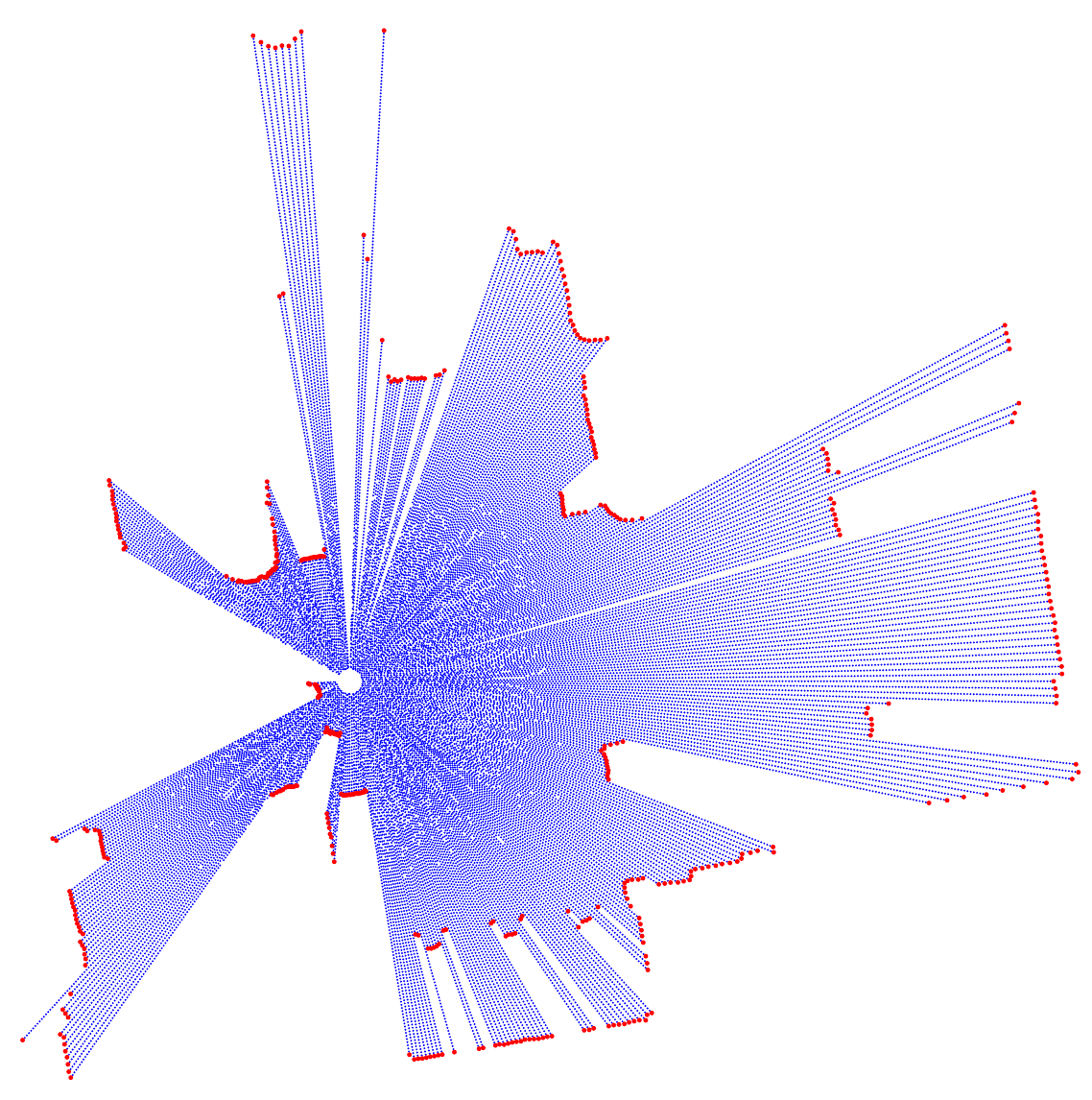}}
\caption{Sampling strategy for generating observations from a laser scan: (a) Equidistant sampling on a beam, with red indicating occupied and blue indicating free states. The distance between two consecutive points is the resolution $s$. (b) All sampled observation points at a given time step.}
\label{fig_scan_sampling}
\end{figure}

We now introduce our sampling strategy for generating observations from laser scans, which are used in our NLLS formulation. 

Each scan data consists of a number of beams. On each beam, the endpoint indicates the presence of an obstacle, while the other points before the endpoint indicate the absence of obstacles. Here, we sample each beam using a fixed resolution $s$ to get the observations, as shown in Fig. \ref{fig_sampling_strategy}. Specifically, $\mathbf{q}_{ij}=[x_{q_{ij}},y_{q_{ij}}]^\top$ denotes the position of $j$th sampling point at time step $i$ in the local robot/laser coordinate frame and
\begin{equation}
z_{ij} = \ln \frac{p(\mathbf{q}_{ij} \in occ)}{1-p(\mathbf{q}_{ij} \in occ)} \label{eq_occ_obs}
\end{equation}denotes the corresponding occupancy value. In the same way as the occupancy map representation described in Section \ref{sec_discrete_occupancy}, we also use the evidence to represent the occupancy value here. In our implementation, following \cite{hornung2013octomap,ProbabilisticRobotics}, we use $p(\mathbf{q}_{ij} \in occ) = 0.7$ for an occupied point (red in Fig. \ref{fig_sampling_strategy}), and use $p(\mathbf{q}_{ij} \in occ) = 0.4$ for a free point (blue in Fig. \ref{fig_sampling_strategy}). Fig. \ref{fig_scan} shows an example of all sampled points in one scan.
By constant equidistant sampling of all the beams for the scan collected at time step $i$, a sampling point set
\begin{equation}
\mathbb{S}_i=\{ \mathbb{S}_{ij} \triangleq \{\mathbf{q}_{ij},z_{ij}\}\}_{1 \leq j \leq k_i}
\label{S_i}
\end{equation}
can be obtained. It should be noted that since the total length of all the beams at different time step $i$ is different, the number of sampling points $k_i$ obtained by the equidistant sampling strategy varies for different time step $i$.

Suppose there are $n+1$ laser scans collected from robot poses $0$ to $n$, $\mathbb{S}=\{\mathbb{S}_i\}_{0\leq i \leq n}$ is the available observation information collected at all different robot poses using our sampling strategy and will be used as observations in our NLLS formulation.

\subsection{Relationship Between Observations and Occupancy Map}\label{sec_relationship}

In Section \ref{sec_discrete_occupancy}, we defined the discrete occupancy map $\mathbb{M}$ as part of the state vector in our optimization problem. Section \ref{Sec_Info_1} detailed the observation generation process. In this section, we explain how the relationship between observations and the occupancy map is established through robot poses, forming the basis of our joint optimization problem.

\subsubsection{Local to Global Projection}
First, the $j$th scan point at time step $i$ can be projected to the occupancy map using the robot pose $\mathbf{x}^P_i$, and the projected position on the occupancy map can be calculated by 

\begin{equation}
	\mathbf{p}_{ij}
=\frac{\mathbf{R}_i^\top \mathbf{q}_{ij}+\mathbf{t}_i}{s} \label{P-project}
\end{equation}
where $s$ is the resolution of the vertices in the occupancy map $\mathbb{M}$ (the distance between two adjacent cell vertices represents $s$ meters in the real world). Here, we use the same resolution as that used in generating observations from laser scans in Section \ref{Sec_Info_1}. Then, the occupancy value at the projected point $\mathbf{p}_{ij}$ can be obtained using (\ref{eq_interp}), expressed as $M(\mathbf{p}_{ij})$.

\subsubsection{Relationship Between Sampling Points and Occupancy Map w.r.t. Occupancy Values}
As outlined in Sections \ref{sec_discrete_occupancy} and \ref{Sec_Info_1}, evidence is used to define the occupancy value, where multiple observations of the same cell result in the occupancy values from individual observations being cumulatively added to the cell's total occupancy value \cite{hornung2013octomap}. If the robot's poses are accurate and repeated observations of the same cell consistently indicate the same occupancy state, the cell's occupancy value becomes the product of the occupancy value of each observation and the number of times the cell is observed. For a unique coordinate $\mathbf{p}_{ij}$ in (\ref{P-project}), if both the robot pose $\mathbf{x}^P_i$ and the occupancy map $\mathbb{M}$ are accurate, the occupancy value $z_{ij}$ of its associated sampling point $\mathbb{S}_{ij}$, should closely approximate the occupancy value at $\mathbf{p}_{ij}$, $M(\mathbf{p}_{ij})$, divided by the number of times $\mathbf{p}_{ij}$ is ``observed", $N(\mathbf{p}_{ij})$.

Thus, if the number of times the point $\mathbf{p}_{ij}$ is ``observed" can be calculated, the relationship between the observations and the state vector (occupancy map and robot poses), can be determined.

\subsubsection{Hit Map and Hit Number Lookup}\label{sec_hit}

We now explain how $N({\mathbf{p}_{ij}})$ can be calculated. To quickly query the number of times an arbitrary point is ``observed", we need to count the number of times all cell vertices have been observed to form the discrete hit map $\mathbb{N}$ associated with the occupancy map $\mathbb{M}$. 

When a sampling scan point is projected into a coordinate by a given robot pose, this coordinate is considered to have been observed once, and then we distribute the hit number ``1" of the coordinate to the discretized cell vertices. Since the occupancy value $M(\mathbf{p}_{ij})$ is derived by bilinear interpolation of occupancy values of discrete cell vertices in (\ref{eq_interp}), in order to maintain the correspondence between the hit number and the occupancy values, we distribute this ``1" hit to the four surrounding cell vertices by inverse bilinear interpolation. For example, if a sampling point is projected into the center of a cell, then each of the 4 nearby cell vertices gets a hit number of 0.25. In addition, the hit number also accumulates with multiple observations of the same cell vertex, i.e.,
\begin{equation}
\left[ N(\mathbf{m}_{00}),\cdots,N(\mathbf{m}_{c_wc_h}) \right] 
= \sum_{i=0}^n \sum_{j=1}^{k_i} H(\mathbf{p}_{ij})
\label{eq_NP}
\end{equation}
where $H(\cdot)$ is the inverse process of bilinear interpolation. The hit number at all these discrete cell vertices consists of discrete hit map $\mathbb{N}=\{N(\mathbf{m}_{wh})\}_{0 \leq w \leq c_w, 0 \leq h \leq c_h}$.

After the discrete hit map $\mathbb{N}$ is obtained, the equivalent hit multiplier $N(\mathbf{p}_{ij})$ (representing the number of times $\mathbf{p}_{ij}$ is ``observed") for an arbitrary continuous point $\mathbf{p}_{ij}$ can be easily obtained using bilinear interpolation, similar to (\ref{eq_interp}). 

\subsection{The NLLS Formulation} 
We now formulate the NLLS problem to jointly optimize the robot poses and the occupancy map. The objective function of the NLLS problem is defined as
\begin{equation}
f(\mathbf{x})=w_Z f^Z(\mathbf{x})+w_O f^O(\mathbf{x})+w_S f^S(\mathbf{x}). 	\label{eq_objective_func}
\end{equation}
The objective function consists of the observation term $f^Z(\mathbf{x})$, the smoothing term $f^S(\mathbf{x})$, and the odometry term $f^O(\mathbf{x})$. $w_Z$, $w_S$ and $w_O$ are their corresponding weights, and we set $w_O = 0$ if there is no odometry information. We now explain the three terms one by one.

\subsubsection{Observation Term $f^Z(\mathbf{x})$}
Based on the relationship between the observations and the occupancy map w.r.t. occupancy values described in \ref{sec_relationship}, we can formulate the observation term as follows. 

Given the observation information $\mathbb{S}$ in (\ref{S_i}), the observation term in the objective function (\ref{eq_objective_func}) is formulated as
\begin{equation}
	f^Z(\mathbf{x}) =
	\sum_{i=0}^n \sum_{j=1}^{k_i}  \left\|z_{ij} - F_{ij}^Z(\mathbf{x})\right\|^2, 
\label{obs-term}
\end{equation}
where
\begin{equation}
	F_{ij}^Z(\mathbf{x})  = \frac{M(\mathbf{p}_{ij})}{N({\mathbf{p}_{ij}})}.\\ \label{eq_MN}
\end{equation}
Here, $\mathbf{p}_{ij}$ represents a coordinate in the map where the $j$th sampling scan point at time step $i$ is projected using the robot pose $\mathbf{x}^P_i$, as calculated by (\ref{P-project}). $N(\mathbf{p}_{ij})$ denotes the equivalent hit multiplier at $\mathbf{p}_{ij}$, as detailed in Section \ref{sec_hit}. 

In (\ref{obs-term}), we suppose the errors of occupancy values of different sampled points in the observations $\mathbb{S}$ are independent and have the same uncertainty. Therefore, the weights on all terms are the same, which is equivalent to setting all the weights as $1$. Thus, we use norms instead of weighted norms in equation (\ref{obs-term}).

\subsubsection{Odometry Term $f^O(\mathbf{x})$}\label{sec_odometry}

The odometry information $\mathbb{O} = \{\mathbf{o}_i\}_{1 \leq i \leq n}$ might be available. We assume the odometry input is the relative pose between two consecutive steps. 
The odometry from robot pose $\mathbf{x}^P_{i-1}$ to pose $\mathbf{x}^P_{i}$ is expressed as
\begin{equation} 
\mathbf{o}_i=\left[ (\mathbf{o}_i^t)^\top,o_i^\theta \right]^\top~~(1 \leq i \leq n)
\label{O_i}
\end{equation}
where $\mathbf{o}_i^t$ is the translation part and $o_i^\theta$ is the rotation angle part of the odometry. The odometry term can be formulated as
\begin{equation}
\begin{aligned}
f^O(\mathbf{x})&=\sum_{i=1}^n \left\|\mathbf{o}_i -
F_i^O(\mathbf{x})
\right\|^2_{\mathbf{\Sigma}^{-1}_{O_i}}
\\&=\sum_{i=1}^n\left\|
\begin{bmatrix}
\mathbf{o}_i^t-\mathbf{R}_{i-1}\left(\mathbf{t}_i - \mathbf{t}_{i-1} \right)\\
\wrap\left({o}_i^\theta- \theta_i + \theta_{i-1}\right)
\end{bmatrix}
\right\|^2_{\mathbf{\Sigma}^{-1}_{O_i}}  \label{eq_odometry_term}
\end{aligned}
\end{equation}
in which $\mathbf{\Sigma}_{O_i}$ is the covariance matrix representing the uncertainty of $\mathbf{o}_i$, and $\wrap(\cdot)$ wraparounds the rotation angle to $(-\pi,\pi]$.
\subsubsection{Smoothing Term $f^S(\mathbf{x})$}

It can be easily found out that minimizing the objective function with only the observation term (and the odometry term) is not easy since there are a large number of local minima. Especially when the initial robot poses are far away from the global minimum, it is very difficult for an optimizer to converge to the correct solution. 

In order to enlarge the region of attraction and develop an algorithm that is robust to initial values, we introduce a smoothing term. The smoothing term requires the occupancy values of nearby cell vertices to be close to each other thus resulting in the occupancy map being smoother for derivative calculation. In our case, based on the derivative calculation method we use (see Appendix \ref{Sec_J_P}), we penalize the difference between the occupancy value of each cell vertex and the occupancy values of the two neighboring cell vertices to its right and below, i.e.,
\begin{equation}
\begin{aligned}
f^S(\mathbf{x})
& =\left\|F^S(\mathbf{x}) \right\|^2\\
& = \sum_{w=0}^{c_w-1} \sum_{h=0}^{c_h-1}  \left\|\begin{bmatrix} M(\mathbf{m}_{wh})-M(\mathbf{m}_{{(w+1)}h})\\
M(\mathbf{m}_{wh})-M(\mathbf{m}_{{w}{(h+1)}})
\end{bmatrix} \right\|^2 \\
& + \sum_{h=0}^{c_h-1}  \left\| M(\mathbf{m}_{c_wh})-M(\mathbf{m}_{{c_w}{(h+1)}})\right\|^2 \\
& + \sum_{w=0}^{c_w-1}  \left\| M(\mathbf{m}_{wc_h})-M(\mathbf{m}_{{(w+1)}{c_h}})\right\|^2,
\end{aligned} \label{eq_smoothing_term}
\end{equation} where the second and third terms are used to handle cell vertices located in the bottom row and the rightmost column. It should be noted that $F^S(\mathbf{x})$ is a linear function of $\mathbf{x}^M$ in the state. The coefficient matrix is constant and can be calculated prior to the optimization. For more details, please refer to Appendix \ref{Sec_J_S}.

\section{Iterative Solution to the NLLS Formulation}\label{Sec_Algorithm_1}
In Section \ref{sec_formulation}, we introduced our NLLS formulation for the joint poses and occupancy map optimization problem. In this section, we provide the details of a Gauss-Newton based algorithm for solving the NLLS problem.

\begin{algorithm}[t]
\small
\caption{Our Joint Poses and Occupancy Map Optimization Algorithm}\label{alg_1}
\SetKwInput{KwInput}{Input}                
\SetKwInput{KwOutput}{Output}              
\SetKwInput{KwParam}{Params}
\SetAlgoLined
\DontPrintSemicolon
\SetKw{Return}{End Function}
  \KwParam{Threshold $\tau_k$, $\tau_{\Delta}$, weight matrix $\mathbf{W}$, resolution $s$}
  \KwInput{Observations $\mathbb{S}$, odometry $\mathbb{O}$, and initial poses $\mathbf{x}^P(0)$}
  \KwOutput{Optimized poses $\hat{\mathbf{x}}^P$ and optimized map $\hat{\mathbf{x}}^M$}
\SetKwFunction{FuncFirstStage}{FirstStage}

\SetKwProg{Fn}{Function}{:}{}
\Fn{\FuncFirstStage{$\mathbf{x}^P(0)$, $\mathbb{S}$, $\mathbb{O}$, $\tau_k$, $\tau_{\Delta}$, $s$, $\mathbf{W}$}}
{
Initialize $\mathbf{x}^M(0)$ and $\mathbb{N}(0)$ using $\mathbf{x}^P(0)$ and $\mathbb{S}$ \;

Pre-calculate smoothing term coefficient $\mathbf{A}$ using (\ref{eq_A})\;

\SetKwFunction{FuncGN}{OccupancyGN}
\SetKwFunction{FuncReturn}{return}

\SetKwProg{Fn}{Function}{:}{}
\For {$k=0$; $k <= \tau_k \; \& \; \| \mathbf{\Delta}(k) \|^2 >= \tau_{{\Delta}}$; $k++$}{
\Fn{\FuncGN{$\mathbf{x}^M(k)$, $\mathbb{N}(k)$, $\mathbf{x}^P(k)$, $\mathbb{S}$, $\mathbb{O}$, $\mathbf{A}$, $\mathbf{W}$}}{
Calculate gradient $\mathbf{\nabla} \mathbf{x}^M(k)$ of $\mathbf{x}^M(k)$

Calculate $\mathbf{J}$, as described in appendices

Evaluate $F(\mathbf{x})$ at $\mathbf{x}^P(k)$ and $\mathbf{x}^M(k)$

Solve $\mathbf{J}^\top \mathbf{W} \mathbf{J} \mathbf{\Delta}(k) =-\mathbf{J}^\top \mathbf{W} F(\mathbf{x})$, where $\mathbf{\Delta}(k) = {\left[{\mathbf{\Delta}^P(k)}^\top,{\mathbf{\Delta}^M(k)}^\top\right]}^\top$

Update $\mathbf{x}^P(k+1)=\mathbf{x}^P(k) + \mathbf{\Delta}^P(k)$ and $\mathbf{x}^M(k+1)=\mathbf{x}^M(k)+\mathbf{\Delta}^M(k)$

Recalculate $\mathbb{N}(k+1)$ using $\mathbf{x}^P(k+1)$ and $\mathbb{S}$

\FuncReturn{$\mathbf{x}^P(k+1)$, $\mathbf{x}^M(k+1)$}
}
\Return
}
$\hat{\mathbf{x}}^P \Leftarrow \mathbf{x}^P(k)$, $\hat{\mathbf{x}}^M \Leftarrow \mathbf{x}^M(k)$

\FuncReturn{$\hat{\mathbf{x}}^P$, $\hat{\mathbf{x}}^M$}
}

\Return
\end{algorithm}

In the equation below, we assume the odometry inputs are available. Let
\begin{equation}
\begin{aligned}
F(\mathbf{x}) = [&\cdots,z_{ij}-F_{ij}^Z(\mathbf{x}),\cdots,{(\mathbf{o}_i-F_i^O(\mathbf{x}))}^\top,\\
&\cdots,{F^S(\mathbf{x})}^\top]^\top\\
\mathbf{W} = \;\; &\diag(\cdots,w_Z,\cdots,w_O \mathbf{\Sigma}^{-1}_{O_i}, \cdots,w_S, \cdots)\\
\end{aligned}
\end{equation}
combine all the error functions and the weights of the three terms in (\ref{eq_objective_func}). Then, the NLLS problem in (\ref{eq_objective_func}) seeks $\mathbf{x}$ such that
\begin{equation}\label{Least Squares}
f(\mathbf{x})=\|F(\mathbf{x})\|^2_{\mathbf{W}} =
{F(\mathbf{x})}^\top \mathbf{W}
F(\mathbf{x})
\end{equation}
is minimized.

A solution to (\ref{Least Squares}) can be obtained iteratively by starting with an initial guess $\mathbf{x}(0)$ and updating with $\mathbf{x}(k+1) = \mathbf{x}(k) + \mathbf{\Delta}(k)$. \rule{0pt}{1em}The update vector $\mathbf{\Delta} (k) = [{\mathbf{\Delta}^P(k)}^\top,{\mathbf{\Delta}^M(k)}^\top]^\top$ is the solution to
\begin{equation}\label{Gauss-Newton}
\mathbf{J}^\top \mathbf{W} \mathbf{J} \mathbf{\Delta} (k) = -\mathbf{J}^\top \mathbf{W} F(\mathbf{x}(k))
\end{equation}
where $\mathbf{J}$ is the linear mapping represented by the Jacobian matrix
$\partial F / \partial \mathbf{x}$ evaluated at $\mathbf{x}(k)$.

The iterative method for solving the proposed NLLS problem is shown in Algorithm \ref{alg_1}, in which $\tau_k$ and $\tau_{\Delta}$ represent the thresholds of iteration number $k$ and the incremental vector $\mathbf{\Delta}$. Unlike the standard Gauss-Newton iterative method, the hit map needs to be additionally recalculated after updating the poses in each iteration. With this approach, the implicit data association is established at each iteration and updated during the optimization.

Since the robot poses and the occupancy map are optimized simultaneously, the Jacobian $\mathbf{J}$ in (\ref{Gauss-Newton}) is very important and quite different from those used in the traditional SLAM algorithms. More details of the Jacobians are described in appendices.

\section{Multi-resolution Joint Optimization Strategy} \label{Sec_multi}
Algorithm \ref{alg_1} provides a solution to our NLLS problem (\ref{eq_objective_func}) to jointly optimize the poses and the occupancy map. However, directly using Algorithm \ref{alg_1} with the high-resolution map is time-consuming and requires an accurate initial value of robot poses \cite{Zhao-RSS-22}, which is challenging to obtain. To overcome these limitations, we propose a multi-resolution joint optimization strategy in this section.

\subsection{Discussion on Map Resolution in Optimization }

The resolution of the occupancy map has a significant impact on the optimization results since $\mathbf{x}^M$ is part of the state vector in our NLLS formulation (\ref{eq_objective_func}). 


A high-resolution map enables precise relationships between observations and occupancy values of projected points. However, it results in a dramatic increase in the optimization problem's size, raising computational costs. Additionally, in a high-resolution map, occupancy values in adjacent cell vertices may exhibit sharper variations compared to those in a low-resolution map, leading to noisy gradients when poor initial robot poses are used. Even with the introduction of the smoothing term, the use of a high-resolution map may cause poor convergence of Algorithm \ref{alg_1}.

A low-resolution map provides advantages in faster computation and reduced memory usage. Moreover, gradients are less sensitive to pose accuracy. With our occupancy map representation and smoothing term, these advantages enable the algorithm to quickly converge to a reasonable solution, even with poor initial robot poses. However, low resolution may cause inaccurate links between observations and occupancy values near boundaries, preventing the optimization from achieving greater accuracy.

To combine the advantages of different resolution map representations, we propose a multi-resolution strategy to optimize the occupancy values of different resolution cell vertices together with robot poses at various stages. Unlike the conventional coarse-to-fine scheme, in the second stage of our strategy, we use the selected high-resolution map that only includes high-resolution cell vertices possibly in need of further optimization instead of the full high-resolution map. Optimizing only those selected high-resolution cell vertices further improves the efficiency of our algorithm. 
\subsection{Our Multi-resolution Joint Optimization Strategy}

Firstly, we obtain low-resolution observations $\mathbb{S}^{l} = \{\mathbb{S}_i^{l}\}_{0\leq i \leq n}$ by down-sampling from the high-resolution observations $\mathbb{S}^{h} = \{\mathbb{S}_i^{h}\}_{0\leq i \leq n}$, which are obtained by the equal sampling strategy described in Section \ref{Sec_Info_1} with a sampling distance $s^{h}$. Here, we set the map resolution and sampling resolution to be the same. Therefore, the low resolution $s^{l}=r \times s^{h}$, where $r$ is the resolution ratio between the low-resolution map and the high-resolution map. The size of the low-resolution map $\mathbb{M}^{l}$ is $(c_w+1) \times (c_h+1)$.

Initialized by the odometry inputs or scan matching, we perform Algorithm \ref{alg_1} to quickly obtain relatively accurate poses. The state vector in the first stage is ${\mathbf{x}}^{l} = {\left[{\mathbf{x}^P}^\top,{{\mathbf{x}^{lM}}}^\top\right]}^\top $, where $\mathbf{x}^{lM}$ includes all occupancy values at the cell vertices of the low-resolution map $\mathbb{M}^{l}$. In this stage, the hit map, observation information, coefficient matrix, weight matrix, and resolution are represented as $\mathbb{N}^{l},  \mathbb{S}^{l}, \mathbf{A}^{l}$, $\mathbf{W}^{l}$, and $s^{l}$, respectively. 

In the first stage of optimization, the low-resolution occupancy map reduces both the dimension of $\mathbf{x}^{lM}$ and the number of observations in $\mathbb{S}^{l}$. Since the occupancy values at cell vertices change relatively gradually in the low-resolution map, the directions of the map's gradients are closer to the correct ones when the poses are initialized by odometry inputs or scan matching, making it easier for Algorithm \ref{alg_1} to converge to a relatively good result quickly.

After the first stage, we use Algorithm \ref{alg_2} to select the cell vertices that need to be further optimized to compose the selected high-resolution map $\mathbb{M}^{s}$ and find their corresponding observations $\mathbb{S}^{s}$. Details are described in Section \ref{select_index_set}.

\begin{algorithm}[tp]
\small
\caption{Finding the Selected High-resolution Map and Corresponding Observations}\label{alg_2}

\SetKwInput{KwInput}{Input}                
\SetKwInput{KwOutput}{Output}              
\SetKwInput{KwParam}{Params}
\SetKw{Return}{End Function}
\SetAlgoLined
\DontPrintSemicolon
\KwParam{Resolution $s^{h}$, selection distance $d$, convolution kernel size $q$}
  \KwInput{Observations $\mathbb{S}^{h}$, and poses ${{\hat{\mathbf{x}}}}^{\tilde{P}}$ from the first stage using Algorithm \ref{alg_1}}
  \KwOutput{Observations $\mathbb{S}^{s}$ and map part of the state vector in the second stage $\mathbf{x}^{sM}$}
  \SetKwFunction{FuncSel}{Selection}
\SetKwProg{Fn}{Function}{:}{}
\Fn{\FuncSel{${{\hat{\mathbf{x}}}}^{\tilde{P}}$, $\mathbb{S}^{h}$, $s^{h}$, $d$, $q$}}{

Build a full high-resolution map $\mathbb{M}^{h}$ using $\hat{\mathbf{x}}^{\tilde{P}}$ and $\mathbb{S}^{h}$ 

Calculate the binary map $\mathbb{B}$ using $\mathbb{M}^{h}$

Calculate the convoluted map $\mathbb{C}$ with kernel size $q$

Calculate the set $\mathbb{I}^{h}$, which includes the indices of all boundary vertices in $\mathbb{M}^{h}$, using $\mathbb{C}$

Calculate the set $\mathbb{I}^{s}$, which includes the indices of all selected vertices, using $\mathbb{I}^{h}$ and $d$

Define the map part of the state vector in the second stage $\mathbf{x}^{sM}$ and the selected high-resolution map $\mathbb{M}^{s}$ by $\mathbb{I}^{s}$

Find observations $\mathbb{S}^{s}$ for $\mathbb{M}^{s}$ using the set $\mathbb{I}^{s}$, $\mathbb{S}^{h}$ and ${{\hat{\mathbf{x}}}}^{\tilde{P}}$

\FuncReturn{$\mathbb{S}^{s}$, $\mathbf{x}^{sM}$}
}
\Return
\end{algorithm}

In the second stage, the state vector is represented as $\mathbf{x}^{s} = {\left[{{\mathbf{x}}^P}^\top,{\mathbf{x}^{sM})}^\top\right]}^\top $ where $\mathbf{x}^{sM}$ includes all occupancy values at cell vertices of the selected high-resolution map $\mathbb{M}^{s}$. We perform Algorithm \ref{alg_3} using poses obtained from the first stage as initial guesses and observations $\mathbb{S}^{s}$ to refine poses. The NLLS optimization problem in the second stage can be formulated similarly as (\ref{Least Squares}). Additionally, the differences in the Jacobian calculation between Algorithm \ref{alg_1} and Algorithm \ref{alg_3} are described in Appendix \ref{Sec_J_Select}. 

The full multi-resolution joint optimization strategy is outlined in Algorithm \ref{alg_flowchart}.

\begin{algorithm}[t]
\small
\caption{The Algorithm for the Second Stage of the Multi-resolution Joint Optimization Strategy}\label{alg_3}
\SetKwInput{KwParam}{Params}
\SetKwInput{KwInput}{Input}                
\SetKwInput{KwOutput}{Output}              
\SetAlgoLined
\DontPrintSemicolon
\SetKw{Return}{End Function}

  \KwParam{Threshold $\tau_k^{s}$, $\tau_{\Delta}^{s}$, weight matrix $\mathbf{W}^{s}$, resolution $s^{h}$}
  \KwInput{Observations $\mathbb{S}^{s}$, odometry $\mathbb{O}$, and poses ${{\hat{\mathbf{x}}}}^{\tilde{P}}$ from the first stage using Algorithm \ref{alg_1}}
  \KwOutput{Optimal poses $\hat{\mathbf{x}}^P$ and map $\hat{\mathbf{x}}^{sM}$}
\SetKwFunction{FuncSecondStage}{SecondStage}

$\mathbf{x}^P(0) \Leftarrow {{\hat{\mathbf{x}}}}^{\tilde{P}}$

\SetKwProg{Fn}{Function}{:}{}
\Fn{\FuncSecondStage{$\mathbf{x}^P(0)$, $\mathbb{S}^{s}$, $\mathbb{O}$, $\tau_k^{s}$, $\tau_{\Delta}^{s}$, $s^{h}$, $\mathbf{W}^{s}$}}
{

Initialize $\mathbf{x}^{sM}(0)$ and $\mathbb{N}^{s}(0)$ using $\mathbb{S}^{s}$ and $\mathbf{x}^P(0)$

Pre-calculate smoothing term coefficient matrix $\mathbf{A}^{s}$

\For {$k=0$; $k <= \tau_k^{s} \; \& \; \| \mathbf{\Delta}(k) \|^2 >= \tau_{\Delta}^{s}$; $k++$}{

$\mathbf{x}^P(k+1)$, $\mathbf{x}^{sM}(k+1)$
$\leftarrow$  \FuncGN{$\mathbf{x}^{sM}(k)$, {$\mathbb{N}^{s}(k)$, $\mathbf{x}^P(k)$, $\mathbb{S}^{s}$, $\mathbb{O}$, $\mathbf{A}^{s}$, $\mathbf{W}^{s}$}}
}

$\hat{\mathbf{x}}^P \Leftarrow \mathbf{x}^P(k)$, $\hat{\mathbf{x}}^{sM} \Leftarrow \mathbf{x}^{sM}(k)$

\FuncReturn{$\hat{\mathbf{x}}^P$, $\hat{\mathbf{x}}^{sM}$}

}
\Return
\end{algorithm}

\begin{algorithm}[t]
\small
\caption{Our Multi-resolution Joint Optimization Strategy}\label{alg_flowchart}
\SetKwInput{KwParam}{Params}
\SetKwInput{KwInput}{Input}                
\SetKwInput{KwOutput}{Output}              

\SetAlgoLined
\DontPrintSemicolon

\SetKwFunction{FuncDown}{DownSampling}
\SetKwFunction{FuncInitPose}{InitializePose}
\SetKwFunction{FuncSM}{ScanMatching}
  \KwParam{Threshold $\tau_k^{l}$, $\tau_{\Delta}^{l}$, $\tau_k^{s}$, $\tau_{\Delta}^{s}$, weight matrix $\mathbf{W}^{l}$, $\mathbf{W}^{s}$, ratio of resolutions $r$, resolution $s$, selection distance $d$, convolution kernel size $q$}
  \KwInput{Observations $\mathbb{S}^{h}$, odometry $\mathbb{O}$}
  \KwOutput{Optimal poses $\hat{\mathbf{x}}^P$ and map $\hat{\mathbf{x}}^{sM}$}

$\mathbb{S}^{l}$ $\leftarrow$ \FuncDown{$\mathbb{S}^{h}$, $r$}

\uIf {$w_O \neq  0$}
{
$\mathbf{x}^P(0)$ $\leftarrow$ \FuncInitPose{$\mathbb{O}$}
}
\Else
{
$\mathbf{x}^P(0)$ $\leftarrow$ \FuncSM{$\mathbb{S}^{l}$}
}

${{\hat{\mathbf{x}}}}^{\tilde{P}}$, $\hat{\mathbf{x}}^{lM}$ $\leftarrow$ \FuncFirstStage{$\mathbf{x}^P(0)$, $\mathbb{S}^{l}$, $\mathbb{O}$, $\tau_k^{l}$, $\tau_{\Delta}^{l}$, $s^{l}$, $\mathbf{W}^{l}$}

$\mathbb{S}^{s}$, $\mathbf{x}^{sM}$ $\leftarrow$ \FuncSel{${{\hat{\mathbf{x}}}}^{\tilde{P}}$, $\mathbb{S}^{h}$, $s^{h}$, $d$, $q$}

${\hat{\mathbf{x}}^P}$, $\hat{\mathbf{x}}^{sM}$ $\leftarrow$
\FuncSecondStage{${{\hat{\mathbf{x}}}}^{\tilde{P}}$, $\mathbb{S}^{s}$,$\mathbb{O}$, $\tau_k^{s}$, $\tau_{\Delta}^{s}$, $s^{h}$, $\mathbf{W}^{s}$}

\end{algorithm}

\begin{figure}[t]
\centering 
\subfigure[Full High-resolution Map]{ 
\includegraphics[width=0.23\textwidth]{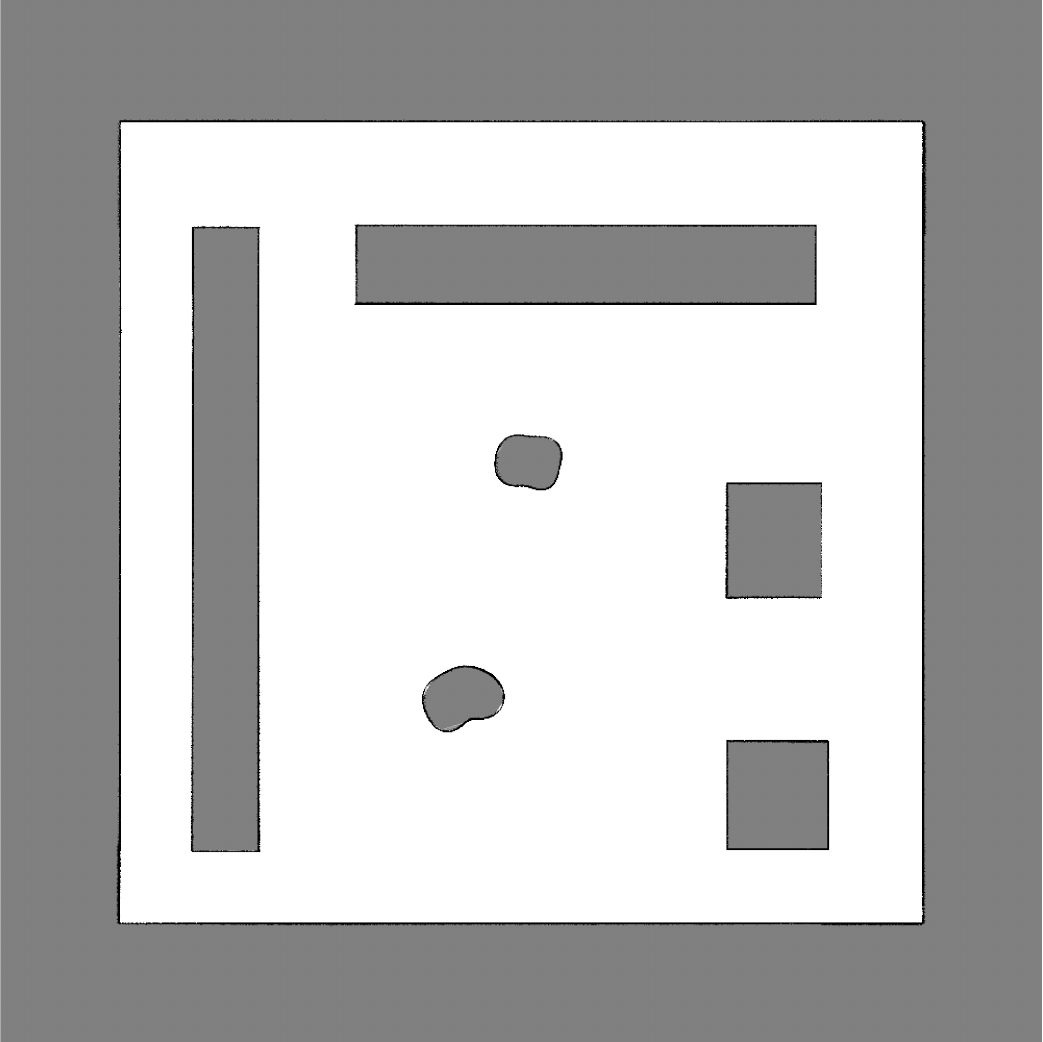}}
\subfigure[Selected High-resolution Map (In White and Black)]{\label{fig_select_example_b}
\includegraphics[width=0.23\textwidth]{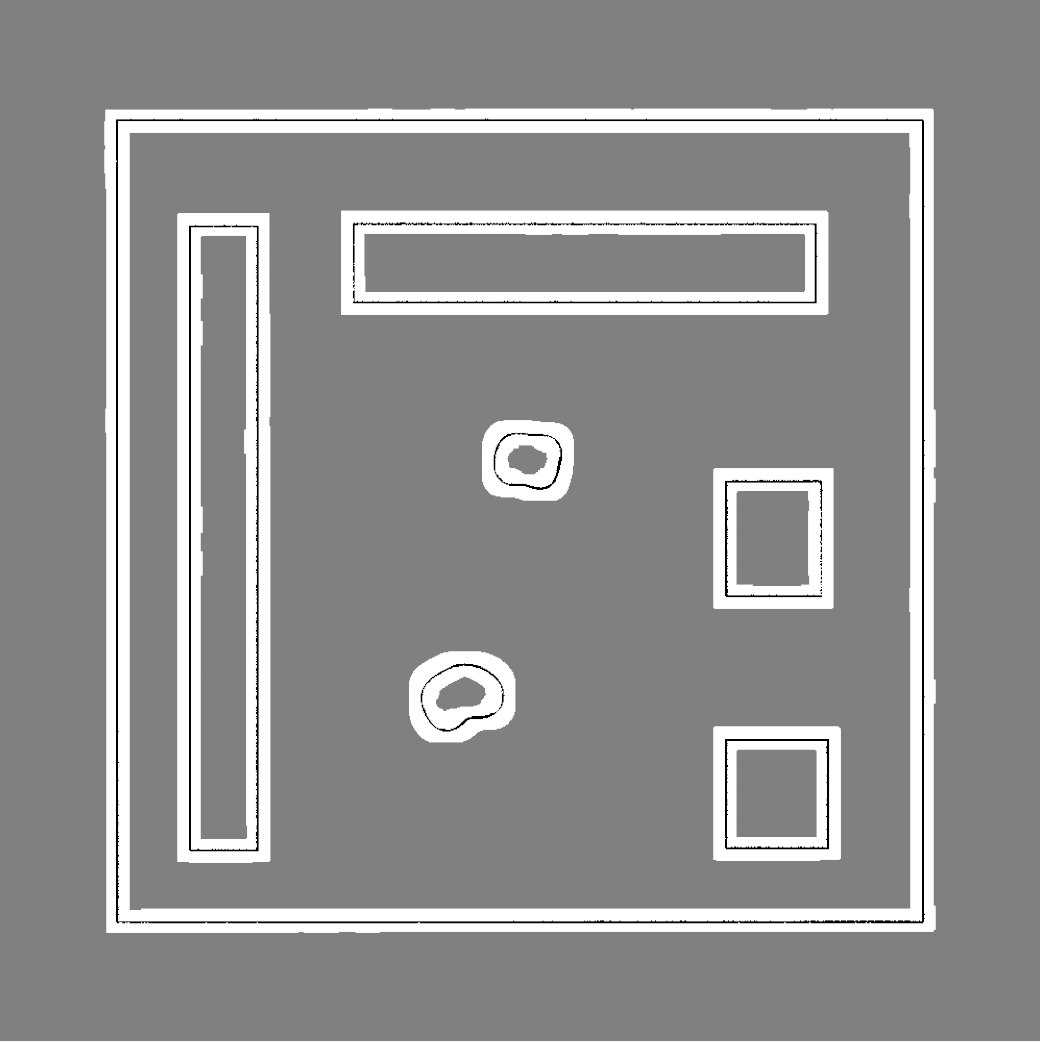}}
\caption{An example of the selected high-resolution map from a full high-resolution map in a simulation dataset. (a) The full high-resolution map generated using poses from the first-stage optimization and scans, forming the basis for selection. (b) The recolored selected high-resolution map: gray marks dropped (stable) areas, white and black denote selected areas, with black highlighting obstacle boundaries.}
\label{fig_select_example}
\end{figure}

\subsection{Selected High-resolution Map and Observations}\label{select_index_set}

After the first stage optimization using the low-resolution map $\mathbb{M}^{l}$, the robot poses $\hat{\mathbf{x}}^{\tilde{P}}$ become relatively accurate. Subsequently, the full high-resolution map $\mathbb{M}^{h}$, with dimensions $(r*c_w+1) \times (r*c_h+1)$, is built using the Bayesian occupancy mapping method \cite{ProbabilisticRobotics}, based on observations $\mathbb{S}^{h}$ and poses ${{\hat{\mathbf{x}}}}^{\tilde{P}}$. In this case, most cell vertices of $\mathbb{M}^{h}$ are considered stable in terms of occupancy state. Semantically, these stable cell vertices have the same occupancy state as the surrounding cell vertices (typically free or unknown cells). This characteristic leads to map gradients near zero at these stable cell vertices. In contrast, the cell vertices that require further updates are typically located at the edges of objects, where the occupancy values significantly differ from those of surrounding cell vertices. Therefore, the gradient at these cell vertices is larger. An example illustrating this is shown in Fig. \ref{fig_select_example_b}, where the selected area (in white and black) is clearly distinct from the stable area (in gray). Based on this idea, we propose a strategy to select the cell vertices located around the boundaries to compose the selected high-resolution map $\mathbb{M}^{s}$, which is used in the second stage of optimization.

\begin{figure}[t]
\centering 
\includegraphics[width=0.48\textwidth]{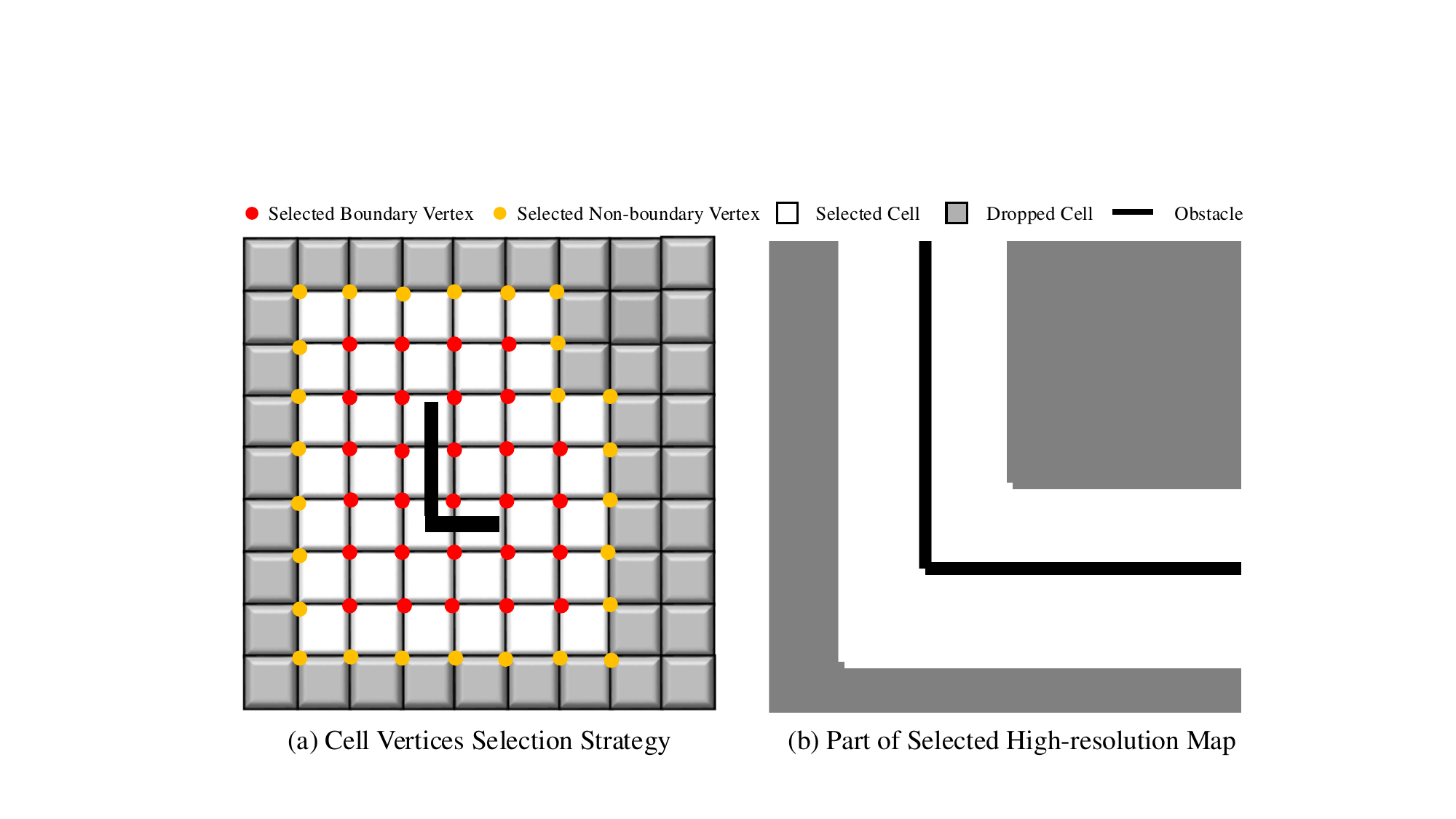}
\caption{\label{fig_low_high_select} An illustration of the cell vertices selection strategy and a selected high-resolution map from a simulation dataset. In (a), selected cell vertices are marked in red and yellow, with their indices forming the index set $\mathbb{I}^{s}$.}
\end{figure}


Firstly, we identify cell vertices located at the edges of objects by performing mean-value convolution of the full high-resolution map $\mathbb{M}^{h}$. Specifically, we calculate a binary map $\mathbb{B}=\{B(\mathbf{m}_{id})\}$ by binarizing $\mathbb{M}^{h}$ as
\begin{equation}
B(\mathbf{m}_{id}) =	\begin{cases}
	1, & {M}^{h}(\mathbf{m}_{id}) \geq \tau_{occupied} \\
	0, & {M}^{h}(\mathbf{m}_{id}) < \tau_{occupied} \\
\end{cases},
\end{equation}
where $\mathbf{m}_{id}$ represents a cell vertex, and $\tau_{occupied}$ is the threshold used to classify a cell vertex as occupied or free. A mean-value convolution kernel $\mathbf{K}$ is defined as
\begin{equation}
	\mathbf{K} = \dfrac{1}{q^2} \cdot \bold{1}_{q\times q}
\end{equation}
where $\bold{1}_{q\times q}$ represents a $q \times q$ matrix of ones. The convoluted map $\mathbb{C}=\{C(\mathbf{m}_{id})\}$ is then derived by convolving $\mathbb{B}$ with $\mathbf{K}$, where $C(\mathbf{m}_{id})$ indicates whether the $q \times q$ cell vertices around $\mathbf{m}_{id}$ are all in the same occupancy state. Compared to other edge detection methods like Sobel \cite{duda1973pattern} and Canny \cite{canny1986computational}, this conservative method more reliably selects cell vertices that may require further optimization. 

Using this method, the set of indices for all boundary cell vertices in the high-resolution map is defined as 

\begin{equation}
\begin{aligned}
	\mathbb{I}^{h} = \{id | 0<C(\mathbf{m}_{id})<1 \}.
\end{aligned}
\end{equation}
The cell vertices indexed in $\mathbb{I}^{h}$ are marked in red in Fig. \ref{fig_low_high_select}(a).

To account for pose uncertainties from the first stage, the selection is expanded to include cell vertices within a distance $d$ from all boundary cell vertices. The indices of the selected cell vertices in the high-resolution map form the set $\mathbb{I}^{s}$, illustrated in Fig. \ref{fig_low_high_select}(a), where the selected cell vertices are highlighted in red and yellow with $d=1$. An example of a selected high-resolution map from a simulation dataset is shown in Fig. \ref{fig_low_high_select}(b).

Consequently, the map component of the state vector in the second stage is expressed as
\begin{equation}
	\mathbf{x}^{sM} = {[\cdots, M^{h}(\mathbf{m}_{wh}), \cdots]}^\top, ~~ wh\in \mathbb{I}^{s}.
\end{equation}

\begin{figure}[t]
\centering 
\includegraphics[width=0.48\textwidth]{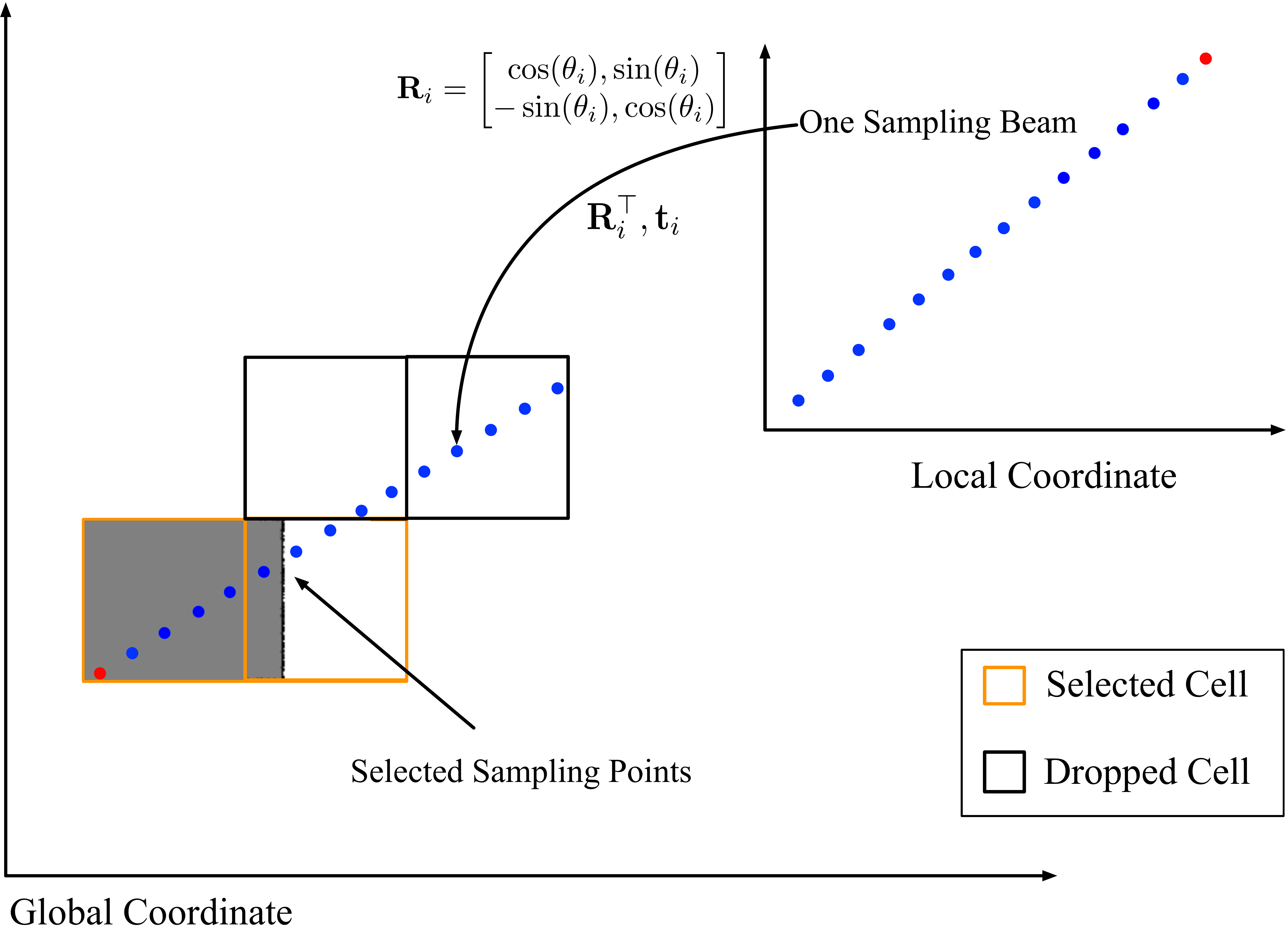}
\caption{\label{fig_select_sampling_point} An example of the selected sampling points of a beam at time step $i$, where points projected onto the selected cells are chosen.}
\end{figure}

Next, we select observations to optimize $\mathbf{x}^{sM}$. Cells surrounded by vertices with indices in $\mathbb{I}^s$ are designated as selected cells, shown in white in Fig. \ref{fig_low_high_select}(a) and Fig. \ref{fig_low_high_select}(b). Subsequently, sampling point selection is carried out, as illustrated in Fig. \ref{fig_select_sampling_point}. Specifically, sampling points in $\mathbb{S}^{h}$ are first projected onto the global coordinate system using the poses optimized in the first stage. All sampling points located on the selected cells are then included to form the set $\mathbb{S}^{s}$. 

\section{Submap Joining} \label{Sec_submap}
In Section \ref{Sec_multi}, we introduced a multi-resolution joint optimization strategy to efficiently solve our NLLS problem. For large-scale occupancy SLAM with long robot trajectories, the number of poses to optimize can be very large. To make the computational complexity mainly depend on the environment size rather than the trajectory length, in this section we propose an occupancy submap joining method. The key idea is to reduce the number of poses that need to be optimized to the number of local submaps.

\subsection{Inputs and Outputs of Submap Joining Problem} 
We first separate the observation information into multiple parts and perform Algorithm \ref{alg_flowchart} to build several submaps. The inputs of submap joining problem are a sequence of local occupancy submaps. 
Let us denote the $n_L+1$ submaps as $\mathbb{M}_L = \left\{\mathbb{M}_{L_0}, \cdots, \mathbb{M}_{L_{n_L}}\right\}$ and the associated coordinate frames of these local occupancy maps are denoted as $ \{\mathbf{x}^P_{0}, \cdots, \mathbf{x}^P_{n_L}\}$, where $\mathbb{M}_{L_{i_L}}$ and $\mathbf{x}^P_{i_L}$represents the $i_L$th local occupancy map and its associated coordinate frame. In addition, the global occupancy map is represented as $\mathbb{M}_G=\left\{M_G(\mathbf{m}^G_{00}), \cdots,{M}_G\left(\mathbf{m}^G_{c_w^Gc_h^G}\right)\right\}$. Both the global map and local maps follow the same definition as described in Section \ref{sec_discrete_occupancy}. The outputs of submap joining problem are the optimal solution of the local submap coordinate frames and the optimal global occupancy map.

\subsection{NLLS Formulation of Submap Joining Problem} 
First, the cell vertex $\mathbf{m}^G_{wh}$ in the global occupancy map $\mathbb{M}_G$ can be projected to local submap coordinate by pose $\mathbf{x}^P_{i_L}$, i.e., 
\begin{equation}
	\mathbf{p}_{i_L}^{wh} = \frac{ \mathbf{R}_{i_L} (\mathbf{m}^G_{wh} \cdot s_G  - \mathbf{t}_{i_L})}{s_L}.
\end{equation}
Here, $\mathbf{p}_{i_L}^{wh}$ represents the position in the local submap's coordinate where the cell vertex $\mathbf{m}_{wh}^G$ from the global map is projected using the pose $\mathbf{x}^P_{i_L}$. The resolutions of the global occupancy map and local submaps are denoted by $s_G$ and $s_L$, respectively.

The submap joining problem aims to find the optimal global occupancy map and the poses of submap coordinate frames. Thus, the state vector for this problem is defined as 
\begin{equation}
	\mathbf{x}_G = \left[{\mathbf{x}^P_L}^\top, {{\mathbf{x}^M_G}}^\top\right]^\top,
\end{equation}
where 
\begin{equation}
\begin{aligned}
\mathbf{x}^P_L & =\left[\left(\mathbf{x}^P_1\right)^\top, \cdots,\left(\mathbf{x}^P_{n_L}\right)^\top\right]^\top \\
\mathbf{x}^M_G & =\left[{M}_G\left(\mathbf{m}^G_{00}\right), \cdots, {M}_G\left(\mathbf{m}^G_{c_w^Gc_h^G}\right)\right]^\top.
\end{aligned}
\end{equation}
As with most submap joining problem formulations, we fix the first local map coordinate frame as the global coordinate frame. Therefore, $\mathbf{x}^P_L$ consists of $n_L$ local map coordinate frames and $\mathbf{x}^M_G$ includes $(c_w^G+1) \times (c_h^G+1)$ discrete cell vertices of global occupancy map. 

By the global-to-local projection relationship, all cell vertices of global occupancy map $\mathbb{M}_G$ can be projected to corresponding submaps to compute the difference in occupancy values. Thus, the NLLS problem of occupancy submap joining can be formulated to minimize 
\begin{equation}
\begin{adjustbox}{max width=\linewidth}
$
g(\mathbf{x}_G) =  \sum\limits_{i_L=0}^n\sum\limits_{ wh \in \mathbb{S}^G_{i_L}} \left\| \omega(i_L,\mathbf{m}^G_{wh}) {M}_G(\mathbf{m}^G_{wh}) - {M}_{L_{i_L}}(\mathbf{p}_{i_L}^{wh}) \right\|^2,
$
\end{adjustbox}
\label{eq_NLLS_joining}
\end{equation}
where $\mathbb{S}^G_{i_L}$ represents the set of indices of cell vertices in the global occupancy map $\mathbb{M}_G$ that are projected onto the local submap $\mathbb{M}_{L_{i_L}}$.

In (\ref{eq_NLLS_joining}), $\omega(i_L,\mathbf{m}^G_{wh})$ is the weight to establish an accurate relationship between the global occupancy map and local submaps w.r.t. occupancy values, which can be calculated by
\begin{equation}
    \omega(i_L,\mathbf{m}^G_{wh}) = \frac{{N}_{{L_{i_L}}}(\mathbf{p}_{i_L}^{wh})}{{N}_{G}(\mathbf{m}^G_{wh})}.
\end{equation}
Here, ${N}_{{L_{i_L}}}(\cdot)$ is the local hit number lookup function for submap $\mathbb{M}_{L_{i_L}}$, derived as described in Section \ref{sec_hit}. It approximates the hit number at coordinate $\mathbf{p}_{i_L}^{wh}$ using bilinear interpolation. Similarly, ${N}_{G}(\cdot)$ represents the global hit number lookup function associated with $\mathbb{M}_G$.

In (\ref{eq_NLLS_joining}), the submap joining problem is formulated as a NLLS problem, which can be solved iteratively by Gauss-Newton based method similar to Algorithm \ref{alg_1}.

 \begin{table}[htp]
		\centering
		\caption{Parameters of Datasets. \label{tab_dataset}}
		\label{tab_comparison}
		\setlength{\tabcolsep}{0.7mm}{
		\begin{tabular}{lccccc}\toprule
		Dataset	& No. Scans & Duration  & Map Size &  Odometry & Resolution\\ \hline
		Simulation 1 & 3640  &117 s& $50$ m  $\times$ $50$ m & yes & 0.05 m\\
        Simulation 2 & 3720  &121 s& $50$ m $\times$ $50$ m & yes & 0.05 m\\
		Simulation 3  & 2680  & 83 s& $50$ m $\times$ $50$ m & yes & 0.05 m\\
		Car Park  & 1642 & 164 s& $50$ m $\times$ $40$ m & yes & 0.1 m\\
		C5  & 3870  &136 s& $50$ m $\times$ $40$ m & yes & 0.1 m\\
		Museum b0 & 5522 &152 s& $85$ m $\times$ $95$ m &no & 0.1 m \\
		Museum b2 & 51833 &1390 s &  $250$ m $\times$ $200$ m &no & 0.1 m\\
        C3 &24402 &610 s& $150$ m $\times$ $125$ m  & no & 0.1 m\\
		\hline
		\end{tabular}
		}
\end{table}

\section{Experimental Results} \label{Sec_experiment}

In this section, we evaluate our algorithm on several datasets and compare its performance with Cartographer \cite{hess2016real}, the current state-of-the-art algorithm, which significantly outperforms other methods such as Hector-SLAM \cite{kohlbrecher2011flexible} and Karto-SLAM \cite{konolige2010efficient}. To ensure fair comparisons, we adjust some parameters in Cartographer based on the sensor configurations of the respective datasets for optimal performance.

The dataset parameters are summarized in Table \ref{tab_dataset}. For practical datasets, Deutsches Museum b0 and Deutsches Museum b2 are Cartographer demo datasets collected at the Deutsches Museum. The Car Park \cite{zhao20212d} dataset is gathered in an underground car park, while C5 and C3 are collected in a factory environment using a Hokuyo UTM-30LX laser scanner. Consistent map resolutions $s$ are applied across all methods to display the map results, with ratio $r$ set to $10$ for all simulation experiments and $5$ for all practical experiments unless stated otherwise. For each dataset, 20\% of scans and corresponding poses are uniformly selected as key frames for the key frame option in our method.

To ensure fair comparisons, we use an identical number of poses (synchronizing the poses from the results with the ground truth poses using timestamps) and their corresponding observations to generate results for visualization and quantitative evaluation across all compared methods, with the exception of our method that employs keyframes. Furthermore, the same occupancy mapping algorithm is applied consistently across all approaches to produce the occupancy grid map results for comparison.

\subsection{Simulation Experiments}\label{simu_experiment}

\begin{figure*}[tp]
\centering \subfigure[Simulation 1] {\label{fig_trajectory_1}
\includegraphics[width=0.28\textwidth]{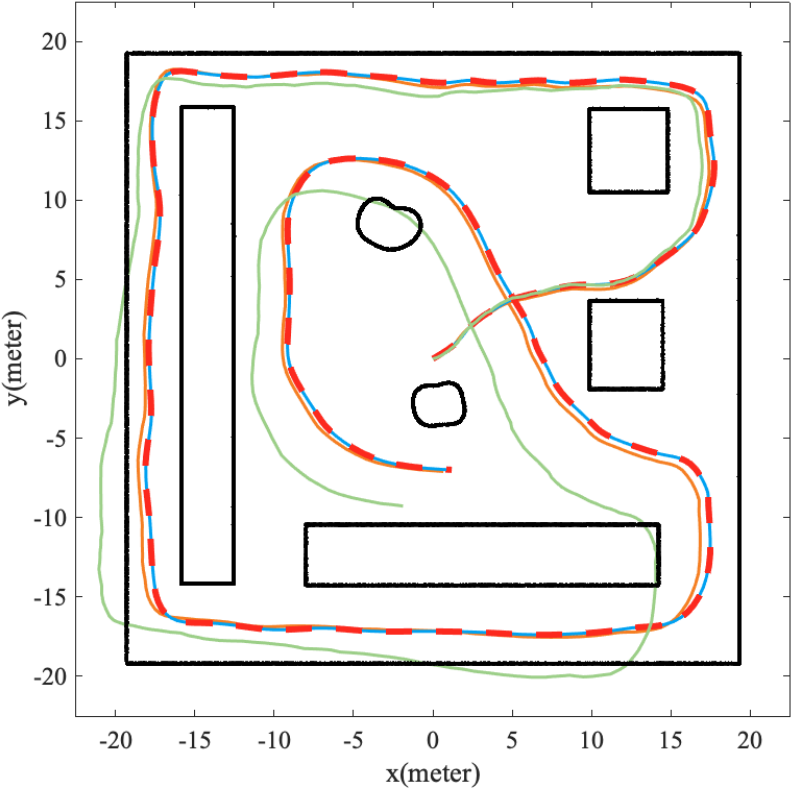}}
\centering \subfigure[Simulation 2] {\label{fig_trajectory_2}
\includegraphics[width=0.28\textwidth]{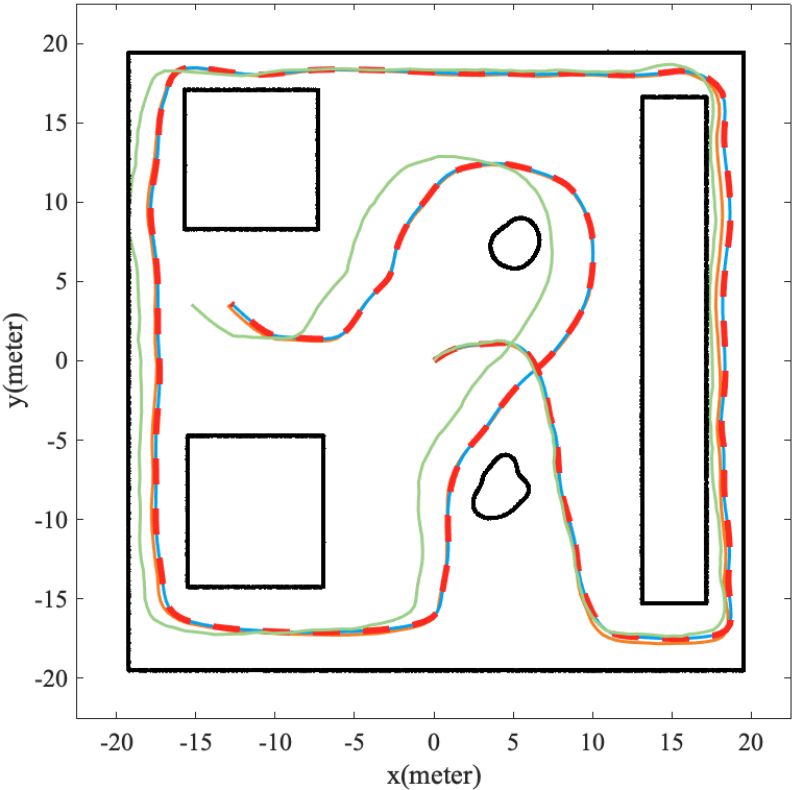}}
\centering \subfigure[Simulation 3] {\label{fig_trajectory_3}
\includegraphics[width=0.343\textwidth]{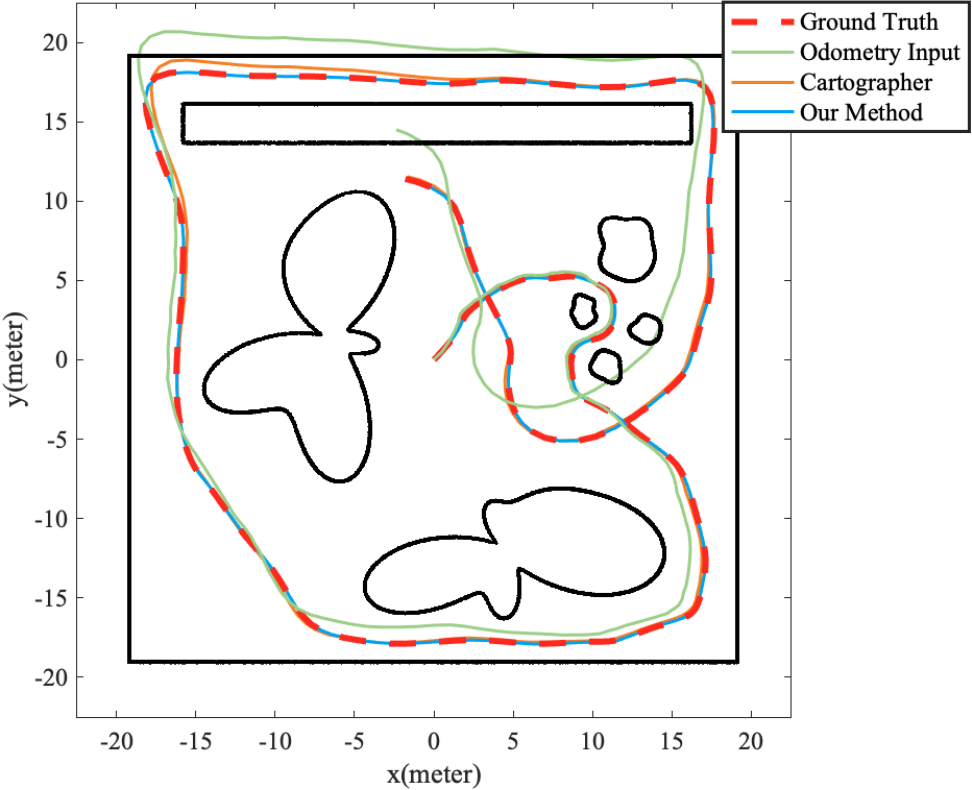}}
\caption{\label{fig_trajectory_compare}Simulation environments and robot trajectory results. (a), (b) and (c) show the simulation environments (the black lines indicate the obstacles in the scene) and the trajectories of ground truth, odometry inputs, Cartographer \cite{hess2016real}, and our approach for one dataset in each of the three simulation environments.}
\end{figure*}

We use three different simulation environments with varying levels of nonlinearity and nonconvex obstacles to design three different simulation experiments. Since Cartographer needs a high-frequency scanning rate to ensure the good performance of scan matching, while our approach performs well for scan data with low scanning frequency, only 10\% scans listed in Table \ref{tab_dataset} are used in our method. 

We utilize the open-source 2D LiDAR simulator from \cite{zhao20212d} to generate simulated datasets. Each scan includes 1081 laser beams spanning angles from -135 degrees to 135 degrees, mimicking the specifications of a Hokuyo UTM-30LX laser scanner. To emulate real-world data acquisition, random Gaussian noise with zero mean and standard deviation of $0.02$ m is added to each beam of the simulated scan data. Similarly, zero-mean Gaussian noise is introduced to the odometry inputs derived from the ground truth poses, with standard deviation of $0.04$ m for $x$-$y$ and $0.003$ rad for orientation. Five datasets with different noise realizations are generated for each simulation environment.

\begin{figure*}[t]
\centering \subfigure[Simulation 1] {\label{fig_time_error_1}
\includegraphics[width=0.32\textwidth]{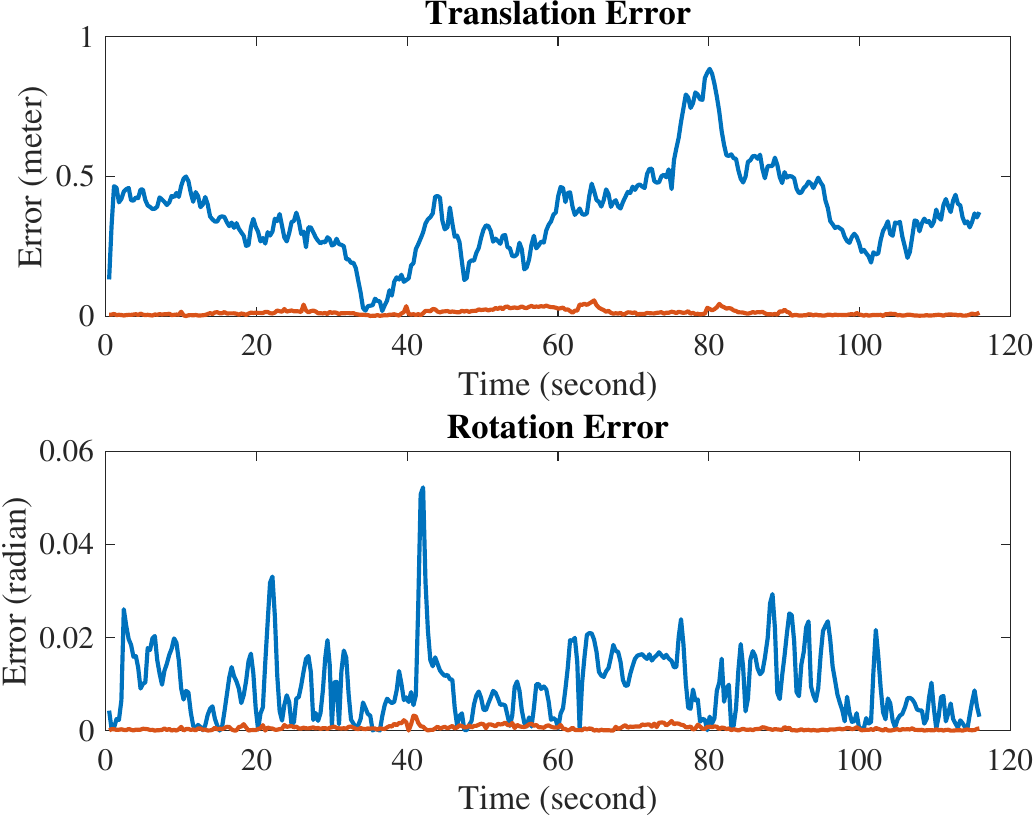}}
\centering \subfigure[Simulation 2] {\label{fig_time_error_2}
\includegraphics[width=0.32\textwidth]{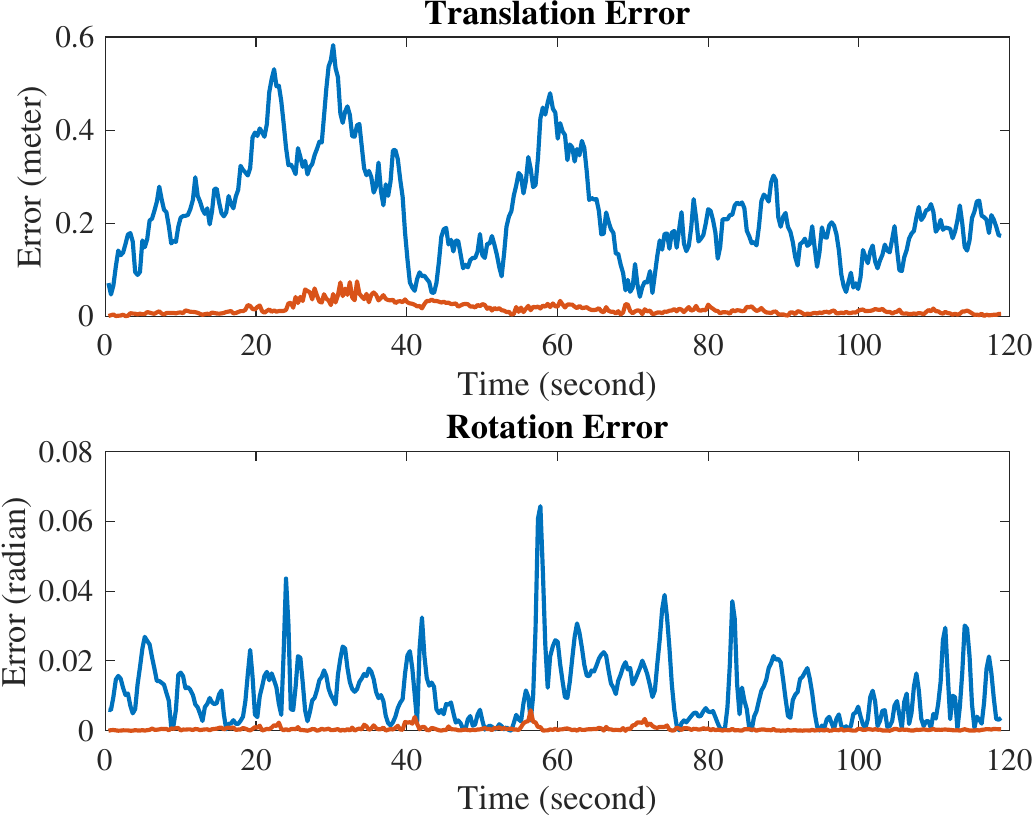}}
\centering \subfigure[Simulation 3] {\label{fig_time_error_3}
\includegraphics[width=0.32\textwidth]{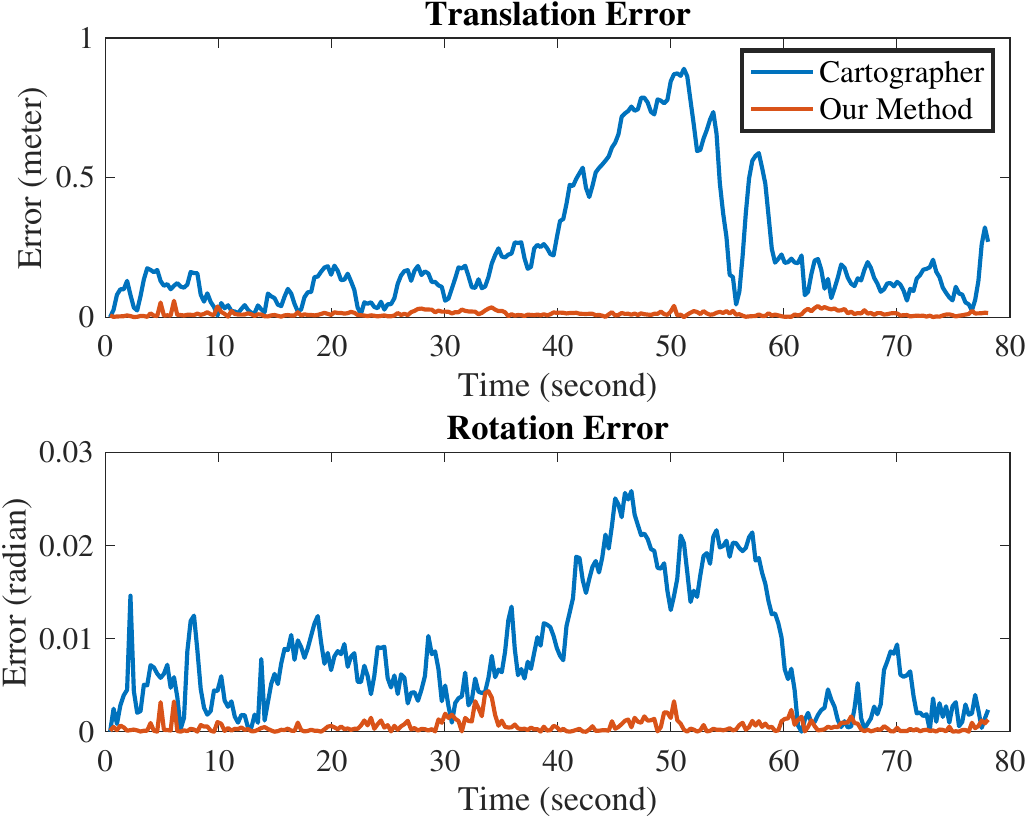}}
\caption{Comparison of translation and rotation errors at different time steps using simulation datasets.}
\label{fig_error_compare_time}
\vspace{-1em}
\end{figure*}


The trajectory results of our method and Cartographer, compared to ground truth and odometry, are shown in Fig. \ref{fig_trajectory_compare}. It is evident that our trajectories align more closely with the ground truth, particularly in areas with significant rotational movements. Fig. \ref{fig_error_compare_time} illustrates translation and rotation errors over time, demonstrating that our method consistently achieves substantially smaller errors compared to Cartographer.

\begin{figure}[t]
\centering
\includegraphics[width=0.48\textwidth]{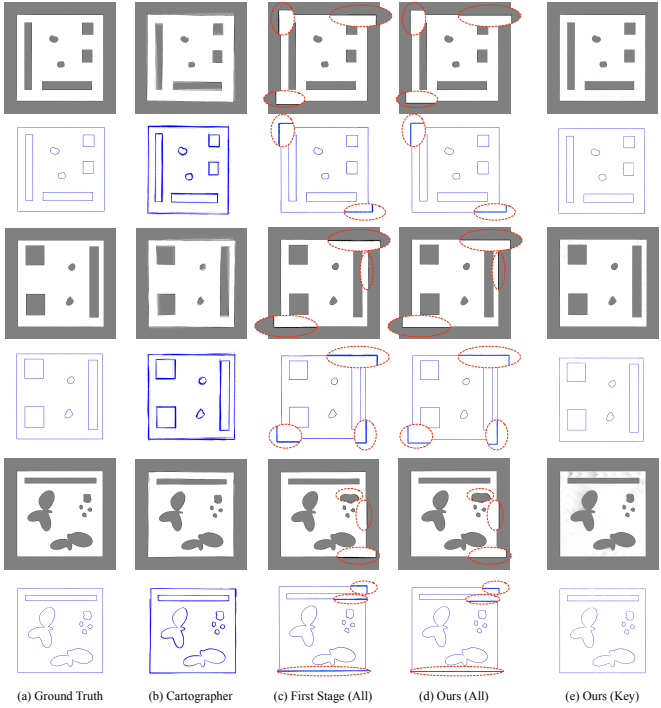}
\caption{\label{fig_simulation} The occupancy grid maps and point cloud maps generated from ground truth poses and different approaches for each simulation dataset. The areas marked with red dots highlight where our method outperforms the results of the first-stage optimization alone.}
\vspace{-0.5em}
\end{figure}

We performed quantitative and qualitative comparisons of pose estimation errors using all fifteen datasets from Simulations 1, 2, and 3. Table \ref{tab_comparison} presents, in order, the quantitative results for odometry inputs, Cartographer, the first stage of our method (Algorithm \ref{alg_1} with low-resolution) using all frames, our method using all frames, and our method using key frames. Metrics such as mean absolute error (MAE) and root mean squared error (RMSE) evaluate translation errors (in meters) and rotation errors (in radians). Our method consistently achieves the best performance across all metrics, significantly outperforming Cartographer even when using only key frames or the first stage. Fig. \ref{fig_simulation}(a) to Fig. \ref{fig_simulation}(e) further illustrates occupancy grid maps and point cloud maps generated using poses from the ground truth, Cartographer, and the three options of our method. The maps produced by our method exhibit noticeably clearer boundaries, demonstrating its ability to jointly optimize robot poses and occupancy maps for more accurate results.

\begin{table}[t]
		\centering
		\caption{Quantitative Comparison of Robot Pose Errors in Simulations.}
		\label{tab_comparison}
		\setlength{\tabcolsep}{0.7 mm}{
		\begin{tabular}{lccccc}\toprule
			& Odom & Carto & Ours (First) & Ours (All) & Ours (Key) \\ \hline
		Simulation 1& & & &\\
		\quad MAE / Trans (m) & 0.78270 & 0.25336  & 0.02206 &\textcolor{red}{\textbf{0.00640}} & \textcolor{blue}{\textbf{0.01024}}\\
		\quad MAE / Rot (rad) & 0.04912 & 0.01394  & 0.00098 &\textcolor{red}{\textbf{0.00060}} & \textcolor{blue}{\textbf{0.00084}}
\\
		\quad RMSE / Trans (m) & 0.98404 & 0.29920  & 0.02680 &\textcolor{red}{\textbf{0.00974}} & \textcolor{blue}{\textbf{0.01430}}\\
		\quad RMSE / Rot(rad) & 0.05506 & 0.01562  & 0.00162 &\textcolor{red}{\textbf{0.00102}} & \textcolor{blue}{\textbf{0.00126}}\\\hline
		
		Simulation 2& & & &\\
		\quad MAE / Trans (m) & 0.80544 & 0.11914 &0.03224 & \textcolor{red}{\textbf{0.00858}} & \textcolor{blue}{\textbf{0.01082}}
\\
		\quad MAE / Rot (rad) & 0.02538 & 0.00666  &0.00220 &\textcolor{red}{\textbf{0.00062}} & \textcolor{blue}{\textbf{0.00096}}\\
		\quad RMSE / Trans (m) & 0.97152 & 0.14810  & 0.04188 &\textcolor{red}{\textbf{0.01198}} & \textcolor{blue}{\textbf{0.01244}}\\
		\quad RMSE / Rot (rad) & 0.02936 & 0.00916 &0.00220 & \textcolor{red}{\textbf{0.00104}} & \textcolor{blue}{\textbf{0.00178}}\\\hline
		
		Simulation 3& & & &\\
		\quad MAE / Trans(m) & 0.75352 & 0.14262  &0.02624 &\textcolor{red}{\textbf{0.00726}} &  \textcolor{blue}{\textbf{0.00998}}\\
		\quad MAE / Rot (rad) & 0.05180 & 0.00682 &0.00164  & \textcolor{red}{\textbf{0.00058}} &  \textcolor{blue}{\textbf{0.00090}}\\
		\quad RMSE / Trans (m) & 0.96866 & 0.18782 & 0.03238 &\textcolor{red}{\textbf{0.00952}} &  \textcolor{blue}{\textbf{0.01338}}\\
		\quad RMSE / Rot (rad) & 0.05926 & 0.00914  & 0.00204 &\textcolor{red}{\textbf{0.00088}} &  \textcolor{blue}{\textbf{0.00134}}\\\hline
		\end{tabular}
	\begin{tablenotes}
     \item \textcolor{red}{\textbf{Red}} and  \textcolor{blue}{\textbf{blue}} indicate the best and second best results, respectively.
   \end{tablenotes}
		}
\end{table}

\begin{table*}[htp]
\centering
\caption{Occupancy Grid map Precision of Our Method Using All Frames, Our Method Using Key Frames, and Cartographer.}
\label{tab_map_accuracy}
\setlength{\tabcolsep}{2.4mm}
\renewcommand{\arraystretch}{1.2}

\begin{tabular}{c c c c c c c c c c c c}
\toprule
 \multirow{2}{*}{} & \multirow{2}{*}{\textbf{Ground Truth}}& \multicolumn{3}{c}{Our Method (All Frames)} & \multicolumn{3}{c}{Our Method (Key Frames)} & \multicolumn{3}{c}{Cartographer} \\
\cmidrule(lr){3-5} \cmidrule(lr){6-8} \cmidrule(lr){9-11}
& & Unknown & Free & Occupied & Unknown & Free & Occupied & Unknown & Free & Occupied \\
\midrule
\multirow{3}{*}{Simulation 1} 
& Unknown  & \textcolor{red}{\textbf{99.798\%}}  & 0.020\% & 0.182\% & \textcolor{blue}{\textbf{99.598\%}} & 0.072\% & 0.330\% & 95.616\% & 2.822\% & 1.562\% \\
& Free     & 0.022\%  & \textcolor{red}{\textbf{99.938\%}} & 0.040\% & 0.094\% & \textcolor{blue}{\textbf{99.824\%}} & 0.082\% & 1.290\% & 97.678\% & 1.032\% \\
& Occupied & 5.436\%  & 2.334\%  & \textcolor{red}{\textbf{92.230\%}} & 13.562\% & 3.142\% & \textcolor{blue}{\textbf{83.296\%}} & 30.053\% & 53.280\% & 16.667\% \\
\midrule
\multirow{3}{*}{Simulation 2} 
& Unknown  & \textcolor{red}{\textbf{99.846\%}}  & 0.010\% & 0.144\% & \textcolor{blue}{\textbf{99.696\%}} & 0.062\% & 0.242\% & 96.868\% & 1.743\% & 1.389\% \\
& Free     & 0.016\%  & \textcolor{red}{\textbf{99.846\%}} & 0.138\% & 0.076\% & \textcolor{blue}{\textbf{99.848\%}} & 0.076\% & 0.593\% & 98.584\% & 0.823\% \\
& Occupied & 6.434\%  & 3.738\%  & \textcolor{red}{\textbf{89.828\%}} & 11.694\% & 2.806\% & \textcolor{blue}{\textbf{85.500\%}} & 23.943\% & 50.795\% & 25.262\% \\
\midrule
\multirow{3}{*}{Simulation 3} 
& Unknown  & \textcolor{red}{\textbf{99.812\%}}  & 0.032\% & 0.156\% & \textcolor{blue}{\textbf{99.258\%}} & 0.430\% & 0.312\% & 96.968\% & 1.574\% & 1.458\% \\
& Free     & 0.036\%  & \textcolor{red}{\textbf{99.928\%}} & 0.036\% & 0.554\% & \textcolor{blue}{\textbf{99.352\%}} & 0.094\% & 1.018\% & 98.110\% & 0.872\% \\
& Occupied & 4.500\%  & 2.358\%  & \textcolor{red}{\textbf{93.142\%}} & 16.788\% & 3.736\% & \textcolor{blue}{\textbf{79.476\%}} & 26.420\% & 44.928\% & 28.652\% \\
\bottomrule
\end{tabular}
\end{table*}

From the results of our first stage shown in Fig. \ref{fig_simulation}(c), it is evident that further optimization is needed at the edges of objects. This is due to sampling points with different occupancy values being projected onto the coarse grid cells at object boundaries, causing inaccurate data associations. These results highlight the necessity of the second stage in our multi-resolution strategy (Algorithm \ref{alg_3}) to improve accuracy. Additionally, as shown in Fig. \ref{fig_simulation}(c), the non-edge areas (stable areas) of the occupancy grid maps are well optimized, supporting the fact that including occupancy cell vertices of non-edge areas in the state variables for further optimization is unnecessary.

For a quantitative comparison of the occupancy maps, we apply the same threshold across all methods to convert occupancy values into occupancy states. The mapping problem is treated as a classification task, categorizing each grid cell as free, occupied, or unknown. The mapping performance of our method and Cartographer is summarized in Table \ref{tab_map_accuracy}, clearly showing that both variants of our method, one using all frames and the other using keyframes, significantly outperform Cartographer in terms of map accuracy.

\begin{table}[ht]
		\centering
		\caption{Accuracy of the Occupancy Grid Map.}
		\label{tab_auc}
		\setlength{\tabcolsep}{4.8 mm}{
		\begin{tabular}{llccccc}\toprule
		& & AUC & Precision   \\ \hline
		\multirow{3}{*}{Simulation 1}& Cartographer & 0.90878 & 0.95651 \\ & Ours (All) &\textcolor{red}{\textbf{0.99999}} &\textcolor{red}{\textbf{0.99773}}\\ & Ours (Key) & \textcolor{blue}{\textbf{0.99902}} & \textcolor{blue}{\textbf{0.99548}} \\ \hline 
			\multirow{3}{*}{Simulation 2}& Cartographer & 0.96132 &  0.96829 \\ & Ours (All) & \textcolor{red}{\textbf{0.99926}} & \textcolor{red}{\textbf{0.99721}} \\ & Ours (Key) & \textcolor{blue}{\textbf{0.99914}} & \textcolor{blue}{\textbf{0.99638}}\\ \hline
		\multirow{3}{*}{Simulation 3} & Cartographer & 0.92696 &0.96592 \\ & Ours (All)& \textcolor{red}{\textbf{0.99974}} & \textcolor{red}{\textbf{0.99771}} \\ & Ours (Key) & \textcolor{blue}{\textbf{0.99748}} & \textcolor{blue}{\textbf{0.99113}}\\
		 \hline
		\end{tabular}
  }
\end{table}

We also assess performance using AUC (Area under the ROC curve) \cite{bradley1997use} and precision, with ground truth labels generated from the occupancy map based on ground truth poses. To ensure a fair comparison, all unknown cells are excluded from this evaluation, as AUC is a binary classification metric \cite{bradley1997use}. Table \ref{tab_auc} presents the results, showing that our method using all frames achieves the highest performance in both metrics. Even with only key frames, our method surpasses Cartographer. A key factor resulting in Cartographer's lower mapping quality is its lack of a batch optimization method to address errors during submap construction. Although its scan-to-map matching approach reduces cumulative errors more effectively than scan-to-scan matching, its accuracy still falls short compared to our algorithm. Cartographer performs pose graph optimization to adjust the coordinate frames of submaps only when loop closure is detected, leaving errors within the submaps uncorrected. Although global pose graph optimization is applied at the end of the process, it often suffers from an excess of inaccurate and conflicting relative measurements, as well as its susceptibility to local minima, limiting its effectiveness in correcting these errors. Moreover, pose graph optimization typically does not enhance the local details of maps, as it focuses solely on optimizing poses without jointly considering the map. This further highlights the advantage of our approach, which jointly optimizes both robot poses and the occupancy map.

\subsection{Comparisons using Practical Datasets} \label{sec_practical}

\begin{figure}[t]
\centering
\includegraphics[width=0.46\textwidth]{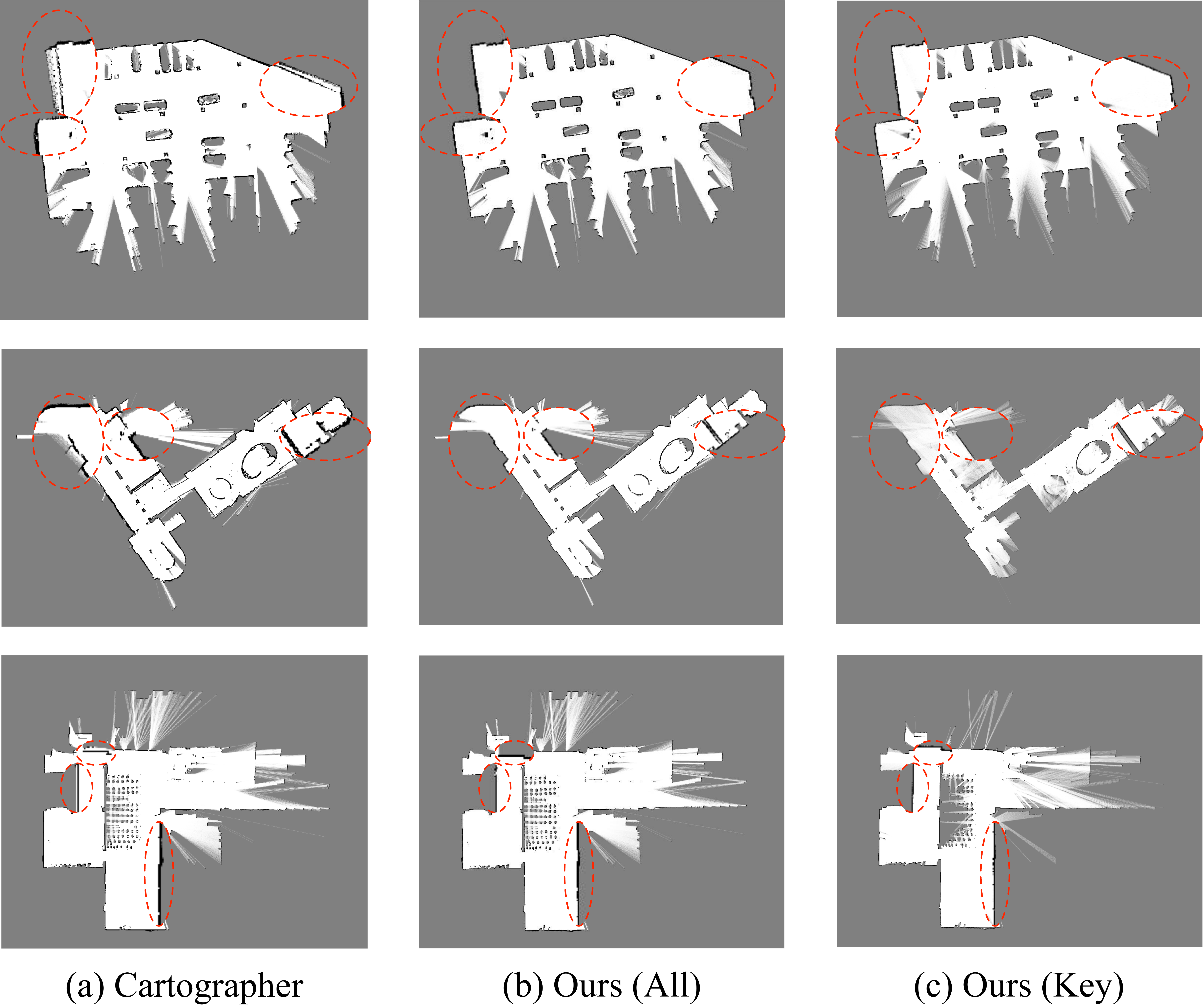}
\caption{\label{fig_result_compare_OGM} The occupancy grid maps from Cartographer, our method using all frames, and our method using key frames. }
\end{figure}

\begin{figure}[t]
\centering
\includegraphics[width=0.46\textwidth]{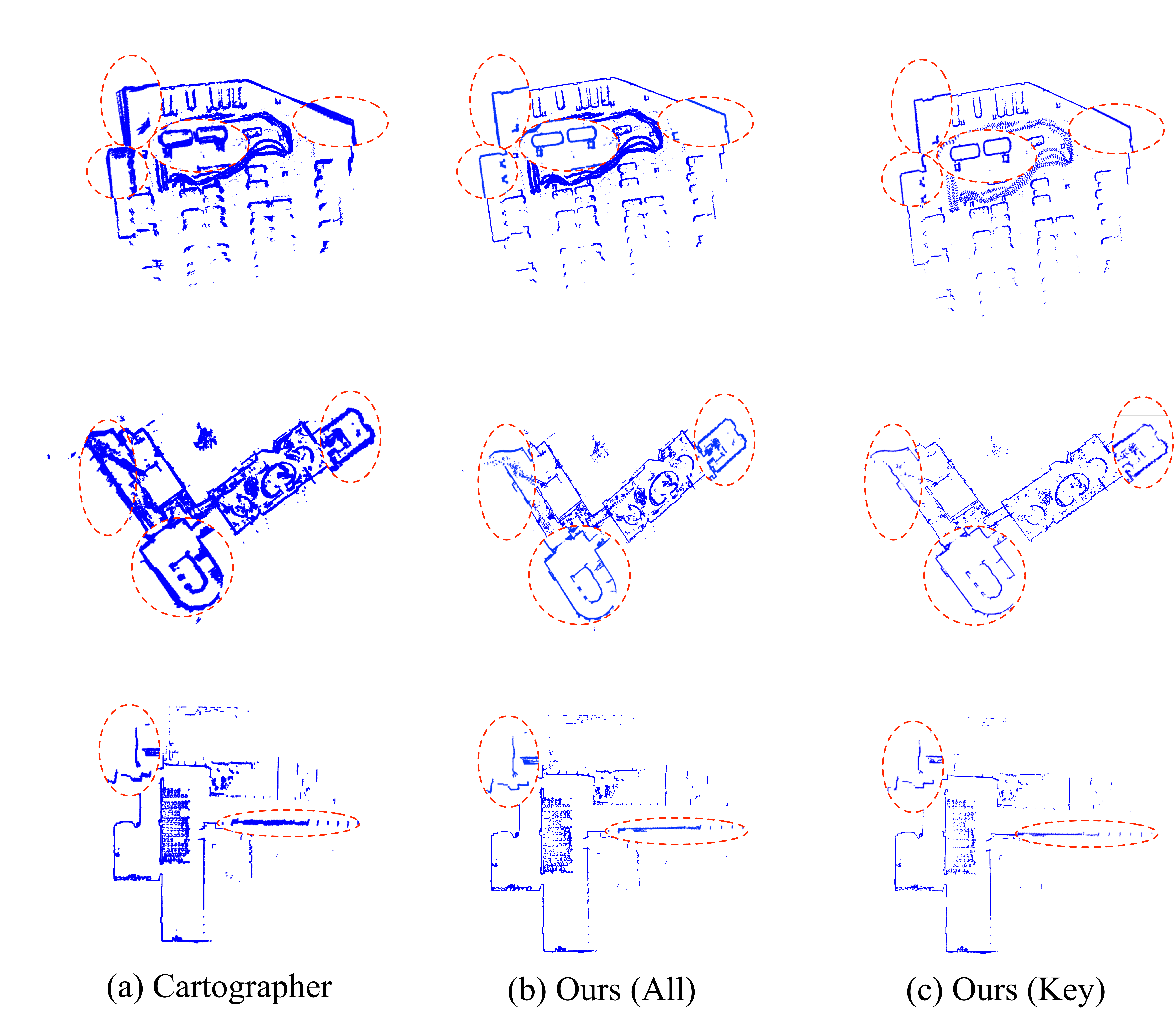}
\caption{\label{fig_result_compare_scan} The point cloud maps from Cartographer, our method using all frames, and our method using key frames.}
\end{figure}

We use three normal-scale practical datasets, namely Deutsches Museum b0 \cite{hess2016real}, Car Park \cite{zhao20212d} and C5, to compare our method with Cartographer in terms of the constructed occupancy grid maps and optimized poses. 

The mapping quality is evaluated by comparing the details of the constructed maps. Additionally, point cloud maps, which are generated using the endpoint projections of scan points and optimized poses, serve as a reference for pose accuracy. For the Car Park and C5 datasets, our method is initialized with poses from odometry inputs, whereas for the Museum b0 dataset, initialization relies on poses from scan matching due to the absence of odometry. The occupancy grid maps and point cloud maps generated by Cartographer, our method using all frames, and our method using key frames for the three datasets are shown in Fig. \ref{fig_result_compare_OGM} and Fig. \ref{fig_result_compare_scan}. Red dotted lines highlight areas where our results outperform Cartographer in both the occupancy grid maps and point cloud maps. Comparing Fig. \ref{fig_result_compare_OGM}(a) and Fig. \ref{fig_result_compare_OGM}(b), our method provides more precise boundaries for the occupancy grid maps due to joint optimization of robot poses and the occupancy map. Similarly, the comparison between Fig. \ref{fig_result_compare_scan}(a) and Fig. \ref{fig_result_compare_scan}(b) illustrates that our method achieves more accurate poses. 

Moreover, our method outperforms Cartographer when using only key frames, as evident from the comparison of Fig. \ref{fig_result_compare_OGM}(a) and Fig. \ref{fig_result_compare_scan}(a) with Fig. \ref{fig_result_compare_OGM}(c) and Fig. \ref{fig_result_compare_scan}(c). These results show that, despite Cartographer introducing loop closure detection, it still produces non-negligible pose errors, leading to point clouds that fail to fully overlap observations of the same obstacle at different poses. While the point cloud maps generated by our method also have non-overlapping parts, these areas are significantly smaller compared to those from Cartographer. 


Additionally, we assess the time consumption of our method and Cartographer on these three datasets. Table \ref{table_time_compare} shows that our method consistently requires less time than Cartographer across all datasets when using all frames and achieves significantly better efficiency when using selected key frames.

\begin{table}[t]
		\centering
		\caption{Time Consumption of Different Algorithms.}
		\label{table_time_compare}
		\setlength{\tabcolsep}{2.5 mm}{
		\begin{tabular}{lccc}\toprule
		Dataset& & Computational Time (s)& \\ \hline
			     & Cartographer  & Ours (All) & Ours (Key)  \\ 
		Car Park & 168  & 119  & \textbf{44} \\
		Museum b0 & 152  & 126 & \textbf{38} \\
		C5 & 146 & 137 & \textbf{35} \\
		\hline
		\end{tabular}
		}
\end{table}

Finally, it is worth noting that some well-known public datasets, such as Radish \cite{Radish}, were collected before 2014 with outdated sensors, leading to low-quality data with poor scanning frequency and odometry accuracy. These issues hinder the performance of Cartographer, often requiring meticulous parameter tuning but still yielding suboptimal results. In contrast, our method performs well on these datasets. Although we do not include these comparisons in this paper, we make our results available on our code page\footnote{\url{https://github.com/WANGYINGYU/Occupancy-SLAM}}.

\subsection{Assessment of Robustness to Initial Guess}


While our method has demonstrated robustness when initialized with odometry inputs or scan matching under reliable sensor conditions in both simulation and real-world experiments, this subsection highlights its capability to converge even when initialized with significantly noisy poses. We use all frames in this subsection to assess robustness.

\begin{figure}[t]
\centering
\includegraphics[width=0.47\textwidth]{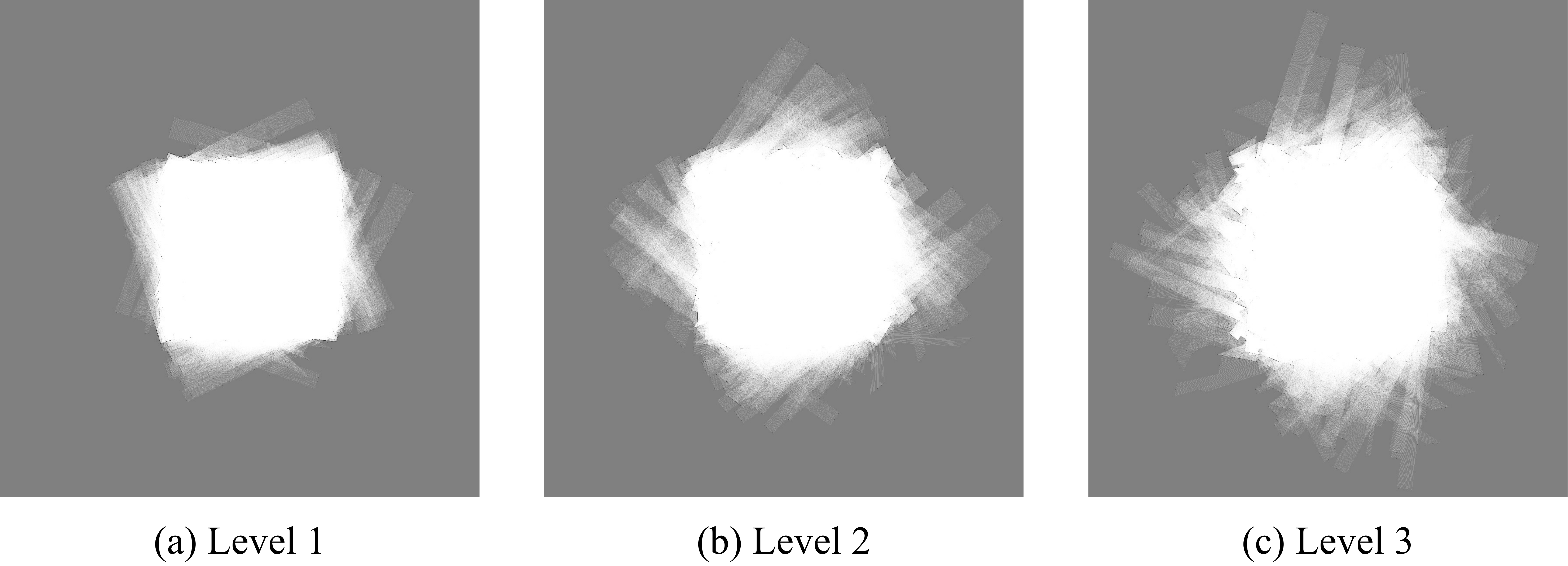}
\caption{\label{fig_OGM_246} Examples of occupancy grid maps generated from poses with different noise levels as shown in Table \ref{table_robustness} using Simulation 1 dataset.}
\end{figure}

First, we use Simulation 1 dataset to quantitatively evaluate the convergence percentage and the accuracy of optimized poses under different noise levels. We add zero-mean uniformly distributed noises with different bounds to the ground truth of the poses to generate each group of ten sets of initial poses for the experiments to count convergence rates and average errors. Specifically, for noise level 1, the noise for translation is within $[-2$ m, $2$ m$]$ and the noise for rotation is within $[-0.5$ rad, $0.5$ rad$]$; for level 2, $[-4$ m, $4$ m$]$ and $[-1$ rad, $1$ rad$]$; for level 3, $[-6$ m, $6$ m$]$ and $[-1.5$ rad, $1.5$ rad$]$. The poses with different noise levels of Simulation 1 dataset are visualized using the generated occupancy grid maps, as shown in Fig. \ref{fig_OGM_246}. The convergence results are depicted in Table \ref{table_robustness}, showing that our method can $100\%$ converge when initialized with challenging noisy poses of level 1 and level 2. Our method still has a high convergence percentage ($80\%$) when initialized with noisy poses of level 3. Our algorithm using other simulation datasets has similar robustness performance.

\begin{table}[t]
		\centering
		\caption{Robustness to Initialization.}
		\label{table_robustness}
		\setlength{\tabcolsep}{0.9 mm}{
		\begin{tabular}{lccc}\toprule
		\thead{Noise Level} & \thead{Convergence\\ Percentage} & \thead{Average MAE of \\Translation (m)} & \thead{Average MAE of\\ Rotation (rad)}\\ \hline
		Level 1 (2 m, 0.5 rad)  & 100\% & 0.00679 & 0.0005   \\
		Level 2 (4 m, 1 rad)  & 100\% & 0.00682 & 0.0005  \\
		Level 3 (6 m, 1.5 rad)  & 80\% & 0.01742  & 0.0012 \\
		\hline
		\end{tabular}
		}
\end{table}

\begin{figure}[t]
\centering
\includegraphics[width=0.5\textwidth]{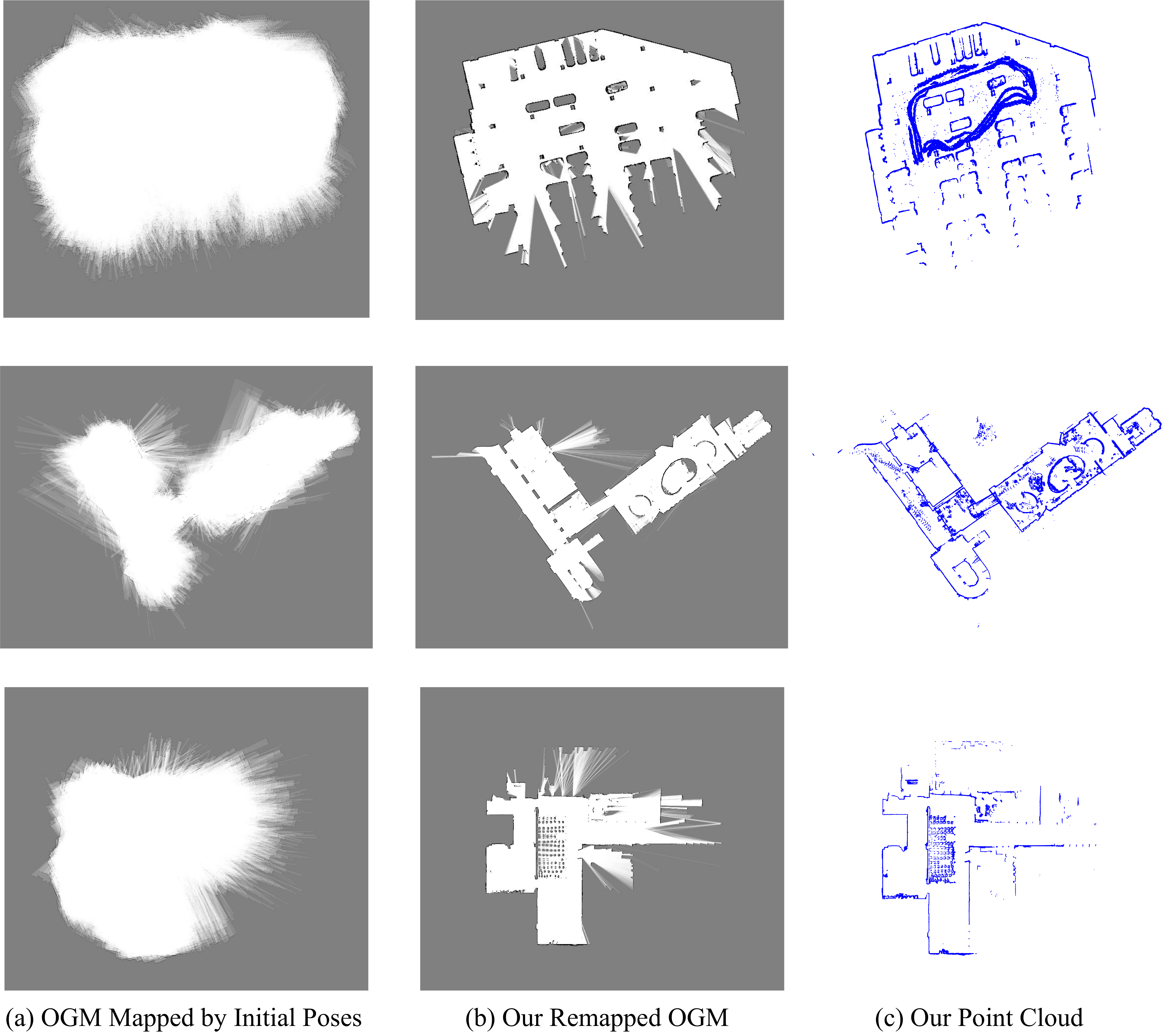}
\caption{\label{fig_noise_initial}The occupancy grid maps and point cloud maps generated using noisy poses for initialization by our approach. (a) and (b) display the remapped occupancy maps generated from the noisy initial poses and our optimized poses, respectively, and (c) shows the point cloud maps created by projecting the endpoints of scans using our optimized poses.}
\end{figure}

Moreover, for all practical datasets, we additionally add random zero-mean uniform distribution noises ($[-2$ m, $2$ m$]$ for translation and $[-0.5$ rad, $0.5$ rad$]$ for rotation) to the poses obtained from Cartographer as the initial guess. The initial occupancy maps obtained by using the noisy initial poses are shown in Fig. \ref{fig_noise_initial}(a). Fig. \ref{fig_noise_initial}(b) shows the remapped occupancy grid maps using our optimized poses, and Fig. \ref{fig_noise_initial}(c) shows the point cloud maps using our optimized poses. This experiment shows that our approach can converge from very poor initial guesses and generate good results.

\subsection{Discussion about the Effectiveness of Different Stages} \label{sec_discuss}

\begin{figure*}[tp]
\centering \subfigure[Simulation 1] {\label{fig_group_error_1}
\includegraphics[width=0.32\textwidth]{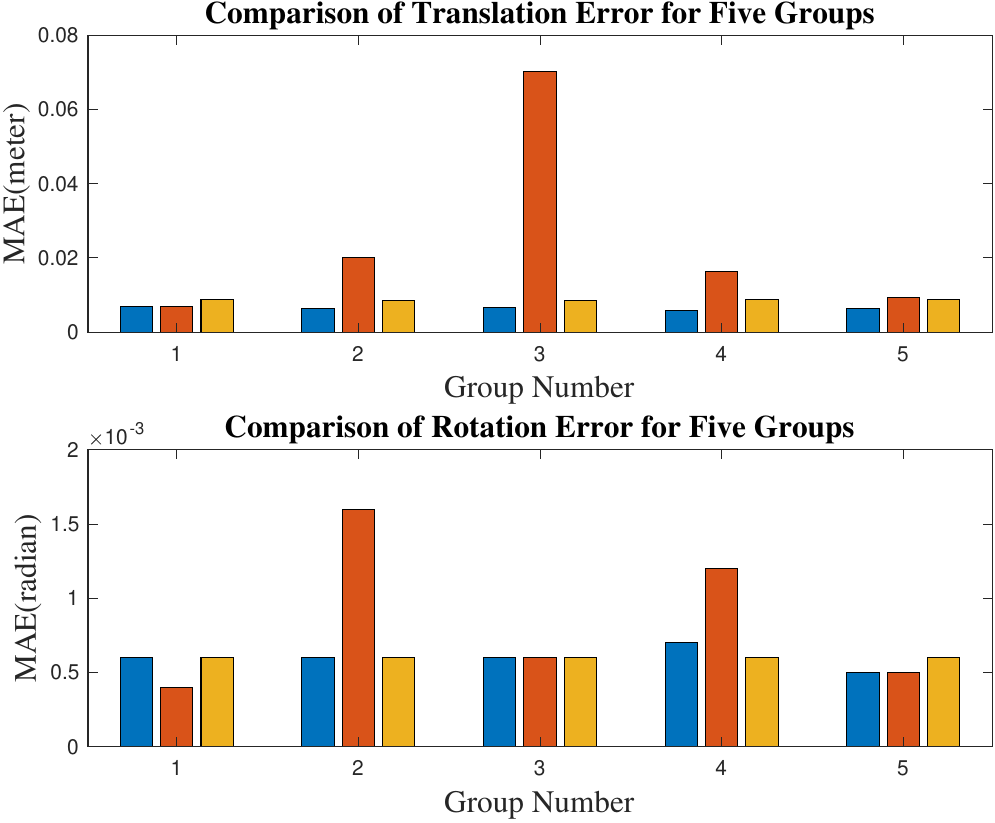}}
\centering \subfigure[Simulation 2] {\label{fig_group_error_2}
\includegraphics[width=0.32\textwidth]{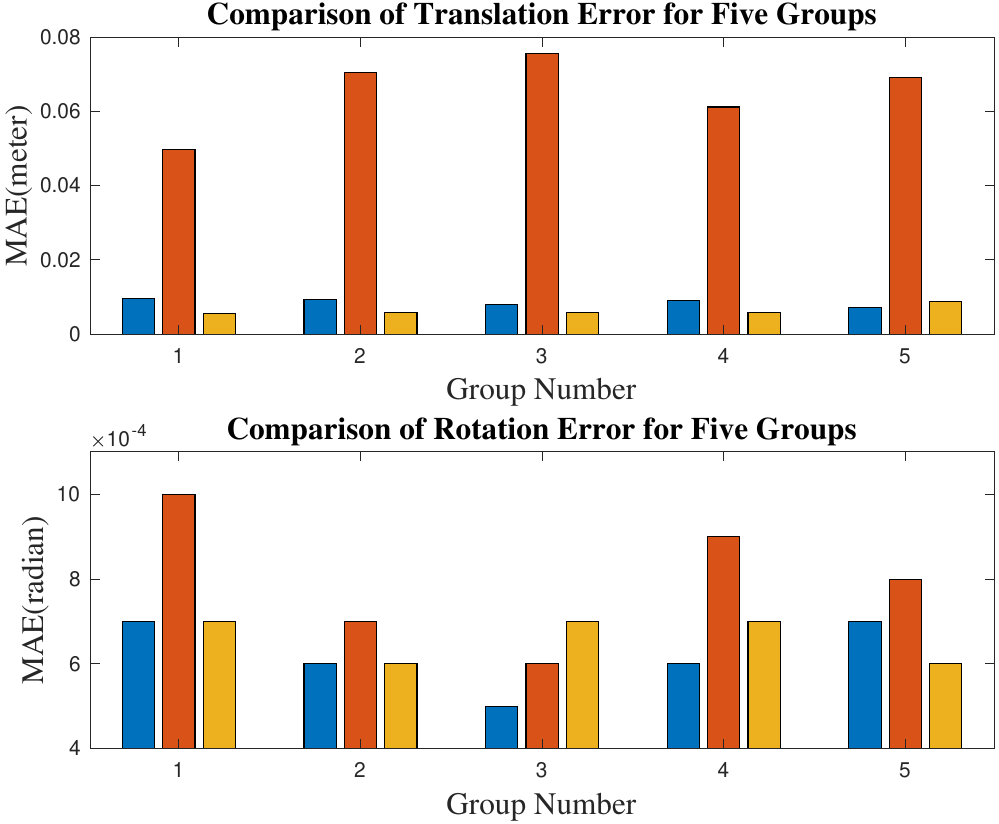}}
\centering \subfigure[Simulation 3] {\label{fig_group_error_3}
\includegraphics[width=0.32\textwidth]{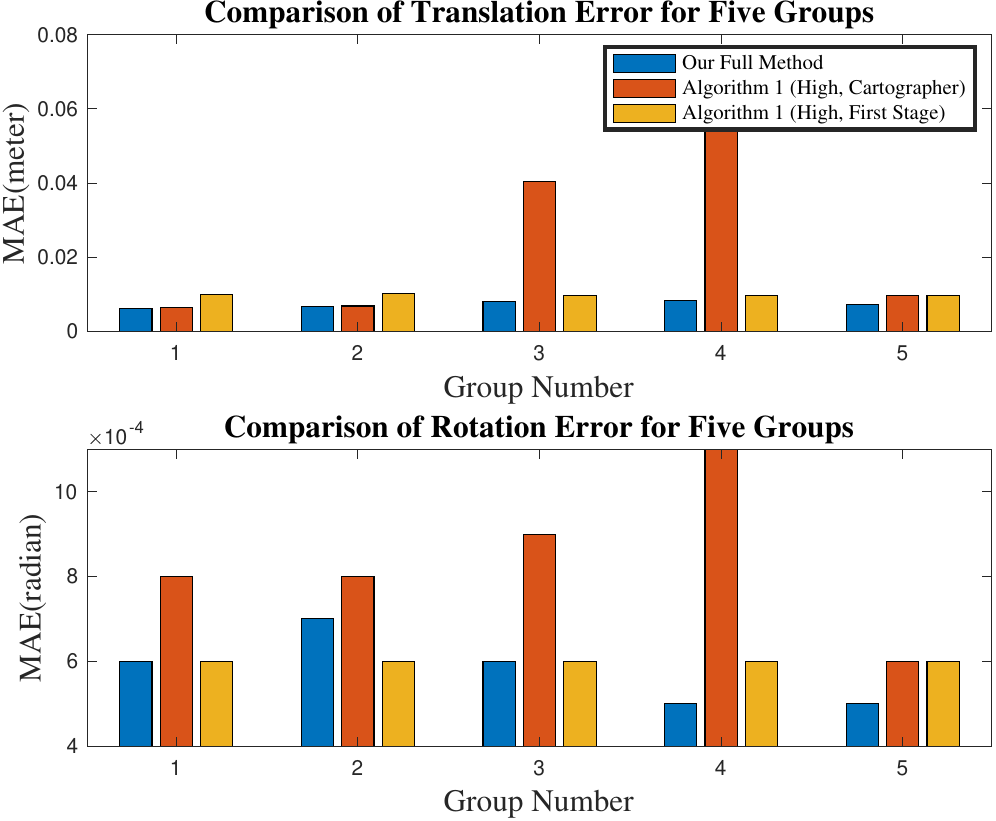}}
\caption{Comparison of translation and rotation errors for simulated datasets using three methods: our full method (Algorithm \ref{alg_flowchart}), our Algorithm \ref{alg_1} initialized by Cartographer's poses with a high-resolution map \cite{Zhao-RSS-22}, and our Algorithm \ref{alg_1} initialized by poses obtained from our first stage with a high-resolution map.}
\label{fig_group_error}
\end{figure*}

In previous sections, we demonstrated the accuracy, robustness, and efficiency of our proposed method. In this section, we discuss the effectiveness of its different parts.

As demonstrated in Table \ref{tab_comparison}, Fig. \ref{fig_simulation}, Fig. \ref{fig_result_compare_OGM}, and Fig. \ref{fig_result_compare_scan}, the accuracy of the poses and the map obtained from our full approach (Algorithm \ref{alg_flowchart}) is much better than those obtained from Cartographer. This confirms the advantage of jointly optimizing both the robot poses and the occupancy map.



One question is whether using only Algorithm \ref{alg_1} with a high-resolution map would yield even better results. To investigate this, we compared our full approach with Algorithm \ref{alg_1} using a high-resolution map. We tested three initialization: (1) \textit{Algorithm \ref{alg_1} (High, O/S)}: initialization using odometry inputs or scan matching; (2) \textit{Algorithm \ref{alg_1} (High, Carto)}: initialization using Cartographer's poses (as proposed in our conference paper \cite{Zhao-RSS-22}); and (3) \textit{Algorithm \ref{alg_1} (High, First)}: initialization using the poses obtained by our first stage.



First, \textit{Algorithm \ref{alg_1} (High, O/S)} fails to converge on most datasets, while our full method converges successfully, indicating the improved robustness of our multi-resolution strategy.

The comparison between our full method, \textit{Algorithm \ref{alg_1} (High, Carto)}, and \textit{Algorithm \ref{alg_1} (High, First)} across all five simulation groups is shown in Fig. \ref{fig_group_error}. It can be observed that the accuracy of our full method remains stable across all groups, while the accuracy of \textit{Algorithm \ref{alg_1} (High, Carto)} varies drastically. This suggests that the approach in our conference paper \cite{Zhao-RSS-22} not only requires an accurate initial guess but also produces less accurate poses than our new method. Moreover, comparing \textit{Algorithm \ref{alg_1} (High, Carto)} with \textit{Algorithm \ref{alg_1} (High, First)} further confirms that the poses obtained in our first stage are more accurate than those of Cartographer.


It is also worth noting that our full method utilizes the selected high-resolution map for optimization in the second stage. As shown in Fig. \ref{fig_group_error}, in certain experiments, the accuracy of our full approach surpasses that of the optimization using the full high-resolution map (\textit{Algorithm \ref{alg_1} (High, First)}). A possible explanation is that once relatively accurate poses and occupancy maps are obtained, the dropped cell vertices and corresponding observations contain little information. Retaining all cell vertices and corresponding observations for optimization may prevent the algorithm from finding the optimal solution, as all observation items are given uniform weights. Another reason could be the smoothing term in (\ref{eq_objective_func}), which spreads occupancy values to unknown cell vertices, potentially introducing errors that affect the convergence of the algorithm.

In terms of time consumption, our full approach is much more efficient than \textit{Algorithm \ref{alg_1} (High, Carto)}. For instance, \textit{Algorithm \ref{alg_1} (High, Carto)} consumes over 21,000 seconds with the Car Park dataset. In comparison, the time consumption of our full approach using all frames is 119 seconds (less than 0.6\%), and using only key frames, it takes only 44 seconds (about 0.2\%). This substantial reduction in time consumption highlights the efficiency improvements of our method over our conference paper \cite{Zhao-RSS-22}.


The reduction in time consumption stems from both the multi-resolution strategy, which reduces time per iteration, and the fewer iterations needed in the second stage due to the selected high-resolution map. Our experiments show that only about two iterations are needed in the second stage with the selected high-resolution map, fewer than in \textit{Algorithm \ref{alg_1} (High, First)}. This is probably because the selected high-resolution map focuses on critical states, with observations containing the most relevant information, enabling faster convergence.

In summary, compared to our conference paper \cite{Zhao-RSS-22}, our new multi-resolution method does not require precise initialization, is far more efficient, and achieves higher accuracy.

\subsection{Ablation Study on the Resolution Ratio}

\begin{table}[t]
		\centering
		\caption{Impact of First-Stage Resolution Settings.}
		\label{table_ablation}
		\setlength{\tabcolsep}{1mm}{
		\begin{tabular}{llcccc}\toprule
		& & $r=20$ & $r=10$  & $r=5$ & $r=2$ \\   \hline   

		 \multirow{5}{*}{Simulation 1} & MAE/Trans (m) First& 0.02352  &  0.02206 & \textbf{0.02118} & 0.17318\\
		\quad & MAE/Rot (rad) First& 0.00116  &  \textbf{0.00098} & 0.00124 & 0.01066\\
		\quad & MAE/Trans (m) All & 0.00812  & \textbf{0.00640}  & 0.00728 & 0.16066\\
		\quad & MAE/Rot (rad) All&  0.00062 & 0.00060  & \textbf{0.00054} & 0.01008\\ 
		\quad & Total Time (s) & \textbf{118}  &  148 & 262 & 2183\\
		\hline

		\multirow{5}{*}{Simulation 2} & MAE/Trans (m) First & 0.03938 & 0.03224 & \textbf{0.01984} & 0.09160\\
		\quad & MAE/Rot (rad) First&  0.00332 & 0.00220  & \textbf{0.00108} & 0.00314\\
		\quad & MAE/Trans (m) All& 0.01742  & 0.00858  & \textbf{0.00584} & 0.08018\\
		\quad & MAE/Rot (rad) All& 0.00064  & 0.00062  & \textbf{0.00052} & 0.00286\\ 
		\quad & Total Time (s)& \textbf{149}  & 193  & 321 & 2685\\
		\hline

		\multirow{5}{*}{Simulation 3} & MAE/Trans (m) First& 0.06708  &  0.02624  & \textbf{0.01776} & 0.03570\\
		\quad & MAE/Rot (rad) First&  0.00384 & 0.00164  & \textbf{0.00124}  & 0.00278\\
		\quad & MAE/Trans (m) All& 0.01586  & \textbf{0.00726}  & 0.00816 & 0.02082\\
		\quad & MAE/Rot (rad) All& 0.00100  & \textbf{0.00058}   & 0.00068 & 0.00102\\ 
		\quad & Total Time (s)& \textbf{125}  &  132 & 185 & 1041\\
		\hline
		\end{tabular}
		}
\end{table}

In this section, we perform ablation experiments on simulation datasets to analyze the impact of varying resolution settings in the first stage of the multi-resolution strategy on overall optimization performance.

We assess the accuracy and computational time using three simulation datasets, with the resolution in the second stage fixed at $s^{h} = 0.05$ m. The resolution ratios $r$ between the first and second stages are set to 2, 5, 10, and 20, respectively. To ensure consistency, a fixed selection range of $d=1.5$ m is applied uniformly across all datasets.

The results, shown in Table \ref{table_ablation}, reveal that $r=10$ achieves the best trade-off between time consumption and accuracy. While $r=20$ minimizes time consumption, it reduces the accuracy of poses in the first stage, adversely impacting final optimization accuracy. Conversely, $r=5$ improves pose accuracy in the first stage at the cost of higher time consumption but does not consistently enhance final accuracy. Notably, $r$ may need adjustment for other high-resolution settings.

\subsection{Using Submap Joining in Large-scale Environments}

We have demonstrated that our approach accurately and robustly handles normal-scale simulated and practical environments. In this section, we evaluate its efficiency and effectiveness in large-scale environments and long-term trajectories by integrating our Occupancy-SLAM algorithm with the proposed occupancy submap joining approach. The dataset is divided into multiple segments, where Algorithm \ref{alg_flowchart} is used to construct submaps, followed by applying the submap joining method in Section \ref{Sec_submap} to generate the optimized global occupancy map and robot trajectory.

\begin{figure}[t]
\centering \subfigure[b2] { \label{fig_large_b2}
\includegraphics[width=0.253\textwidth]{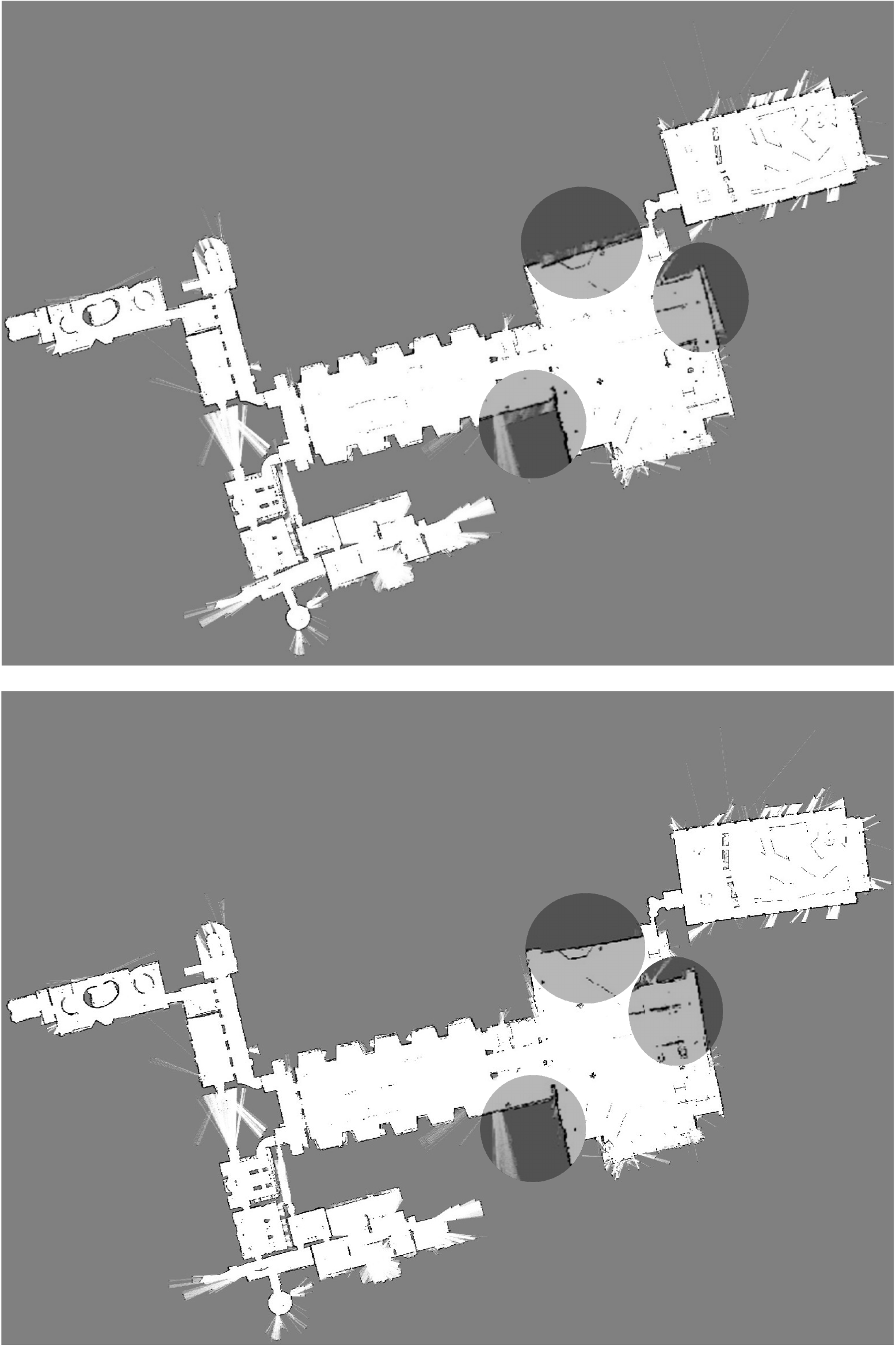}}\hspace{-0.4em}
\centering \subfigure[C3] {\label{fig_large_C3}
\includegraphics[width=0.2195\textwidth]{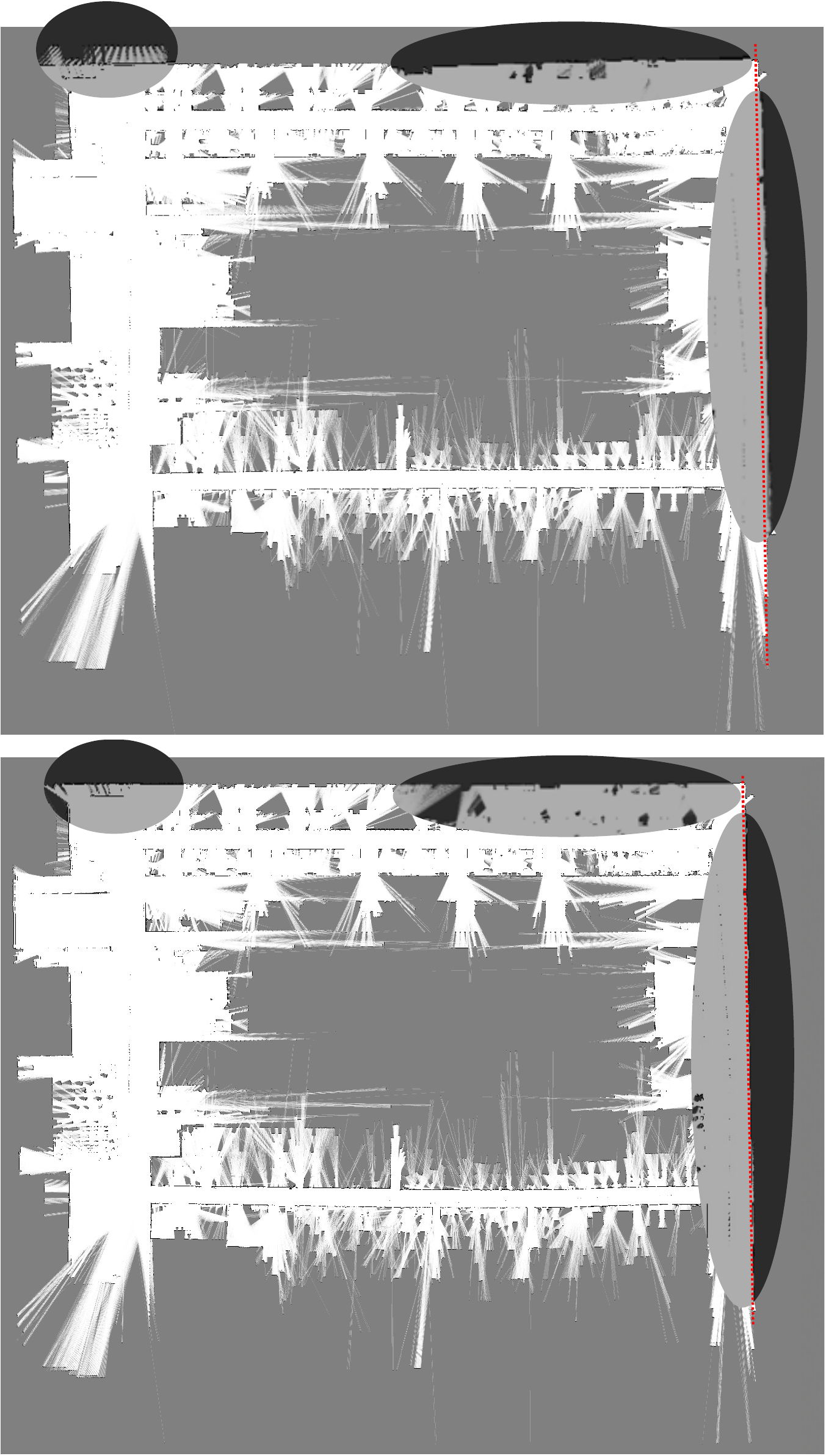}}
\caption{\label{fig_large_environment} Comparison of results between our method and Cartographer on two large-scale practical datasets. The first row shows Cartographer's results, while the second row shows ours. In (b), the red dotted lines serve as references, highlighting that Cartographer's right wall appears more curved, whereas our result aligns more closely with a straight line.}
\end{figure}

We validate our method on two large-scale datasets, Deutsches Museum b2 \cite{hess2016real} and C3, and compare it with Cartographer. As shown in Fig. \ref{fig_large_environment}, our occupancy grid maps outperform those of Cartographer, demonstrating the capability of our method to handle large-scale environments and long-term trajectories effectively. 


\subsection{Computational Complexity Analysis}
In this section, we analyze the computational complexity and evaluate the time consumption of our method using large-scale datasets.

The computational cost of the Gauss–Newton method for solving the joint optimization of local maps and poses in (\ref{Least Squares}), as well as submap joining in (\ref{eq_NLLS_joining}), is primarily determined by computing the Jacobian matrix $\mathbf{J}$, constructing and solving the sparse linear system in (\ref{Gauss-Newton}) \cite{konolige2008frameslam}. Since the formulations of the NLLS problems differ between these two methods, we analyze their computational complexities separately.


In the case where odometry inputs are not considered, the objective function of the joint optimization problem within each local submap $\mathbb{M}_{L_{i_L}}$ consists of the observation and smoothing terms. Let ${\lambda(\mathbb{S}^L_{i_L})}$ and $n^L_{i_L}$ denote the number of sampling points $\mathbb{S}^L_{i_L}$ and the number of poses within the $i_L$-th submap, respectively. Let $c^{{i_L}}_w \times c^{{i_L}}_h$ represent the size of the $i_L$-th submap. Then, the number of residual terms in the objective function is $\mathfrak{d}_{row} =\lambda(\mathbb{S}^L_{i_L})+3(n^L_{i_L}-1)+2c^{{i_L}}_w  c^{{i_L}}_h+c^{{i_L}}_w+c^{{i_L}}_h$, and the size of state vector is $\mathfrak{d}_{col}=3n^L_{i_L}+(c^{{i_L}}_w+1)(c^{{i_L}}_h+1)$. Considering Jacobian of the smoothing term $\mathbf{J}_S$ can be pre-calculated, the number of non-zero elements of the Jacobian matrix to be computed per iteration is $\mathfrak{d}_J = 7\lambda(\mathbb{S}^L_{i_L}) + 6(n^L_{i_L}-1)$. Therefore, the per-iteration computation complexities of the Jacobian calculation, constructing (\ref{Gauss-Newton}) and solving (\ref{Gauss-Newton}) are $\mathcal{O}(\mathfrak{d}_J)$, $\mathcal{O}(\mathfrak{d}_{J}\mathfrak{d}_{col})$, and $\mathcal{O}({\mathfrak{d}^3_{col}})$, respectively. The total computation complexity per iteration for the optimization problem is $\mathcal{O}(\mathfrak{d}_{J}+\mathfrak{d}_J{\mathfrak{d}_{col}}+\mathfrak{d}^3_{col})$. Due to our proposed multi-resolution joint optimization strategy and keyframe selection, both $\mathfrak{d}_{J}$ and $\mathfrak{d}_{col}$ remain small during the first and second stages of optimization, making the computation time manageable.

For our submap joining algorithm, the number of observations depends on the total number of cell vertices in the global occupancy map that are observed across all submaps, denoted as $\mathfrak{d}_{obs}^{G}$. Considering that some cell vertices are observed repeatedly from different submaps, this number slightly exceeds the total number of non-unknown cell vertices in the global map. The number of non-zero elements in Jacobian matrix is therefore $\mathfrak{d}_J^G = 4\mathfrak{d}_{obs}^G$, and the size of state vector is $\mathfrak{d}_{col}^{G} = 3n_L+(c_w^G+1)(c_h^G+1)$. The computation complexity per iteration is $\mathcal{O}(\mathfrak{d}_{J}^G+\mathfrak{d}_J^G{\mathfrak{d}_{col}^G}+{\mathfrak{d}_{col}^G}^3)$.
Although the global occupancy map is typically large, the sub-matrix of Hessian w.r.t. the global occupancy map is diagonal. To speed up computation, we apply the Schur complement \cite{zhang2006schur} to make the normal equation solving highly efficient.

To validate our computational complexity analysis, we evaluate the time consumption of our method using both all frames and selected key frames in large-scale environments and compare it with Cartographer. On Museum b2 dataset, Cartographer takes 1424 seconds, while our method takes 1250 seconds when using all frames and 363 seconds with selected key frames. On C3 dataset, Cartographer requires 610 seconds, while our method takes 742 seconds with all frames and 236 seconds with selected key frames. Notably, in both datasets, the submap joining step in our method accounts for less than 10 seconds of the total runtime. These results demonstrate that our method achieves runtime comparable to Cartographer when using all frames, and significantly lower runtime when using selected key frames. This confirms the effectiveness and efficiency of our multi-resolution joint optimization strategy and submap joining approach. 

\section{Preliminary Results in 3D Case}\label{sec_3d}

While this paper primarily focuses on demonstrating the benefits of jointly optimizing the robot poses and occupancy map in 2D, we also present some preliminary 3D results to illustrate that our idea can be extended to 3D applications.

\subsection{Extension of the Algorithms to 3D Case}

Our approach for jointly optimizing robot poses and the occupancy map extends naturally to 3D, where the information, robot poses, and occupancy maps are all represented in 3D. Most problem formulations and algorithms can be adapted with minor adjustments. 

For our local map and poses optimization method, observations transition from 2D laser scans to 3D LiDAR scans, robot poses and odometry involve 6 degree-of-freedom (DoF), and the map representation becomes 3D. Consequently, (\ref{eq_interp}) and (\ref{eq_NP}) need to be replaced from bilinear to trilinear interpolation and its inverse operation. 
For the objective function (\ref{eq_objective_func}), the odometry term (\ref{eq_odometry_term}) should be replaced with a 6 DoF odometry term for 3D, and the smoothing term (\ref{eq_smoothing_term}) should include a smoothing penalty for the z-axis, Jacobians $\mathbf{J}_P$, $\mathbf{J}_M$, $\mathbf{J}_O$, and $\mathbf{J}_S$ described in Appendices need to be adjusted accordingly. 

The submap joining problem in 3D remains largely similar to the 2D case, with the primary difference being that the projection relation extends from 2D-2D to 3D-3D, enabling the estimation of 6 DoF poses and the 3D global occupancy map in the NLLS problem (\ref{eq_NLLS_joining}).

\subsection{3D Experimental Results}
\subsubsection{Evaluation metrics and state-of-the-art methods}
We evaluate our method in 3D using absolute trajectory error (MAE and RMSE), aligning estimated trajectories with ground truth via EVO \cite{grupp2017evo}, as used in \cite{liu2023large,rosinol2021kimera}. For all experiments, we use the dataset-provided odometry as initialization when available. Otherwise, we use FAST-LIO2 \cite{xu2022fast} to obtain the odometry. Our method is compared against state-of-the-art methods, including BALM2 \cite{liu2023efficient}, HBA \cite{liu2023large}, and Voxgraph \cite{reijgwart2019voxgraph}. BALM2 jointly optimizes the planar feature parameters extracted from point clouds and the robot's poses. Building on this idea, HBA introduces a hierarchical bundle adjustment framework to improve scalability for large-scale environments. Voxgraph builds SDF-based submaps from point clouds, performs SDF-to-SDF registration for relative alignment, and incrementally optimizes submap frames. HBA and Voxgraph target large-scale environments, while BALM2 is designed for normal-scale environments.  

\subsubsection{Datasets}
We evaluate our method on three real-world datasets. (1) The Newer College Dataset \cite{ramezani2020newer}: The first five sequences from the \textit{shorter experiment}, collected with a handheld Ouster OS-1 LiDAR scanner at New College, Oxford. The environment includes lawns, buildings, a tunnel, and a garden. Ground truth poses are derived from high-precision 3D maps constructed using a BLK360 LiDAR scanner, achieving centimeter-level accuracy.
(2) KITTI Dataset \cite{Geiger2013IJRR}: Sequence 07, a demo dataset used in HBA, recorded with a vehicle-mounted Velodyne HDL-64E LiDAR. Ground truth poses are provided by RTK-GPS/INS.
(3) Arche Dataset \cite{reijgwart2019voxgraph}: A demo dataset for Voxgraph, collected using an Ouster OS1 LiDAR mounted on a hexacopter MAV in a disaster area. Ground truth positions are provided by an RTK-GNSS system. 

The Newer College Dataset is used to evaluate high-precision performance in normal-scale environments, while the KITTI and Arche datasets are used to assess performance in large environments with long trajectories.

\begin{figure*}[t]
\centering
\includegraphics[width=0.99\textwidth]{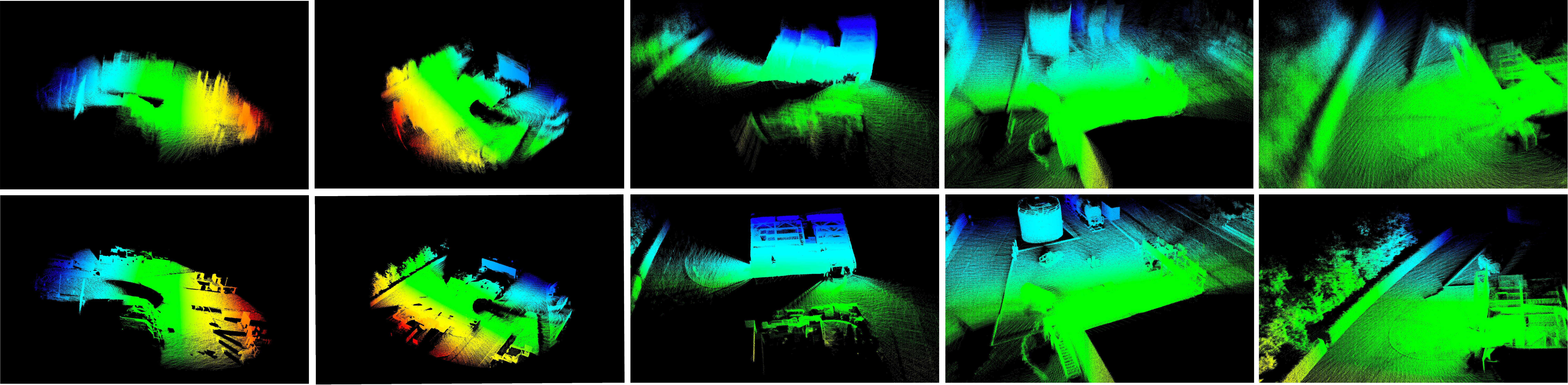}
\caption{\label{fig_3d_pointcloud} Some local point cloud maps from the Arche dataset. The first row shows point cloud maps generated using odometry from ROVIO, which is also used for submap construction in Voxgraph. The second row shows maps generated with our optimized poses using the same LiDAR scans. BALM2 fails to produce results in all these local environments when given the same odometry and scan inputs.}
\end{figure*}
 
\subsubsection{Experiments on normal-scale environments}
We evaluate the performance of our method without submap joining in normal-scale environments.

We first compare FAST-LIO2, BALM2, and our method on the first five sequences of The Newer College Dataset, which encompass all scenarios within the dataset. As shown in Table \ref{tab_comparison_3d_local}, our method outperforms BALM2 across all metrics except RMSE in Seq. 1, and significantly outperforms the odometry inputs from FAST-LIO2 in all metrics. 

\begin{table}[t]
		\centering
		\caption{Absolute Trajectory Error (MAE/RMSE, Meters) in Normal-scale Environments for Different 3D Methods.}
		\label{tab_comparison_3d_local}
		\setlength{\tabcolsep}{0.6 mm}{
		\begin{tabular}{lcccccc}\toprule
		Method	& Seq. 0 & Seq. 1 & Seq. 2 & Seq. 3 & Seq. 4 \\ \hline
		FAST-LIO2 & 0.518/0.717 & 0.181/0.202  & 0.121/0.132 &0.188/0.200&0.571/0.723\\
		BALM2 & 0.283/0.326 & 0.112/\textbf{0.123}  & 0.104/0.109 &0.144/0.158 &0.298/0.344 \\
        Ours & \textbf{0.185}/\textbf{0.232}  & \textbf{0.097}/\textbf{0.123}  & \textbf{0.091}/\textbf{0.099} & \textbf{0.141}/\textbf{0.155} &\textbf{0.238}/\textbf{0.284}\\ \hline
		\end{tabular}
		}
\end{table}

We then assess robustness under challenging conditions using the Arche dataset, collected by a hexacopter MAV in an unstructured environment affected by vibrations, flight dynamics, and external disturbances. Local point cloud maps built using odometry from ROVIO \cite{bloesch2017iterated} (also used to construct submaps in Voxgraph) are shown in the top row of Fig. \ref{fig_3d_pointcloud}. For evaluation, we partition the dataset into several short sequences, each lasting 10–20 seconds. BALM2 fails on all but the initial and final segments due to a lack of planar features of adequate quality for optimization, whereas our method performs reliably across all sequences. The bottom row of Fig. \ref{fig_3d_pointcloud} illustrates some of our results, demonstrating that our method is robust in 3D and does not rely on environmental assumptions. Additionally, the results confirm that our method achieves significantly higher pose accuracy than ROVIO.

\subsubsection{Experiments on large-scale environments}\label{sec_experiment_vox} 

We evaluate our method in large-scale environments with long trajectories using the KITTI and Arche datasets, comparing it with HBA and Voxgraph. For this experiment, we first build submaps by jointly optimizing poses and maps within each submap, then apply our submap joining algorithm to jointly optimize submap frame poses and the global occupancy map.

\begin{figure}[tbp]
\centering \subfigure[KITTI] {\label{fig_trajectory_1}
\includegraphics[height=0.15\textwidth]{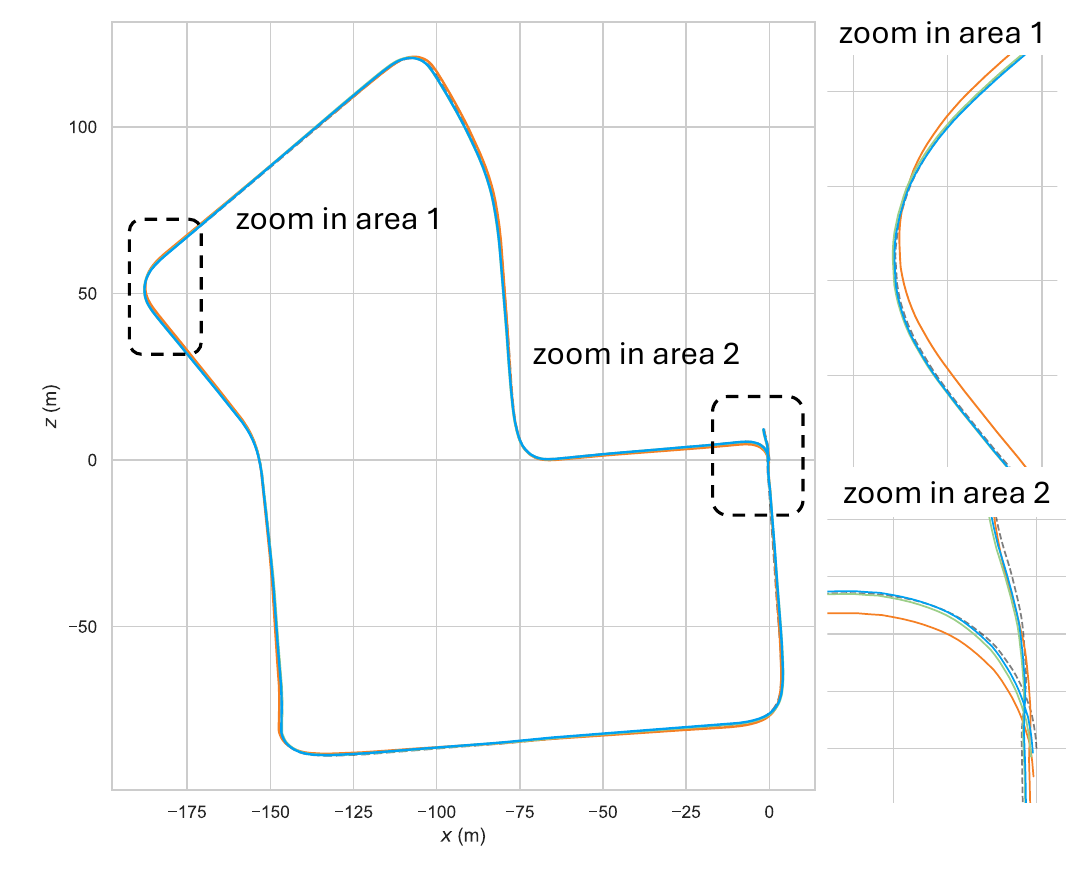}}
\centering \subfigure[Arche] {\label{fig_trajectory_2}
\includegraphics[height=0.15\textwidth]{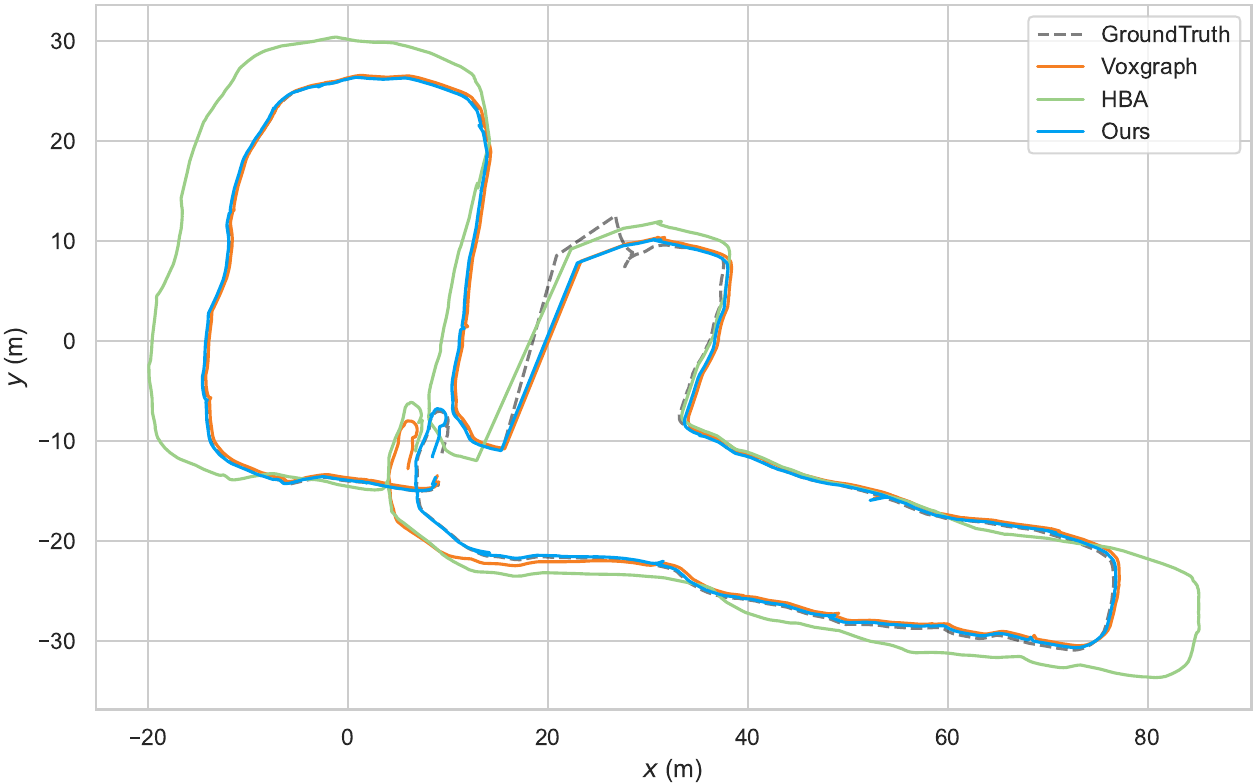}}
\caption{\label{fig_trajectory_3d} Robot trajectory results in large-scale environments. (a) and (b) show the trajectories of ground truth, Voxgraph \cite{reijgwart2019voxgraph}, HBA \cite{liu2023large}, and our method on the KITTI and Arche datasets, respectively.}
\end{figure}

Table \ref{tab_comparison_3d_large} summarizes the absolute trajectory error (in MAE and RMSE), showing that our method achieves the best performance on both datasets. Fig. \ref{fig_trajectory_3d} illustrates that our approach achieves the most accurate global trajectories. Notably, our method significantly outperforms Voxgraph on the KITTI dataset and HBA on the Arche dataset. Voxgraph performs poorly on KITTI due to its reliance on relative measurements from SDF-to-SDF registration, which requires substantial overlap between submaps—often lacking in autonomous driving scenarios. In contrast, our submap joining algorithm jointly optimizes submap poses and the global occupancy map, avoiding this limitation. HBA underperforms on the Arche dataset due to its dependence on planar features, which are difficult to extract in unstructured environments and during MAV motion. In such scenarios, odometry and point cloud quality degrade, and the planarity assumption often breaks down in non-urban environments such as disaster areas.

\begin{table}[t]
		\centering
		\caption{Absolute Trajectory Error (MAE/RMSE, Meters) in Large-scale Environments for Different 3D Methods}
		\label{tab_comparison_3d_large}
		\setlength{\tabcolsep}{7mm}{
		\begin{tabular}{lcc}\toprule
		Method & KITTI & Arche  \\ \hline
		HBA \cite{liu2023large} & 0.342/0.364 & 4.123/4.789  \\
        Voxgraph \cite{reijgwart2019voxgraph} &0.926/1.002 & 0.700/0.833 \\
        Ours & \textbf{0.315}/\textbf{0.339}  & \textbf{0.275}/\textbf{0.378} \\ \hline
		\end{tabular}
		}
\end{table}

\subsection{Discussion}
The experimental results in this section demonstrate that our proposed idea of jointly optimizing the robot pose and the occupancy map also yields improved solutions in 3D scenarios. However, several challenges remain. For instance, 3D LiDAR point clouds are often sparse, particularly in the vertical direction, which can cause observability issues in the optimization. This sparsity also results in inhomogeneous observation density, complicating the accurate representation of the 3D environment in occupancy maps. Furthermore, the high dimensionality of 3D maps introduces significant computational burdens in large-scale SLAM, requiring more efficient solving methods.

To address these challenges, several potential solutions can be explored. First, adopting compact representations such as octree structures (e.g., OctoMap \cite{hornung2013octomap} and \cite{vespa2019adaptive}) can enhance efficiency. Second, combining local map and pose optimization with hierarchical optimization and submap joining methods can further reduce computational cost. Finally, leveraging continuous representations of 3D occupancy maps enables more precise gradient computation, thereby providing better guidance for optimization.

\section{Conclusion} \label{Sec_conclusion}
In this paper, we propose Occupancy-SLAM algorithm, which solves robot poses and occupancy map simultaneously. To enhance efficiency and robustness, we introduce a multi-resolution strategy. The first stage jointly optimizes poses and a low-resolution occupancy map to quickly achieve relatively accurate pose estimates, which then serves as the initial guess for the second stage. The second stage refines poses and a subset of the high-resolution map, focusing on critical boundary areas. We further extend this framework to an occupancy grid-based submap joining algorithm, addressing challenges in large-scale environments and long-term trajectories. Results from both simulated and real-world datasets demonstrate that our method achieves more accurate pose and map estimates than state-of-the-art approaches.

Our findings show that solving poses and occupancy maps simultaneously yields more accurate results compared to first solving pose-graph SLAM and then constructing the map. This joint optimization approach has the potential to revolutionize occupancy map based SLAM frameworks.

The proposed method acts as a batch optimization approach for obtaining high-quality robot poses and maps. Unlike incremental or online methods, batch optimization provides greater accuracy, which is particularly advantageous for applications requiring high-quality maps rather than real-time operation (e.g., offline map creation for precise future localization). Despite typical drawbacks of batch optimization, such as higher computational costs, trajectory-length-dependent complexity, and reliance on accurate initial guesses, our method effectively overcomes these limitations: 1) our method is efficient due to the proposed multi-resolution joint optimization strategy, and the computation time is comparable to online methods; 2) our method can use selected keyframes to further reduce the computational cost without losing too much accuracy; 3) our proposed occupancy submap joining approach can overcome the limitation that the computational complexity related to the length of the robot trajectories; and 4) our method is very robust to the initial guess and can be initialized from odometry inputs or scan matching, so it does not require initialization from the result of incremental/online methods.    

In future work, we aim to further explore problem formulation and solution techniques for 3D environments, with the goal of developing more efficient and robust algorithms capable of addressing various challenges of 3D SLAM.


{\appendix

The Jacobian $\mathbf{J}$ in (\ref{Gauss-Newton}) consists of four parts, i.e. the Jacobian of the observation term w.r.t. the robot poses $\mathbf{J}_P$ (See Appendix \ref{Sec_J_P}), the Jacobian of the observation term w.r.t. the occupancy map $\mathbf{J}_M$ (See Appendix \ref{Sec_J_D}), the Jacobian of the odometry term w.r.t. robot poses $\mathbf{J}_O$ (See Appendix \ref{Sec_J_O}) and the Jacobian of the smoothing term w.r.t. the occupancy map $\mathbf{J}_S$ (See Appendix \ref{Sec_J_S}). In addition, the difference in the calculation of Jacobians between Algorithm \ref{alg_1} and Algorithm \ref{alg_3} is shown in Appendix \ref{Sec_J_Select}. 

\subsection{Jacobian of the Observation Term w.r.t. Robot Poses}\label{Sec_J_P}

The Jacobian of $F_{ij}^Z(\mathbf{x})$ in the observation term w.r.t. the robot poses $\mathbf{x}^P_i$ can be calculated using the chain rule
\begin{equation}
	\begin{aligned}
		\mathbf{J}_P=\frac{ \partial F_{ij}^Z(\mathbf{x}) }{ \partial \mathbf{x}^P_i } = \frac{\partial F_{ij}^Z(\mathbf{x}) }{ \partial \mathbf{p}_{ij} } \cdot \frac{\partial \mathbf{p}_{ij}  }{ \partial \mathbf{x}^P_i}.	
	\end{aligned}
\end{equation}
Here, $\dfrac{\partial \mathbf{p}_{ij}  }{ \partial \mathbf{x}^P_i}$ is calculated as
\begin{equation}
\dfrac{\partial \mathbf{p}_{ij}}{\partial \mathbf{x}^P_i}=\left[\begin{array}{ll}
\dfrac{\partial \mathbf{p}_{ij}}{\partial \mathbf{t}_i} & \dfrac{\partial \mathbf{p}_{ij}}{\partial \theta_i}
\end{array}\right]=\dfrac{1}{s} \left[\begin{array}{ll}
\mathbf{E}_{2} & \left(\mathbf{R}_i^{\prime}\right)^{\top} \mathbf{p}_{ij}
\end{array}\right],
\end{equation}
where $\mathbf{R}_i^\prime$ is the derivative of the rotation matrix $\mathbf{R}_i$ w.r.t. rotation angle $\theta_i$, and $\mathbf{E}_2$ means $2 \times 2$ identity matrix.

$\dfrac{\partial F_{ij}^Z(\mathbf{x}) }{ \partial \mathbf{p}_{ij} }$ can be calculated by
\begin{equation}
\dfrac{\partial F_{ij}^Z(\mathbf{x}) }{ \partial \mathbf{p}_{ij} } = \dfrac{1}{N(\mathbf{p}_{ij})} \dfrac{\partial M(\mathbf{p}_{ij})}{\partial \mathbf{p}_{ij}}.
\end{equation}
Here $\dfrac{\partial M(\mathbf{p}_{ij})}{\partial \mathbf{p}_{ij}}$ can be considered as the gradient of the occupancy map at point $\mathbf{p}_{ij}$, which can be approximated by bilinear interpolation of the gradients of the occupancy at the four adjacent cell vertices $\mathbf{\nabla} M(\mathbf{m}_{wh}),\cdots,\mathbf{\nabla} M(\mathbf{m}_{(w+1)(h+1)})$ around $\mathbf{p}_{ij}$ as
\begin{equation} 
\dfrac{\partial M(\mathbf{p}_{ij})}{\partial \mathbf{p}_{ij}}= 
\left[
\begin{aligned}
a_1b_1\\a_0b_1\\a_1b_0\\a_0b_0\\
\end{aligned}\right]^\top
\left[
\begin{aligned}
&\mathbf{\nabla} M(\mathbf{m}_{wh})\\&\mathbf{\nabla} M(\mathbf{m}_{(w+1)h})\\&\mathbf{\nabla} M(\mathbf{m}_{w(h+1)})\\&\mathbf{\nabla} M(\mathbf{m}_{(w+1)(h+1)})
\end{aligned}\right].\label{eq_14}
\end{equation} 
The gradient $\nabla M(\cdot)$ at all cell vertices can be easily calculated from $\mathbf{x}^M$ in the state. The bilinear interpolation used in (\ref{eq_14}) is similar to the method in (\ref{eq_interp}).

We assume the robot poses $\mathbf{x}^P$ change slightly in each iteration. To reduce the computational complexity, the hit map $\mathbb{N}$ is considered as constant during each iteration and is recalculated based on the current robot poses. The derivative of $N(\mathbf{p}_{ij})$ is therefore omitted from the Jacobian computation. 

\subsection{Jacobian of the Observation Term w.r.t. Occupancy Map}\label{Sec_J_D}
Based on (\ref{eq_interp}), the Jacobian of $F_{ij}^Z(\mathbf{x})$ in the observation term w.r.t. the map part in state vector $\mathbf{x}^{M}$ can be calculated as

\begin{equation}
\begin{aligned}
\mathbf{J}_M & = \dfrac{\partial F_{ij}^Z(\mathbf{x})}{\partial \left[ {M}(\mathbf{m}_{wh}),\cdots, {M}(\mathbf{m}_{(w+1)(h+1)}) \right]^\top}\\
&= \dfrac{1}{N(\mathbf{p}_{ij})}\dfrac{\partial M(\mathbf{p}_{ij})}{\partial \left[ {M}(\mathbf{m}_{wh}),\cdots, {M}(\mathbf{m}_{(w+1)(h+1)}) \right]^\top}\\ 
&= \dfrac{\begin{bmatrix}
a_1b_1,a_0b_1,a_1b_0,a_0b_0
\end{bmatrix}}{N(\mathbf{p}_{ij})}
\end{aligned}
\end{equation}
where $\mathbf{m}_{wh}, \cdots, \mathbf{m}_{(w+1)(h+1)}$ are the four nearest cell vertices to $\mathbf{p}_{ij}$ in occupancy map $\mathbb{M}$, and $a_0,a_1,b_0$ and $b_1$ are defined in (\ref{eq_interp}).

\subsection{Jacobian of the Odometry Term}\label{Sec_J_O}
The Jacobian of $F_i^O(\mathbf{x})$ in the odometry term (\ref{eq_odometry_term}) is the partial derivative w.r.t. the robot poses $\mathbf{x}^P$ since it is not related to $\mathbf{x}^M$. Therefore, the Jacobian $\mathbf{J}_O$ can be calculated as
\begin{equation}
\begin{aligned}
\mathbf{J}_O &= \frac{\partial F_i^O(\mathbf{x})}{\partial \left[ {\mathbf{x}^P_{i-1}}^\top, {\mathbf{x}^P_i}^\top \right]^\top }\\ 
&=\begin{bmatrix}
	 \dfrac{\partial F_i^O(\mathbf{x})}{\partial \mathbf{t}_{i-1}} &
	 \dfrac{\partial F_i^O(\mathbf{x})}{\partial \theta_{i-1}} &
	 \dfrac{\partial F_i^O(\mathbf{x})}{\partial \mathbf{t}_i} &
	 \dfrac{\partial F_i^O(\mathbf{x})}{\partial \theta_i}
 \end{bmatrix} 
 \\
 &=\begin{bmatrix}
 	-\mathbf{R}_{i-1} & \mathbf{R}_{i-1}^\prime(\mathbf{t}_i-\mathbf{t}_{i-1}) & \mathbf{R}_{i-1} &\mathbf{0}_2\\
 	\mathbf{0}_2^\top & -1 & \mathbf{0}_2^\top & 1\\
 \end{bmatrix}
\end{aligned}
 \end{equation}
in which $\mathbf{0}_2$ means $2 \times 1$ zero vector.
 
\subsection{Jacobian of the Smoothing Term}\label{Sec_J_S}

The Jacobian $\mathbf{J}_S$ is the derivative of (\ref{eq_smoothing_term}) w.r.t. $\mathbf{x}^M$ 
due to it is not related to the robot poses $\mathbf{x}^P$. It should be mentioned that $F^S(\mathbf{x})$ is linear w.r.t. $\mathbf{x}^M$, i.e.,
\begin{equation}
F^S(\mathbf{x}) = \mathbf{A} \left[ {M}(\mathbf{m}_{00}),\cdots,{M}(\mathbf{m}_{c_wc_h}) \right]^\top
\end{equation}
where the coefficient matrix $\mathbf{A}$ is sparse and with nonzero elements $1$ or $-1$. An example of the coefficient matrix can be shown as
\begin{equation}\label{eq_A}
	\mathbf{A} = \begin{bmatrix}
    \vdots &\vdots  &\vdots  &\vdots  &\vdots  &\vdots  &\vdots &\vdots\\
 	\mathbf{0}^\top & 1 & -1 & 0 & \mathbf{0}^\top & 0 & 0 & \mathbf{0}^\top\\
 	\mathbf{0}^\top & 1 & 0  & 0 & \mathbf{0}^\top & -1 & 0 & \mathbf{0}^\top\\
 	\mathbf{0}^\top & 0 & 1 & -1 & \mathbf{0}^\top & 0 & 0 & \mathbf{0}^\top\\
 	\mathbf{0}^\top & 0 & 1 & 0 & \mathbf{0}^\top & 0 & -1 & \mathbf{0}^\top\\
    \vdots &\vdots  &\vdots  &\vdots  &\vdots  &\vdots  &\vdots &\vdots\\
 \end{bmatrix}.
\end{equation}
Here $\mathbf{0}$ represents a zero vector with appropriate dimensions. The Jacobian of the smoothing term can be calculated as
\begin{equation}\label{eq_JS}
\mathbf{J}_S = \frac{\partial F^S(\mathbf{x})}{\partial \mathbf{x}^M } = \mathbf{A}.\\ 
\end{equation}
Since $\mathbf{A}$ is constant, $\mathbf{J}_S$ can be pre-calculated and directly used in the optimization as shown in Algorithm \ref{alg_1}.

\subsection{Jacobians in the Second Stage of Multi-resolution Strategy for Optimization}\label{Sec_J_Select}
In the second stage of the multi-resolution strategy (Algorithm \ref{alg_3}), the Jacobians to be calculated are similar to those in Algorithm \ref{alg_1}. A specific challenge arises in handling the selected cell vertices adjacent to the dropped cell vertices in the high-resolution map, particularly when calculating Jacobians $\mathbf{J}_P$ and $\mathbf{J}_S$.

 For Jacobian $\mathbf{J}_P$, partial derivatives w.r.t. all the cell vertices are required for (\ref{eq_14}). However, not all vertices are included in the state vector in the second stage, which makes it challenging to compute the partial derivatives w.r.t. some cell vertices because their surrounding nodes are discarded. From a semantic perspective, the discarded cell vertices share the same occupancy state as the edge vertices, which justifies their exclusion. Consequently, the gradients at these edge vertices are assumed to be negligible. Based on this assumption, the partial derivatives w.r.t. all edge cell vertices are set to zero when computing (\ref{eq_14}).
 
Jacobian $\mathbf{J}_S$ can be calculated using the same idea as (\ref{eq_JS}). In the second stage of our multi-resolution strategy, (\ref{eq_JS}) is reformulated as 
 \begin{equation}
 	\mathbf{J}_S = \frac{\partial F^S_{s}(\mathbf{x}^s)}{\partial \mathbf{x}^{sM} } = \mathbf{A}^{s}\\ 
 \end{equation}
 where $\partial F^S_{s}(\mathbf{x}^s)$ is similar to (\ref{eq_smoothing_term}), but only applies to vertices in $\mathbb{M}^s$. The coefficient matrix $\mathbf{A}^{s}$ has the same form as (\ref{eq_A}) but with dimension corresponding to the number of elements in $\mathbf{x}^{sM}$.  
 
 }

\bibliographystyle{IEEEtran}
\bibliography{references.bib}

\newpage



\begin{IEEEbiography}[{\includegraphics[width=1in,height=1.25in,clip,keepaspectratio]{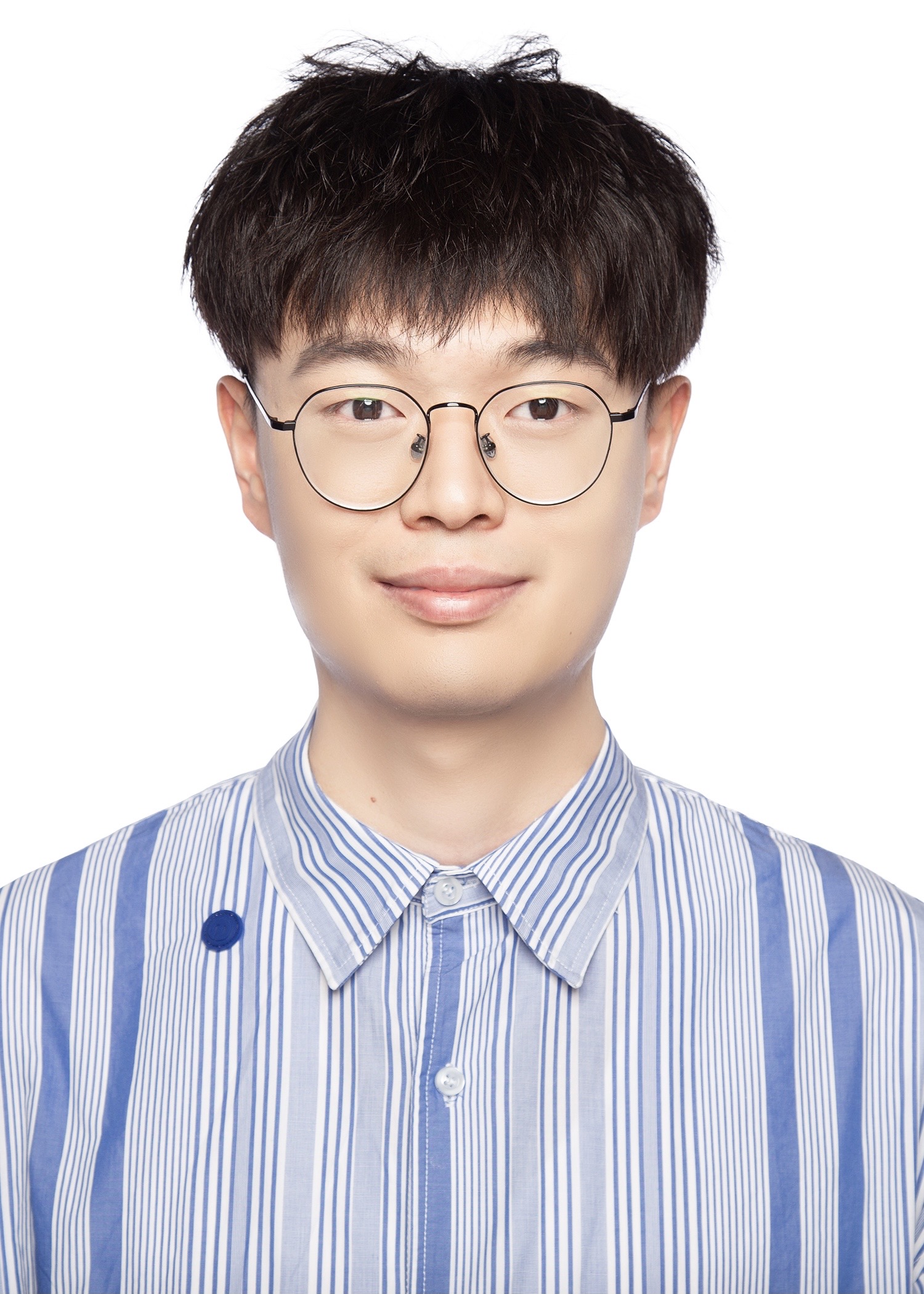}}]{Yingyu Wang} is a PhD candidate in the Robotics Institute at the University of Technology Sydney, where he conducts research under the supervision of Prof. Shoudong Huang and A/Prof. Liang Zhao. He earned his bachelor’s degree from the Chengdu University of Technology, Chengdu, China in 2017 and his master’s degree from the Ocean University of China, Qingdao, China in 2020. His research interests primarily focuses on simultaneous localization and mapping (SLAM), mapping, perception, and state estimation for autonomous robotics systems. 
\end{IEEEbiography}

\begin{IEEEbiography}
[{\includegraphics[width=1in,height=1.25in,clip,keepaspectratio]{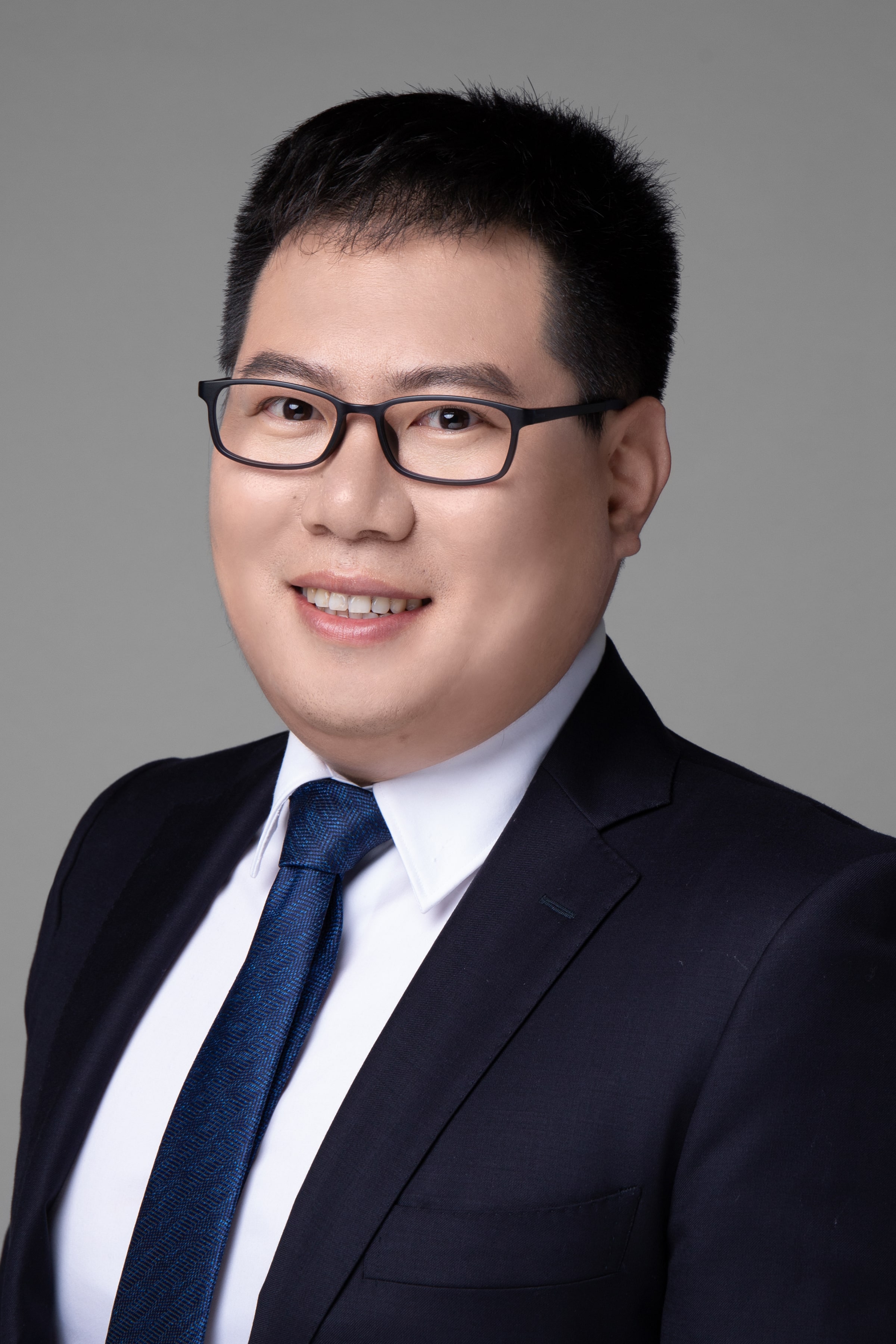}}]{Liang Zhao} (Member, IEEE) received the Ph.D. degree in photogrammetry and remote sensing from Peking University, Beijing, China, in January 2013. From 2014 to 2016, he worked as a Postdoctoral Research Associate with the Hamlyn Centre for Robotic Surgery, Department of Computing, Faculty of Engineering, Imperial College London, London, U.K. From 2016 to 2024, he was a Senior Lecturer and the Director of Robotics in Health at the UTS Robotics Institute, Faculty of Engineering and Information Technology, University of Technology Sydney, Australia. He is currently a Reader in Robot Systems in the School of Informatics, The University of Edinburgh, United Kingdom. His research interests include surgical robotics, autonomous robot simultaneous localization and mapping (SLAM), monocular SLAM, aerial photogrammetry, optimization techniques in mobile robot localization and mapping and image guided robotic surgery. 

Dr. Zhao is an Associate Editor for IEEE Transactions on Robotics, ICRA, and IROS.

\end{IEEEbiography}

\begin{IEEEbiography}
[{\includegraphics[width=1in,height=1.25in,clip,keepaspectratio]{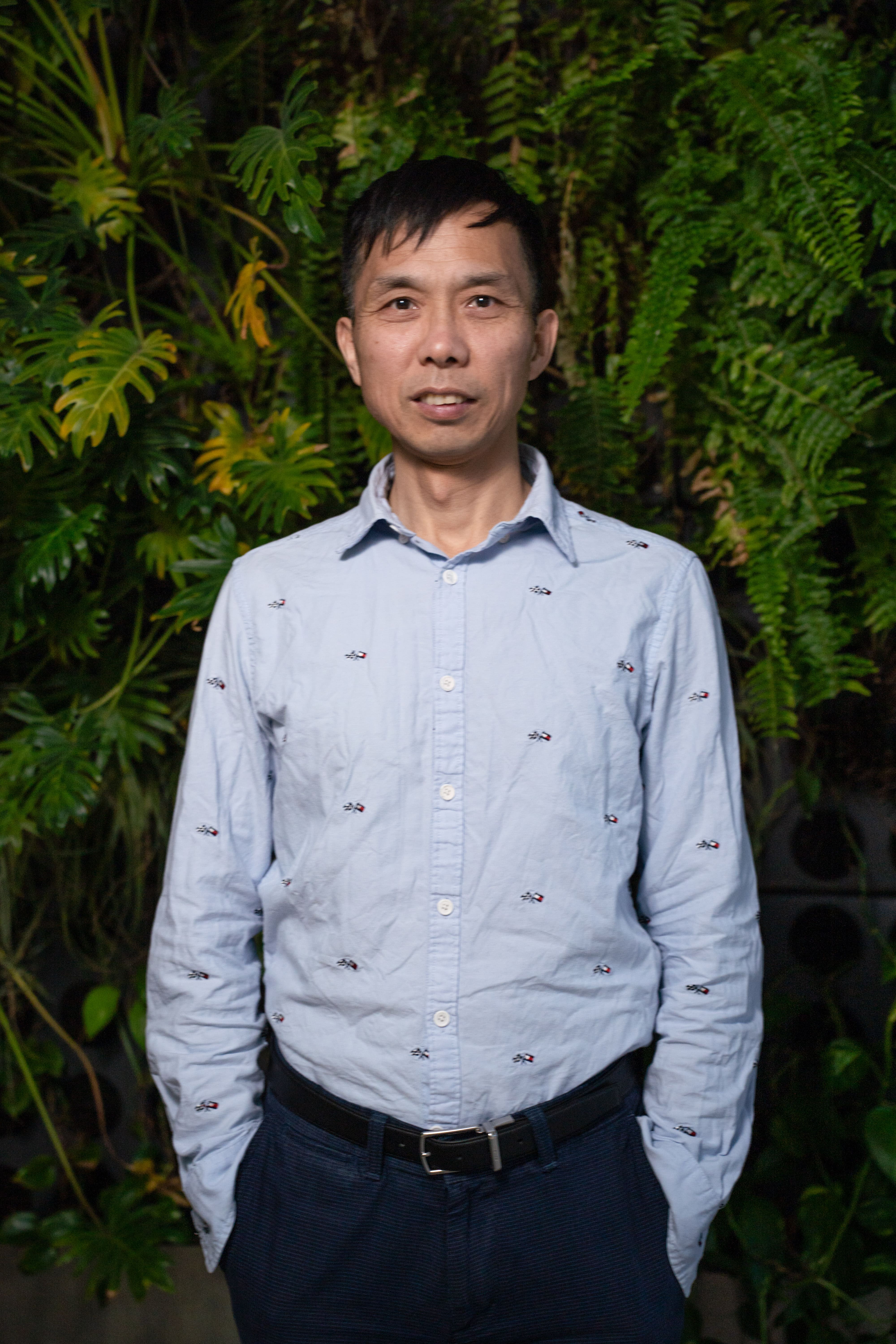}}]{Shoudong Huang} (Senior Member, IEEE) received the Bachelor and Master degrees in Mathematics, Ph.D. in Automatic Control from Northeastern University, PR China in 1987, 1990, and 1998, respectively. He is currently a Professor in the Robotics Institute, University of Technology Sydney, Australia. He is now an Associate Editor for the International Journal of Robotics Research. His research interests include nonlinear system state estimation and control, mobile robot simultaneous localization and mapping (SLAM), and surgical robotics. 

\end{IEEEbiography}



\vfill

\end{document}